\documentclass[runningheads]{llncs}
\usepackage{eccv}

\usepackage{eccvabbrv}

\usepackage{graphicx}
\usepackage{booktabs}
\usepackage{multirow}
\usepackage{wrapfig}

\usepackage[accsupp]{axessibility}

\usepackage{hyperref}
\usepackage{orcidlink}

\usepackage{algorithm}
\usepackage{algpseudocode}
\usepackage{subcaption}

\usepackage{array}

\usepackage{color}
\usepackage{xcolor}
\usepackage{soul}
\usepackage[normalem]{ulem}

\def\mypar#1{\vspace{0.15cm}\noindent{\bf #1.}}

\IfFileExists{darkmode.tex}{
    \usepackage{pagecolor}
    \definecolor{myfg}{gray}{0.94} 
    \definecolor{mybg}{gray}{0}
    \input{darkmode.tex} 
    \pagecolor{mybg}
    \color{myfg}
}{} 

\usepackage{amsmath,amssymb,amsbsy,xspace}

\def\<{\langle}
\def\>{\rangle}

\makeatletter
\DeclareRobustCommand\onedot{\futurelet\@let@token\@onedot}
\def\@onedot{\ifx\@let@token.\else.\null\fi\xspace}

\makeatother

\begin{document}

\title{Policy-based Tuning of Autoregressive Image Models with Instance- and Distribution-Level Rewards}
\titlerunning{Policy-based Tuning of AR Image Models}

\author{
Orhun Bu\u{g}ra Baran\inst{1}\orcidlink{0000-0002-7153-6297} \and
Melih Kandemir\inst{2}\orcidlink{0000-0001-6293-3656} \and
Ramazan Gokberk Cinbis\inst{1,3}\orcidlink{0000-0003-0962-7101}
}

\authorrunning{Baran et al.}

\institute{
Department of Computer Engineering, Middle East Technical University, Ankara, T\"urkiye\\
\email{\{bugra,gcinbis\}@ceng.metu.edu.tr}
\and
Department of Mathematics and Computer Science (IMADA), University of Southern Denmark, Odense, Denmark\\
\email{kandemir@imada.sdu.dk}
\and
ROMER Robotics and AI Center, Middle East Technical University, Ankara, T\"urkiye
}

\maketitle
\footnotetext{Accepted at the European Conference on Computer Vision (ECCV), 2026.}
\begin{abstract}
Autoregressive (AR) models are highly effective for image generation, yet their standard maximum-likelihood estimation training lacks direct optimization for sample quality and diversity. While reinforcement learning (RL) has been used to align diffusion models, these methods typically suffer from output diversity collapse. Similarly, concurrent RL methods for AR models rely strictly on instance-level rewards, often trading off distributional coverage for quality. To address these limitations, we propose a lightweight RL framework that casts token-based AR synthesis as a Markov Decision Process, optimized via Group Relative Policy Optimization (GRPO). Our core contribution is the introduction of a novel distribution-level Leave-One-Out FID (LOO-FID) reward; by leveraging an exponential moving average of feature moments, it explicitly encourages sample diversity and prevents mode collapse during policy updates. We integrate this with composite instance-level rewards (CLIP and HPSv2) for strict semantic and perceptual fidelity, and stabilize the multi-objective learning with an adaptive entropy regularization term. Extensive experiments on LlamaGen and VQGAN architectures demonstrate clear improvements across standard quality and diversity metrics within only a few hundred tuning iterations. The results also show that the model can be updated to produce competitive samples even without Classifier-Free Guidance, and bypass its 2x inference cost.
\end{abstract}
    
\section{Introduction}
\label{sec:intro}

Autoregressive (AR) models have recently re-emerged as a strong and scalable framework for high-fidelity image generation. By modeling pixel or token sequences with Transformer decoders, they unify visual synthesis with the language modeling paradigm underlying GPT-style systems~\cite{gpt2}. Advances such as VQVAE and VQGAN tokenizers~\cite{vqvae,vqgan}, hierarchical scaling~\cite{parti,nextscale}, and continuous-token variants~\cite{fluid,mar} have enabled AR models like LlamaGen~\cite{llamagen} to match or even surpass diffusion models~\cite{diffusion} in both quality and throughput. Despite this progress, most AR generators remain trained solely via maximum likelihood estimation (MLE), which does not directly optimize semantic alignment or perceptual preference and thus limits post-training controllability.

\begin{wrapfigure}{r}{0.54\textwidth}
    \centering
    \includegraphics[width=\linewidth]{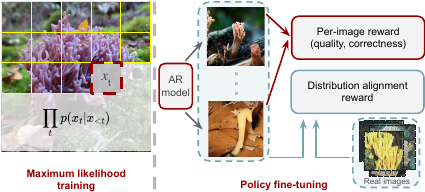}
    \caption{Maximum-likelihood training (Left): maximize true token $x_t$ probability from its groundtruth context. Proposed method (Right): maximize image-level correctness and distribution-level alignment. \label{fig:intro_figure}}
\end{wrapfigure}%
Reinforcement learning (RL) has proven central to shaping the behavior of large language models. Recent progress in large language models (LLMs) has been driven by post-training pipelines that combine supervised instruction tuning,
preference learning, and policy optimization using methods such as PPO~\cite{ppo}, DPO~\cite{dpo}, or GRPO~\cite{grpo}. These approaches produce models that better follow instructions and align with human preferences~\cite{rlhf,finetuninglmsfromhp,instructGPT}.
A parallel trend has emerged in diffusion-based image generation, 
where models originally trained with reconstruction losses are post-trained %
using external reward models such as CLIP~\cite{clip}, HPSv2~\cite{hpscorev2}, or ImageReward~\cite{imagereward}. 
Existing approaches include offline supervised finetuning~\cite{dpok}, differentiating through the denoising process~\cite{alignprop,deeprewardsupervision}, and preference or policy optimization via DDPO~\cite{ddpo}, DRaFT~\cite{draft}, or Diffusion-DPO~\cite{diffusiondpo}.

\mypar{Motivation} We investigate policy-based tuning as an effective post-training mechanism for autoregressive image generators. In contrast to diffusion models, AR decoders naturally follow a token-level policy, enabling stable PPO/GRPO-style updates~\cite{ppo,grpo} with minimal overhead. In addition, unlike the prior work on policy learning for diffusion models~\cite{ddpo,diffusiondpo,draft}, which are known to suffer from output diversity collapse~\cite{barcelo2024modecollapse,imageReFL}, our AR policy learning method allows for the explicit joint optimization of \emph{instance-level alignment} and \emph{distributional coverage}. Furthermore, by embedding instance level rewards and distribution alignment directly into the base policy, our framework produces high-quality samples without relying on Classifier-Free Guidance (CFG), effectively avoiding the $2\times$ computational overhead typically incurred during test-time generation.

\mypar{Contributions} We propose a lightweight RL framework that casts token-based image synthesis as an MDP and efficiently fine-tunes AR image generators via RL, with the following key contributions:
\begin{itemize}
  \item \textbf{Novel distribution-level LOO-FID reward.} To our knowledge, we are the first to explicitly optimize for distribution-level fidelity during AR policy fine-tuning. To overcome the severe instability and high variance of small-batch statistics, we introduce a leave-one-out (LOO) FID reward based on an Exponential Moving Average (EMA) of the generator's feature moments. This tracks each sample's marginal contribution to the global distribution, directly encouraging coverage and preventing mode collapse.
  \item \textbf{Instance-level alignment with adaptive regularization.} Alongside the distribution reward, we integrate composite instance-level signals (CLIP and HPSv2) to enforce strict semantic fidelity and perceptual preference. To stabilize this multi-objective policy learning, we propose a dynamic, adaptive entropy bonus that actively prevents premature convergence and preserves target token diversity throughout training.
  \item \textbf{Efficient improvements and a $\mathbf{2\times}$ inference speedup.} Our approach demonstrates strong generalizability across diverse architectures (LlamaGen~\cite{llamagen} and VQGAN~\cite{vqgan}), yielding clear improvements in both quantitative metrics (FID, IS, CLIPScore, Precision and Recall) and human evaluation. Furthermore, by fundamentally improving the base autoregressive policy, our method enables semantically aligned synthesis without relying on Classifier-Free Guidance (CFG), effectively bypassing the $2\times$ test-time overhead. 
\end{itemize}

Our implementation and pretrained models are publicly available.\footnote{
Project page and source code:
\url{https://github.com/bugrabaran/ar-policy-tuning}}

\section{Related Work}
\label{sec:relwork}

\mypar{Autoregressive image models}
AR modeling, foundational to language models~\cite{gpt2}, was first applied to images via pixel-level generation~\cite{pixelrnn,pixelcnn}. The developments in image tokenization~\cite{vqvae,vqgan} shifted the focus to modeling quantized image patches, enhancing both efficiency and perceptual quality. Subsequent advances brought further improvements: MaskGIT~\cite{maskgit} introduced masked parallel decoding with iterative denoising steps to accelerate generation; Parti~\cite{parti} demonstrated that large-scale AR training with rich token vocabularies can produce photorealistic results; and LlamaGen~\cite{llamagen} showed that scaling AR architectures can rival diffusion models~\cite{imagen,dalle}. 
More recent works focus on compressing sequences (e.g., next-scale prediction~\cite{nextscale}) or continuous latent representations (e.g., TiTok~\cite{titok}, Fluid~\cite{fluid}) to improve efficiency and fidelity. Our approach is orthogonal to these architectural innovations.

\mypar{Policy optimization for generative models} The recent success of large language models
(LLMs) has been strongly driven by policy optimization methods, such as Reinforcement Learning from Human Feedback
(RLHF)~\cite{rlhf, instructGPT}, which maps pretrained AR policies to human intent using reward models trained on human preferences. 
Subsequent works optimized preferences directly~\cite{dpo}, while more recently, Group Relative Policy Optimization (GRPO)~\cite{grpo} 
introduced an on-policy variant that samples multiple responses per prompt and uses their relative rewards to determine the policy gradient. 
This eliminates the value network while improving sample efficiency and stability.

Analogous techniques have been extended to guide diffusion models using reward models like CLIP~\cite{clip} and HPSv2~\cite{hpscorev2} via policy or preference optimization~\cite{ddpo,draft,diffusiondpo,deeprewardsupervision}. However, diffusion-based alignment often suffers from poor sample efficiency and severe diversity collapse, tending to overfit to narrow reward directions~\cite{ddpo,barcelo2024modecollapse,imageReFL}. While some approaches like AIG~\cite{aig} address this reward--diversity trade-off, they rely on modifying diffusion-specific noise schedules or sampling dynamics. In contrast, we formulate autoregressive image generation directly as a discrete policy learning problem.

\mypar{Policy optimization for AR models}
Recently, several concurrent, unpublished preprints have also explored applying reinforcement learning to autoregressive image generators~\cite{argrpo, liao2025vapi, ma2025stage, wang2025simplear, zhang2025groupcritical}. Among these, AR-GRPO~\cite{argrpo} serves as the most direct application of standard policy optimization, making it the natural baseline for our work. The remaining works focus on orthogonal improvements to the optimization pipeline itself, such as variational pixel-aware alignment~\cite{liao2025vapi}, similarity-aware advantage reweighting~\cite{ma2025stage}, multi-stage pretraining pipelines~\cite{wang2025simplear}, and critical-token selection in GRPO~\cite{zhang2025groupcritical}. Our work diverges fundamentally from these approaches: while the aforementioned methods rely strictly on instance/sequence-level rewards for maximization, our primary contribution introduces a novel distributional reward to guide the RL process. To the best of our knowledge, we are the first to explicitly optimize for distribution-level fidelity, encouraging sample diversity and statistical alignment directly during RL fine-tuning. Furthermore, we improve training stability by coupling this with adaptive entropy regularization. Importantly, because our distributional reward formulation is agnostic to the underlying RL optimizer, the GRPO-specific improvements proposed in these concurrent works are entirely orthogonal and can be readily integrated into our framework to yield complementary gains.

\mypar{Reward models for generative image models}
Recent research has shown that learned reward models can effectively align image generation with human aesthetic and semantic preferences. 
Early approaches leveraged CLIP~\cite{clip} directly as a reward function to steer optimization in generative models~\cite{clipguided_diffusion, vqganclip}.
Subsequent works introduced human-trained preference models such as ImageReward~\cite{imagereward}, Human Preference Score v2 (HPSv2)~\cite{hpscorev2}, PickScore~\cite{pickscore} and MANIQA~\cite{maniqa}, which learn regression-based reward functions over large annotated datasets to better capture human judgments of realism and prompt fidelity. 
These reward models have become central to diffusion fine-tuning frameworks including DDPO~\cite{ddpo}, Diffusion-DPO~\cite{diffusiondpo}, AlignProp~\cite{alignprop}, ReNO~\cite{reno} DiffExp~\cite{diffexp}, and DRaFT~\cite{draft}.

\section{Our Approach}
\label{sec:method}

\mypar{Preliminaries} We adopt a standard autoregressive (AR) formulation for class-conditional image generation over discrete image tokens. Let an image be represented as a sequence of tokens $x = (x_1, \dots, x_T)$ produced by a learned tokenizer, and let $c$ denote the class label. The joint distribution is factorized as
$
     p_\theta(x \,|\, c) = \prod_{t=1}^{T} p_\theta(x_t \,|\, x_{<t}, c),
$
where $x_{<t}$ denotes all previously generated tokens. The established AR model training approach \cite{vqgan, titok, llamagen} is maximizing the log-likelihood of observed sequences under this factorization, while sampling proceeds sequentially by drawing each token from $p_\theta(\cdot \,|\, x_{<t}, c)$. Following recent autoregressive image models \cite{vqgan, titok, maskgit,nextscale, llamagen}, we operate in a compressed latent space instead of pixel space. A discrete tokenizer maps each image to a shorter sequence of codebook indices, which both reduces sequence length and preserves semantic fidelity.

In this work we use LlamaGen~\cite{llamagen} family as our baseline. LlamaGen combines a learned discrete tokenizer with a Transformer-based autoregressive decoder that models $p_\theta(x_t \,|\, x_{<t}, c)$ over codebook indices. We keep this architecture and tokenizer fixed, and in the following subsections we describe how we apply policy-based fine-tuning on top of a pretrained autoregressive model using our composite reward and GRPO-style objective.

\subsection{Policy Based Fine-Tuning of AR Image Model}

We cast class-conditional generation as an MDP with state $s_t=(c,x_{<t})$, action $a_t=x_t$, and policy $\pi_\theta(a_t\mid s_t)$. After generating a full sequence $x_{1:T}$, we decode to an image and assign a scalar \emph{reward}: $r\in\mathbb{R}$. We adopt Group Relative Policy Optimization (GRPO), a PPO-style method with group-normalized advantages, which eliminates the need to learn a value function.

Given a group of $G$ samples with rewards $\{r_j\}_{j=1}^G$, advantage is defined as the normalized reward:
\[
\qquad A_j = \frac{r_j-\bar r}{s_r+\varepsilon}
\]
where $\varepsilon$ is a small positive constant, and the per-batch reward statistics are given by
\[
\bar r = \tfrac{1}{G}\sum_{j=1}^G r_j,\qquad s_r = \sqrt{\tfrac{1}{G}\sum_{j=1}^G (r_j-\bar r)^2} .
\]
Based on the sequence-level probability ratio by $\rho_j(\theta)$:
\begin{equation}
\rho_j(\theta) = \frac{\pi_\theta(x_j\mid c_j)}{\pi_{\theta_\mathrm{old}}(x_j\mid c_j)} 
    = \prod_{t=1}^{T}
      \frac{\pi_\theta(x_{j,t} \mid c, x_{j,<t})}
           {\pi_{\theta_{\text{old}}}(x_{j,t} \mid c, x_{j,<t})} ,
\label{eq:ratio_def}
\end{equation}
the GRPO objective is given by
\begin{equation}
\mathcal{L}_{\mathrm{GRPO}}(\theta) = -\frac{1}{G}\sum_{j=1}^G \min\!\Big(\rho_j(\theta)\,A_j,\;\mathrm{clip}(\rho_j(\theta),1-\epsilon,1+\epsilon)\,A_j\Big)
\label{eq:grpo-loss}
\end{equation}
where $\epsilon$ is the clipping threshold.  To induce conservative policy search, following \cite{ppo,instructGPT,grpo}, we use an approximate KL divergence \cite{kl} loss term:
\begin{equation}
D_{\mathrm{KL}}(\pi_\theta \| \pi_{\mathrm{ref}}) 
= 
\frac{\pi_{\mathrm{ref}}(x_i \mid c)}{\pi_\theta(x_i \mid c)}
- 
\log \frac{\pi_{\mathrm{ref}}(x_i \mid c)}{\pi_\theta(x_i \mid c)}
- 1.
\label{eq:kl_approx}
\end{equation}
The subsequent sections explain how we apply this framework to autoregressive image models using a combination of instance-level and distribution-level rewards, and our motivation to introduce a variance-preserving entropy term.

\subsection{Instance-level and Distribution-level Rewards}

We aim to guide the image model towards producing results that are realistic, diverse, and consistent with
the desired conditioning. For this purpose, we define a composite reward function $r_j$:

\begin{equation}
     r_j = r_j^{\mathrm{clip}} + r_j^{\mathrm{hps}} + r_j^{\mathrm{dist}}.
\end{equation}
The instance-level terms
$r_j^{\mathrm{clip}}$ (CLIPScore) and $r_j^{\mathrm{hps}}$ (HPS) encourage text/label consistency and human-preference alignment on a per-image basis. 
More specifically, the CLIPScore term uses prompts of the form ``a photo of \{class-name\}'' and scores each generated
image by the cosine similarity between the CLIP image embedding and the corresponding text
embedding, yielding a higher reward when the image semantically matches its class prompt
\cite{clip}.  Similarly, the HPS term applies the pretrained Human Preference Score v2 (HPSv2) model
\cite{hpscorev2} to the generated image and assigns its scalar output as a reward, thereby
encouraging images that align with human aesthetic and preference judgments by using the same prompts as CLIPScore.  We note that
the maximum-likelihood training optimizes token-level cross-entropy conditioned on {\em ground-truth tokens}, whereas
the per-image reward terms provide end-to-end feedback using the images autoregressively synthesized based 
on {\em generated tokens}, which allows directly targeting perceptual quality and content alignment. 

Per-image terms, however, cannot enforce that the \emph{distribution} of generated images matches the target data distribution; they may score individually plausible samples while tolerating mode collapse, poor coverage, or batch outliers. To address this, we add a \emph{distribution-level} reward $r_j^{\mathrm{dist}}$ that measures how each image contributes to aligning the {\em generated distribution} with the real one in the feature space. 
Concretely, 
let $(\mu_r,\sigma_r) \in \mathbb{R}^D \times \mathbb{R}_{\ge 0}^D$
be the pre-computed reference mean and per-dimension standard deviation, and let $(\hat\mu,\hat\sigma)$ denote estimated 
moments of the generated distribution. Using diagonal covariance for compute efficiency, the FID is
\begin{equation}
\mathrm{FID}_{\mathrm{diag}}(\mu_r,\sigma_r;\hat\mu,\hat\sigma) 
= \|\mu_r-\hat\mu\|_2^2 + \|\sigma_r-\hat\sigma\|_2^2.
\end{equation}
To attribute this distribution-level signal to each image, we define a leave-one-out (LOO) mechanism based reward. For each sample $j$ with Inception~\cite{inceptionnet} features $f_j$, we recompute moments $(\hat\mu_{-j},\hat\sigma_{-j})$ excluding $f_j$:
\begin{equation}
 r_j^{\mathrm{dist}} = \mathrm{FID}_{\mathrm{diag}}(\mu_r,\sigma_r;\hat\mu_{-j},\hat\sigma_{-j}) 
 \, - \, \mathrm{FID}_{\mathrm{diag}}(\mu_r,\sigma_r;\hat\mu,\hat\sigma).
\end{equation}
This value is \emph{positive} only if removing $j$ worsens FID, i.e., when sample $j$ helps align the generated distribution to the reference. 

A naive implementation would compute $(\hat\mu,\hat\sigma)$ from a single \emph{minibatch} only. However, such per-batch estimates depend strongly on the transient composition of the current batch, have high variance, and do not reflect the generator's \emph{running} distribution as training evolves; in practice this can tolerate mode collapse across batches or overreact to outliers. 
Instead, we maintain an {\em exponential moving average} (EMA) of generated moments that tracks the long-run generated distribution while smoothing batch noise. 
Let $(\mu^{(t)},m_2^{(t)})$ be the EMA mean and second raw moment at step $t$, and let $(\hat\mu,\widehat{m_2})$ be the current batch estimates. 
With decay $\alpha\in(0,1)$, EMA update is given by
\begin{align}
    \mu^{(t+1)} &= (1-\alpha)\,\mu^{(t)} + \alpha\,\hat\mu,\qquad \\
    m_2^{(t+1)} &= (1-\alpha)\,m_2^{(t)} + \alpha\,\widehat{m_2},
\end{align}
and $\sigma^{(t+1)}=\sqrt{\max\!\big(m_2^{(t+1)}-\mu^{(t+1)}\!\odot\!\mu^{(t+1)},\,0\big)+\varepsilon}$, where $\varepsilon$ is a small positive constant.
To attribute a per-image contribution with respect to the \emph{EMA} state, we form a hypothetical LOO update by replacing $(\hat\mu,\widehat{m_2})$ with $(\hat\mu_{-j},\widehat{m_2}_{-j})$, yielding $\mu_{-j}^{(t+1)}$ and $m_{2,-j}^{(t+1)}$. 
Let $d_j^{(t+1)} = m_{2,-j}^{(t+1)} - \mu_{-j}^{(t+1)}\!\odot\!\mu_{-j}^{(t+1)}$ and 
$\sigma_{-j}^{(t+1)} = \sqrt{\max\!\big(d_j^{(t+1)},0\big)+\varepsilon}$.
We then define the per-image distribution-level reward directly as the change in EMA-aligned FID:
\begin{align}
r_j^{\text{dist}}
&= \mathrm{FID}_{\mathrm{diag}}\!\big(\mu_r,\sigma_r;\,\mu_{-j}^{(t+1)},\sigma_{-j}^{(t+1)}\big) \\
 &- \mathrm{FID}_{\mathrm{diag}}\!\big(\mu_r,\sigma_r;\,\mu^{(t+1)},\sigma^{(t+1)}\big).
\end{align}
Thus $r_j^{\text{dist}}>0$ when omitting sample $j$ reduces EMA-aligned FID, i.e., improves distributional alignment. EMA helps to both reduce variance and tie the reward to the evolving generator distribution. In Appendix~\ref{app:loo}, we provide full details about the LOO-FID reward and a {\bf 2D distribution-matching example}, which demonstrates that LOO-FID enables stable convergence without mode collapse, mirroring our image-generation results.

\subsection{Adaptive Entropy Regularization}\label{sec:entropy}
To prevent premature collapse to low-entropy token distributions, we include an adaptive entropy bonus. 
At each position $t$, we compute the token entropy $H_t = -\sum_v p_\theta(x_t{=}v\mid s_t)\log p_\theta(x_t{=}v\mid s_t)$ from the softmax distribution over the model logits. 
Let $H_{\max}=\log K$ for vocabulary size $K$. 
We track the normalized entropy fraction $\hat H=\tfrac{1}{T}\sum_t H_t/H_{\max}$ and maintain $\hat H$ near a target $\hat H_{\text{target}}$ by adjusting the coefficient $c$.

The base schedule $c_{\text{sched}}(p)$ uses cosine decay (with warmup), based on training progress $p\!\in[0,1]$. A closed-loop correction applies when the entropy deviates beyond a deadband $\delta$:
\[
c_{\text{eff}} = \mathrm{clip}\!\big(c_{\text{sched}}(p)\, e^{\,k(\hat H_{\text{target}}-\hat H)},\, c_{\min},\, c_{\max}\big),
\]
where $k$ controls sensitivity. The final loss becomes $L(\theta) = \mathcal{L}_{\mathrm{GRPO}}(\theta) - \tfrac{c_{\text{eff}}}{T}\sum_t H_t$.
This dynamic rule increases $c_{\text{eff}}$ when entropy drops below target and reduces it when entropy is excessive, stabilizing exploration throughout training (see Appendix~\ref{app:adaptive} for more details).

\section{Experiments}
\label{sec:experiments}

\mypar{Training setup}
In all our experiments based on \textit{LlamaGen-B/L/XL} and \textit{VQGAN}, we use a $16\times$ downsampling tokenizer with $16384$ codebook entries. For \textit{VQGAN}, we use VQGAN's c-IN architecture. Each model is trained for 600 iterations over four NVIDIA A100 GPUs, taking 10--14 train hours, depending on the backbone (see Appendix~\ref{app:training}). {\bf The same hyper-parameters are used in all experiments, across the backbones.} 

\mypar{Dataset and metrics} 
We use ImageNet~\cite{imagenet} at $256\times256$ resolution for class-conditional generation. We use full 50k examples of ImageNet validation set.
We use Fréchet Inception Distance (FID)~\cite{fid}, Inception Score (IS)~\cite{inceptionscore} and Precision/Recall~\cite{precrecall} as our primary quantitative metrics. To ensure fair comparison, we rerun all pretrained baselines and compute the scores using the official TensorFlow evaluation code from \textit{LlamaGen}~\cite{llamagen}. For individual image quality and semantic consistency, we additionally report \textit{CLIPScore}~\cite{clipscore}
and a human study.

\mypar{LlamaGen results}
Table~\ref{tab:main}~\footnote{In all tables, $\downarrow$ lower is better; $\uparrow$ higher is better.} reports the main results on three different LlamaGen backbones (\textit{B/L/XL}). From the table, we observe consistent improvements in FID, IS, and CLIPScore across all backbones, together with higher precision, while recall is largely preserved (except a slight reduction for \textit{LlamaGen-B}). The results, overall, indicate improved sample quality with strong coverage.

\begin{table}[t]
\centering
\small
\caption{
Effect of our reward-based training. All experiments use CLIPScore, HPSv2, and FID rewards (weight $1.0$ each). 
Training and testing use CFG scale $1.5$. 
}
\label{tab:main}

\setlength{\tabcolsep}{2.5pt}
\resizebox{.8\linewidth}{!}{
\begin{tabular}{p{3.5cm}ccccc}
\toprule
\textbf{Model} & \textbf{FID} $\downarrow$ & \textbf{IS} $\uparrow$ & \textbf{CLIPScore} $\uparrow$ & \textbf{Precision} $\uparrow$ & \textbf{Recall} $\uparrow$\\
\midrule
LlamaGen-B         & 7.06          & 119.52          & 0.2262           & 0.74          & \textbf{0.61} \\
LlamaGen-B + ours  & \textbf{6.31} & \textbf{163.37} & \textbf{0.2347}           & \textbf{0.82}           & 0.54 \\
\midrule
LlamaGen-L         & 4.64          & 196.78          & 0.2350          & 0.78          & 0.63 \\
LlamaGen-L + ours  & \textbf{3.83} & \textbf{215.90} & \textbf{0.2360}          & \textbf{0.79}          & 0.63 \\
\midrule
LlamaGen-XL        & 3.96          & 187.75          & 0.2346          & 0.74          & 0.67 \\
LlamaGen-XL + ours & \textbf{3.82} & \textbf{195.58} & \textbf{0.2370}          & \textbf{0.77}          & 0.67 \\
\bottomrule
\end{tabular}
}
\end{table}

\begin{table}[t]
\centering
\small
\caption{
Effect of reward-based fine-tuning \emph{without} Classifier-Free Guidance (CFG).
All runs use CLIPScore, HPSv2, and FID rewards with equal weights ($1.0$ each).
Training and testing both use CFG scale $1.0$.
}
\label{tab:wo_CFG}

\setlength{\tabcolsep}{6pt}
\resizebox{.85\linewidth}{!}{
\begin{tabular}{p{3.3cm}ccccc}
\toprule
\textbf{Model} & \textbf{FID} $\downarrow$ & \textbf{IS} $\uparrow$ & \textbf{CLIPScore} $\uparrow$ & \textbf{Precision} $\uparrow$ & \textbf{Recall} $\uparrow$  \\
\midrule
LlamaGen-B         & 20.89 & 47.96  & 0.2082  & 0.47  & 0.46 \\
LlamaGen-B + ours  & \textbf{8.91}  & \textbf{112.83} & \textbf{0.2304} & \textbf{0.78} & \textbf{0.54} \\
\midrule
LlamaGen-L         & 10.24 & 82.38  & 0.2200  & 0.53  & 0.48 \\
LlamaGen-L + ours  & \textbf{5.12} & \textbf{143.32} & \textbf{0.2325} & \textbf{0.77} & \textbf{0.63} \\
\midrule
LlamaGen-XL        & 12.66 & 78.02 & 0.2192  & 0.60  & \textbf{0.74} \\
LlamaGen-XL + ours & \textbf{4.55} & \textbf{148.16} & \textbf{0.2305}& \textbf{0.74} & 0.68 \\
\bottomrule
\end{tabular}}
\end{table}

\mypar{Training and testing without CFG}
To further isolate the effect of our proposed reward-based training, we also evaluate all models without Classifier-Free Guidance (CFG).
In this setting, we disable CFG during both training and inference, setting the guidance scale to $1.0$.
This experiment directly measures the contribution of our method to the base autoregressive policy, independent of external conditioning strength.
As shown in Table~\ref{tab:wo_CFG}, our approach achieves substantial improvements across all metrics, confirming that the observed gains come from the learned policy refinement itself rather than reliance on CFG scaling. In addition, the gains in the no-CFG 
setting can be of major practical importance, since it {\bf halves the test-time AR inference steps}.

\mypar{Human Study for Quality Assessment}
Although the quantitative image generation metrics are well established and widely used, we conducted a small scale human study to support our findings. On LlamaGen-L backbone for both CFG and no-CFG cases, we ran paired A/B comparisons against the pretrained model on two tasks: {\em realism} (choose the more realistic image) and {\em reference match} (choose which image best matches 5 ImageNet-val reference images). The participants are asked to select ours / pretrained / tie to reflect their preferences in total for 260 pair per task for CFG and 570 pair per task for no-CFG; we report preferences in Table~\ref{tab:humanstudy}. Under CFG (260 pairs/task), ours is preferred more often than the pretrained baseline for both realism (117 vs.\ 102) and reference matching (135 vs.\ 94). Under no-CFG (570 pairs/task), the preference becomes substantially stronger (realism: 318 vs.\ 149; reference matching: 291 vs.\ 183). Excluding ties, two-sided binomial tests show that the preference is statistically significant for CFG reference matching (59.0\%, $p=0.008$), no-CFG realism (68.1\%,$p<10^{-14}$), and no-CFG reference matching (61.4\%, $p<10^{-6}$), while CFG realism is not as statistically significant (53.4\%, $p=0.34$). These results support that our method improves perceptual quality and semantic alignment, particularly in the no-CFG regime.

\begin{table}[t]
\centering
\caption{Human preference study using pretrained LlamaGen-L baseline versus ours (with CLIPScore, HPSv2 and LOO-FID rewards), for both CFG and no-CFG cases.\label{tab:humanstudy}}
\setlength{\tabcolsep}{6.5pt}
\resizebox{.58\linewidth}{!}{
\begin{tabular}{lcccc}
\toprule
Setting                 & Task                          & Ours                                      & Tie                              & Pretrained   \\ \hline
\multirow{2}{*}{CFG}    & Realism                       & \textbf{117}                     & 41                   & 102 \\
                        & Ref. Match  & \multicolumn{1}{c}{\textbf{135}} & \multicolumn{1}{c}{31} & 94 \\ \hline
\multirow{2}{*}{No CFG} & Realism                       & \textbf{318}                     & 103                     & 149 \\
                        & Ref. Match  & \multicolumn{1}{c}{\textbf{291}} & \multicolumn{1}{c}{96} & 183 \\
\bottomrule
\end{tabular}
}
\end{table}

\mypar{Comparison to AR-GRPO}
AR-GRPO~\cite{argrpo} is the closest related autoregressive RL method to ours. For a fair comparison, we instantiate AR-GRPO within our framework by disabling our adaptive entropy regularization and LOO-FID reward, and train using the same hyperparameters and number of iterations. The results are shown in Table~\ref{tab:argrpo}. Our method achieves substantially better FID in both settings (CLIPScore+HPSv2: 6.37$\rightarrow$\textbf{3.83}; CLIPScore+HPSv2+MANIQA: 9.34$\rightarrow$\textbf{6.43}) and consistently improves Recall (0.56$\rightarrow$\textbf{0.63}; 0.59$\rightarrow$\textbf{0.64}), indicating better distributional coverage. In the CLIPScore+HPSv2+MANIQA setting, our method also improves IS and Precision (147.30$\rightarrow$\textbf{175.34}, 0.70$\rightarrow$\textbf{0.74}). In the CLIPScore+HPSv2 setting, AR-GRPO attains slightly higher IS/Precision (223.95 vs.\ 215.90, 0.82 vs.\ 0.79), suggesting a quality--coverage trade-off while our method remains stronger on distributional fidelity and coverage.

To further verify that the gains do not arise from reduced intra-class diversity,
we report DreamSim~\cite{fu2023dreamsim} and LPIPS~\cite{zhang2018lpips}
diversity metrics using 8 samples per ImageNet class. As shown in Table~\ref{tab:diversity},
our method preserves the diversity of the pretrained model while substantially
improving FID. In contrast, AR-GRPO exhibits lower DreamSim and LPIPS scores,
together with degraded FID and Recall, indicating reduced diversity and coverage.

\begin{table}[t]
\centering
\caption{
Comparison of our approach vs ARGRPO using LlamaGen-L backbone.
Training and testing use CFG scale $1.5$. }
\label{tab:argrpo}
\resizebox{.9\linewidth}{!}{
\begin{tabular}{p{6.8cm}cccc}
\toprule
\textbf{Model} & \textbf{FID} $\downarrow$ & \textbf{IS} $\uparrow$ & \textbf{Precision} $\uparrow$ & \textbf{Recall} $\uparrow$ \\
\hline
AR-GRPO (CLIPScore + HPSv2) & 6.37 & \textbf{223.95} & \textbf{0.82} & 0.56 \\
Ours (CLIPScore + HPSv2) & \textbf{3.83} & 215.90 & 0.79 & \textbf{0.63} \\
\hline
AR-GRPO (CLIPScore + HPSv2 + MANIQA) & 9.34 & 147.30 & 0.70 & 0.59 \\
Ours (CLIPScore + HPSv2 + MANIQA) & \textbf{6.43} & \textbf{175.34} & \textbf{0.74}& \textbf{0.64} \\
\bottomrule
\end{tabular}}
\end{table}

\begin{table}[t]
\centering
\caption{Intra-class diversity evaluation using DreamSim and LPIPS on
LlamaGen-L (CFG 1.5). Higher is better for DreamSim and LPIPS.}
\label{tab:diversity}
\small
\begin{tabular}{lcccccc}
\toprule
Model & DreamSim$\uparrow$ & LPIPS$\uparrow$ & FID$\downarrow$ & IS$\uparrow$ & Prec.$\uparrow$ & Recall$\uparrow$ \\
\midrule
LlamaGen-L & \textbf{0.53} & \textbf{0.73} & 4.64 & 196.78 & 0.78 & 0.63 \\
AR-GRPO & 0.49 & 0.72 & 6.37 & 223.95 & 0.82 & 0.56 \\
Ours & \textbf{0.53} & \textbf{0.73} & \textbf{3.83} & 215.90 & 0.79 & \textbf{0.63} \\
\bottomrule
\end{tabular}
\end{table}

\mypar{Comparison to VA-$\pi$}
VA-$\pi$~\cite{liao2025vapi} is a related concurrent work that also applies RL to improve autoregressive image generation. We use the authors’ released code with LlamaGen-XL checkpoints to generate samples both with and without Classifier-Free Guidance (CFG), and evaluate all outputs using the same evaluation pipeline used throughout this paper. As shown in Table~\ref{tab:vapi}, our method consistently outperforms VA-$\pi$ in the no-CFG setting across all reported metrics (FID: 7.65$\rightarrow$4.55, IS: 102.96$\rightarrow$148.16, Precision: 0.69$\rightarrow$0.74, Recall: 0.66$\rightarrow$0.68), indicating better sample quality and stronger distributional coverage. With CFG enabled, VA-$\pi$ attains higher IS and Precision (212.60/0.80 vs.\ 195.58/0.77), but our method achieves substantially better FID and Recall (6.62$\rightarrow$3.82 and 0.58$\rightarrow$0.67), suggesting a more favorable quality--coverage trade-off and stronger distributional fidelity, highlighting the importance of the distributional reward.

\begin{table}[t]
\centering
\caption{Comparison of our approach vs VA-$\pi$~\cite{liao2025vapi} using LlamaGen-XL backbone.
If applied, CFG scale is 1.5 for each model.}
\label{tab:vapi}
\begin{tabular}{lccccc}
\hline
\textbf{Model} & \textbf{CFG} & \textbf{FID} $\downarrow$ & \textbf{IS} $\uparrow$ & \textbf{Precision} $\uparrow$ & \textbf{Recall} $\uparrow$ \\ \hline
LlamaGen-XL & --         & 12.66         & 78.02           & 0.60          & \textbf{0.74} \\
VA-$\pi$~\cite{liao2025vapi}      & --         & 7.65          & 102.96          & 0.69          & 0.66          \\
Ours        & --         & \textbf{4.55} & \textbf{148.16} & \textbf{0.74} & 0.68          \\ \hline
LlamaGen-XL & \checkmark & 3.96          & 187.75          & 0.74          & \textbf{0.67} \\
VA-$\pi$~\cite{liao2025vapi}      & \checkmark & 6.62          & \textbf{212.60} & \textbf{0.80} & 0.58          \\
Ours        & \checkmark & \textbf{3.82} & 195.58          & 0.77          & \textbf{0.67} \\ \hline
\end{tabular}
\end{table}

\mypar{Architecture generality}
To further demonstrate generalizability across architectures, we evaluate our approach on VQGAN using the same hyperparameters. Table~\ref{tab:vqgan} shows consistent improvements in FID and IS across both sampling settings (no top-$k$ and top-$k{=}100$). Precision also increases (0.57$\rightarrow$0.59 and 0.62$\rightarrow$0.69) while Recall remains stable, indicating improved sample quality without sacrificing distributional coverage.

\begin{table}[t]
\centering
\caption{
Effect of our reward-based training on VQGAN backbone. All experiments use CLIPScore, HPSv2, and FID rewards (weight $1.0$ each). 
Training and testing use no top-k sampling (top-rows) and top-k 100 (bottom rows). 
\label{tab:vqgan}}
\setlength{\tabcolsep}{8pt}
\begin{tabular}{lccccc}
\toprule
\textbf{Model} & \textbf{topk} & \textbf{FID} $\downarrow$ & \textbf{IS} $\uparrow$ & \textbf{Precision} $\uparrow$ & \textbf{Recall} $\uparrow$ \\ \hline
VQGAN          & 0             & 21.90                     & 49.81                     & 0.57                     & \textbf{0.67}                  \\
VQGAN + ours   & 0             & \textbf{15.47}            & \textbf{66.21}            & \textbf{0.59}            & 0.66                  \\ \hline
VQGAN          & 100           & 16.42                     & 78.32                     & 0.62                     & 0.57                  \\
VQGAN + ours   & 100           & \textbf{13.04}            & \textbf{97.23}            & \textbf{0.69}            & \textbf{0.58}         \\
\bottomrule
\end{tabular}
\end{table}

\mypar{Analysis of instance-level rewards}
To assess the contribution of alignment-based rewards, we examine the effects of CLIPScore and HPSv2 independently and in combination, excluding the FID reward. This experiment is conducted using the \textit{LlamaGen-L} backbone to focus on the influence of semantic alignment and perceptual quality signals. The results from Table~\ref{tab:alignmentrewards} show that using both rewards together yields the best overall performance in terms of Inception Score, confirming that CLIPScore and HPSv2 are complementary in guiding the policy toward semantically aligned and visually coherent generations.
However, the FID score degrades in the absence of the diversity-promoting FID reward, highlighting its importance for maintaining distributional coverage.

\begin{table}[t]
\centering
\small
\caption{
Ablation on alignment-based rewards using the \textit{LlamaGen-L} backbone. 
We compare CLIPScore and HPSv2 rewards individually and jointly (w/o FID reward). 
\label{tab:alignmentrewards}}
\begin{tabular}{p{4.7cm}ccrr}
\toprule
\textbf{Model} & \textbf{CLIP} & \textbf{HPS} & \textbf{FID} $\downarrow$ & \textbf{IS} $\uparrow$ \\
\midrule
LlamaGen-L (pretrained)     & --         & --         & \textbf{4.64} & 196.78 \\
LlamaGen-L + CLIP only      & \checkmark & --         & 6.00          & 217.54 \\
LlamaGen-L + HPS only       & --         & \checkmark & 7.30          & 176.73 \\
LlamaGen-L + CLIP + HPS     & \checkmark & \checkmark & 6.21          & \textbf{220.28} \\
\bottomrule
\end{tabular}
\end{table}

\begin{table}[t]
\centering
\small   
\caption{FID reward only policy training, with and without KL regularization.
\label{tab:onlyfid}}
\setlength{\tabcolsep}{6.0pt}
\begin{tabular}{lccc}
\toprule
\textbf{Model} & \textbf{KL weight $\beta$} & \textbf{FID} $\downarrow$ & \textbf{IS} $\uparrow$ \\
\midrule
LlamaGen-L (pretrained)      & --  & 4.64          & \textbf{196.78} \\
LlamaGen-L + FID-only        & 3.0 & 4.07          & 183.91          \\
LlamaGen-L + FID-only        & 0.0 & \textbf{3.97} & 188.08          \\
\bottomrule
\end{tabular}
\end{table}

\begin{table}[t]
\centering
\caption{
Diagonal vs.\ full-covariance LOO-FID with 128-D (PCA) Inception features.
All experiments use CLIPScore, HPSv2, and LOO-FID rewards (weight $1.0$ each).
Training and evaluation use CFG scale $1.5$.
\textbf{Time} reports the LOO-FID reward computation time per iteration (ms) on a single A100.
\label{tab:diagvsfull}}

\setlength{\tabcolsep}{3pt}
\resizebox{\columnwidth}{!}{
\begin{tabular}{lcccc}
\toprule
\textbf{Model} & \textbf{Feat.\ dim} & \textbf{LOO-FID time (ms)} & \textbf{FID} $\downarrow$ & \textbf{IS} $\uparrow$ \\
\midrule
Ours (Diag.\ LOO-FID, PCA)      & 128  & $\sim$30  & 4.30              & \textbf{224.16} \\
Ours (Full-cov.\ LOO-FID, PCA)  & 128  & $\sim$550 & 4.13              & 209.45 \\
Ours (Diag.\ LOO-FID, no PCA)   & 2048 & $\sim$30  & \textbf{3.83}     & 215.90 \\
\bottomrule
\end{tabular}
}
\end{table}

\mypar{Analysis of distribution-level reward}
We also investigate the impact of our distribution-level reward by isolating the FID-only signal during training. 
As shown in Table~\ref{tab:onlyfid}, even without instance-level rewards, optimizing for FID leads to measurable improvements in distributional quality compared to the pretrained baseline. In this case, we observe that KL regularization does not yield any additional improvements, possibly due to the regularizing effect of the FID-based distribution-level reward.

\begin{table}[t]
    \centering
    \begin{minipage}{0.48\textwidth}
        \centering
        \centering
\small
\caption{Effect of KL regularization on \textit{LlamaGen-L} using all rewards.
\label{tab:klweight}}
\setlength{\tabcolsep}{6pt}
\resizebox{.8\textwidth}{!}{
\begin{tabular}{lcc}
\toprule
\textbf{KL weight} $\beta$ & \textbf{FID} $\downarrow$ & \textbf{IS} $\uparrow$ \\
\midrule
3.0 & \textbf{3.83} & 215.90 \\
2.0 & 4.86          & 237.52 \\
1.0 & 5.83          & 273.69 \\
None & 10.14         & \textbf{342.91} \\
\bottomrule
\end{tabular}}

    \end{minipage}
    \hfill
    \begin{minipage}{0.48\textwidth}
        \centering
        \centering
\small
\caption{
Effect of entropy regularization coefficient on the \textit{LlamaGen-L} backbone using all rewards. 
\label{tab:entropycoef}}

\setlength{\tabcolsep}{5.0pt}
\resizebox{.8\textwidth}{!}{
\begin{tabular}{lcc}
\toprule
\textbf{Entropy coeff.} & \textbf{FID} $\downarrow$ & \textbf{IS} $\uparrow$ \\
\midrule
$4\times10^{-3}$ & \textbf{3.83} & 215.90 \\
$4\times10^{-4}$ & 4.89          & 224.24 \\
$0.0$            & 4.96          & \textbf{228.79} \\
\bottomrule
\end{tabular}}

    \end{minipage}
\end{table}

\mypar{Diagonal approximation in LOO-FID}
An important technical detail of LOO-FID estimator is the  diagonal approximation of the covariance matrix.
Table~\ref{tab:diagvsfull} compares diagonal vs.\ full-covariance LOO-FID under a controlled 128-D setting, where we apply PCA to reduce Inception features~\cite{inceptionnet} from 2048 to 128 for both methods. At 128D, full-covariance achieves a slightly lower FID (4.13 vs.\ 4.30) but a worse IS (209.45 vs.\ 224.16), while being $\sim$18$\times$ slower (550\,ms vs.\ 30\,ms).
In contrast, the diagonal approximation maintains essentially the same cost at the native 2048-D features (still $\sim$30\,ms) and enables further FID improvements (3.83). 

\mypar{Contribution of KL regularization}
We further analyze the role of KL regularization by varying its weight $\beta$ during training (using \textit{LlamaGen-L} with all rewards).
As shown in Table~\ref{tab:klweight}, stronger KL regularization (larger $\beta$) leads to lower FID scores, indicating improved distributional consistency and sample diversity.
However, this comes at the cost of reduced Inception Score, suggesting overly strong regularization constrains the policy’s ability to generate visually distinctive samples.
Conversely, weaker or no KL regularization increases IS but degrades FID, confirming the expected trade-off between fidelity and diversity.

\begin{figure*}[t]
    \centering
    \includegraphics[height=.26\textwidth, trim=.5cm 0cm 0cm 0cm, clip]{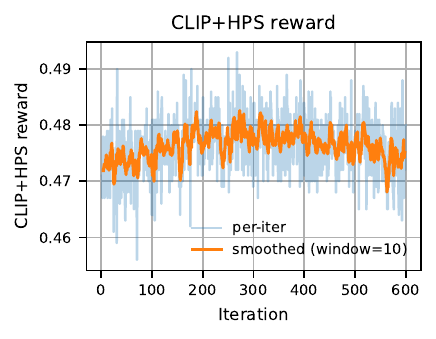}
    \hfill\includegraphics[height=.26\textwidth, trim=.55cm 0cm 0cm 0cm, clip]{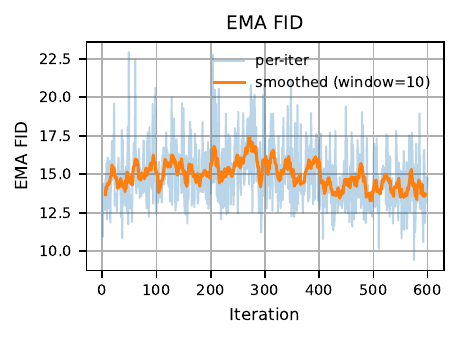}\hfill
    \includegraphics[height=.26\textwidth, trim=.55cm 0cm 0cm 0cm, clip]{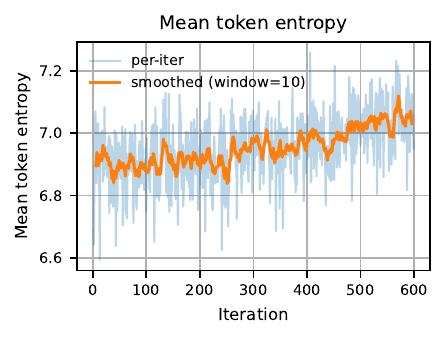}
    \caption{
        \textbf{Training dynamics on \textit{LlamaGen-L} with all rewards, KL regularization, and entropy regularization.} Solid blue lines show a moving average; light traces show per-iteration values. 
    Improvements in alignment (left) co-occur with a gradual reduction in EMA FID (middle) and a controlled rise in entropy (bottom), indicating stable learning with preserved exploration. Titles correspond to y-axis labels. \label{fig:dyn_llamagen_l}}
\end{figure*}

\mypar{Effect of entropy regularization}
We evaluate the role of entropy regularization by varying its coefficient. 
Table~\ref{tab:entropycoef} shows that stronger entropy regularization 
improves FID, suggesting better sample diversity, while lower or no entropy encourages higher IS, indicating sharper but less diverse generations.

\begin{table}[t]
\centering
\small
\caption{Comparison between policy-based post-training and maximum likelihood (MLE) post-training under comparable numbers of additional training iterations.}
\label{tab:mlecomp}

\begin{tabular}{lcc}
\toprule
\textbf{Post-training} & \textbf{FID} $\downarrow$ & \textbf{IS} $\uparrow$ \\
\midrule
LlamaGen-L baseline & 4.64 & 196.78 \\
LlamaGen-L + MLE (500 iterations) & 4.62 & 200.61 \\
LlamaGen-L + MLE (5000 iterations) & 4.54 & 202.43 \\
LlamaGen-L + ours (600 iterations) & \textbf{3.83} & \textbf{215.90} \\
\bottomrule
\end{tabular}
\end{table}

\mypar{Training dynamics of our approach}
We analyze optimization behavior on the \textit{LlamaGen-L} backbone using all rewards (CLIPScore, HPSv2, FID) with KL and entropy regularization (Table~\ref{tab:main}). 
Figure~\ref{fig:dyn_llamagen_l} summarizes three signals over 600 iterations. First, the CLIP\,+\,HPS composite reward exhibits a gradual upward trend with high-frequency noise; the smoothed trace indicates steady improvement in semantic/perceptual alignment. Second, the EMA FID fluctuates due to minibatch stochasticity but follows a slowly decreasing trajectory, consistent with the smoothing induced by the EMA of generator moments. Third, the mean token entropy increases over training, showing that the adaptive entropy bonus preserves exploration and prevents premature collapse while policy quality improves. Together, these curves indicate that our policy update simultaneously strengthens alignment (higher CLIP/HPS reward), improves distributional fidelity (lower EMA FID), and maintains sampling diversity (rising entropy).

\mypar{Comparison with MLE}
 We compare our 
policy-based approach against standard maximum likelihood estimation (MLE) fine-tuning, using \textit{LlamaGen-L} as the baseline. 
As shown in Table~\ref{tab:mlecomp}, conventional MLE training yields only 
modest
improvements even after substantially more updates: running MLE for 500 or 5000 iterations reduces FID slightly and
increases IS only marginally over the pretrained baseline. 
In contrast, our reward-based method achieves clearly better results in both metrics within only 600 training iterations, 
outperforming the best MLE result by a clear margin. These results highlight that policy optimization provides a far more sample-efficient mechanism for improving distributional quality and semantic alignment than additional autoregressive likelihood training.

\mypar{Qualitative comparison}
Figure~\ref{fig:visual_comps_cfg} and Figure~\ref{fig:visual_comps_nocfg} show samples from the pretrained and policy-trained models, with and without CFG. 
Our approach consistently yields clearer structure and more semantically faithful generations. The gains are often more dramatic in the no-CFG setting as the guidance is unable to compensate for model limitations in this case. More examples can be found in Appendix~\ref{app:qualitative}.

\providecommand{\cfgimg}[1]{\includegraphics[width=0.14\linewidth]{#1}}

\begin{figure*}[!t]
\centering

\textbf{Necklace}\\[1pt]
{\small LlamaGen-L (baseline)}\\[1pt]
\cfgimg{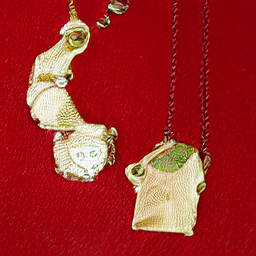}
\cfgimg{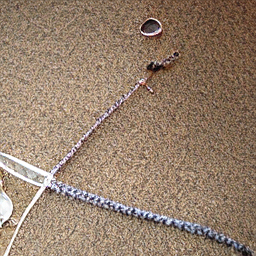}
\cfgimg{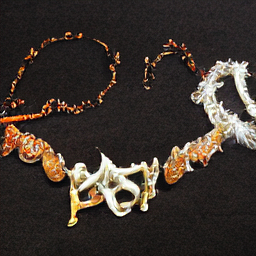}
\cfgimg{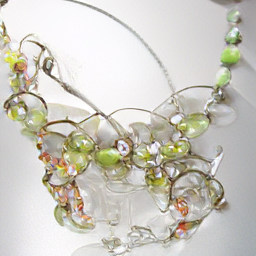}
\cfgimg{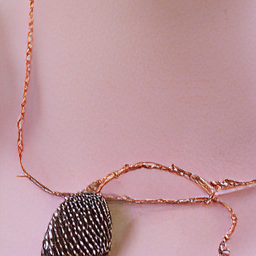}\\[1pt]

{\small LlamaGen-L + ours}\\[1pt]
\cfgimg{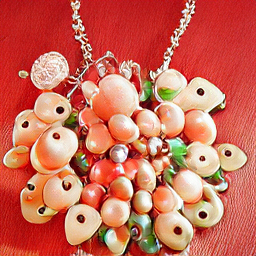}
\cfgimg{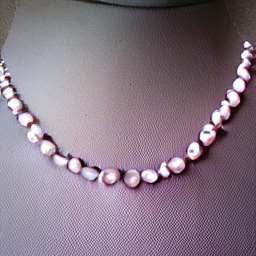}
\cfgimg{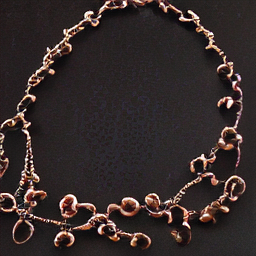}
\cfgimg{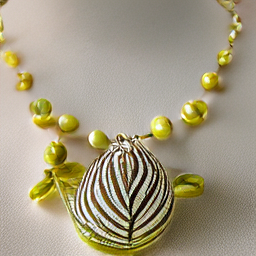}
\cfgimg{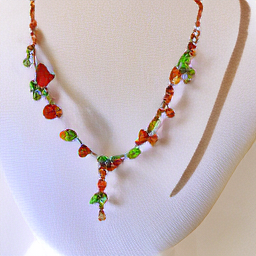}

\caption{Qualitative comparison of models with classifier-free guidance (scale=1.5). \label{fig:visual_comps_cfg} }
\end{figure*}
\providecommand{\cfgimg}[1]{\includegraphics[width=0.14\linewidth]{#1}}

\begin{figure*}[t]
\centering

\textbf{Ambulance}\\[1pt]
{\small LlamaGen-L (baseline)}\\[1pt]
\cfgimg{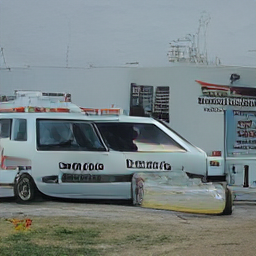}
\cfgimg{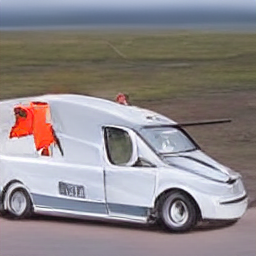}
\cfgimg{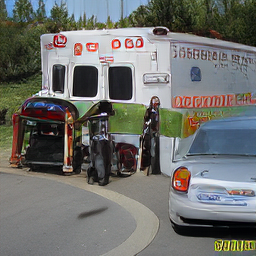}
\cfgimg{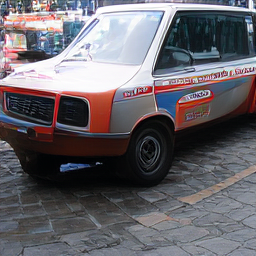}
\cfgimg{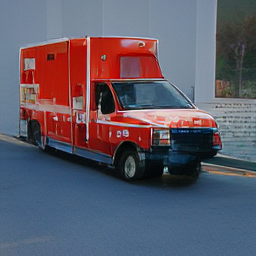}\\[1pt]

{\small LlamaGen-L + ours}\\[1pt]
\cfgimg{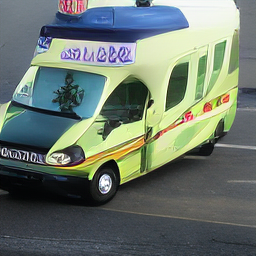}
\cfgimg{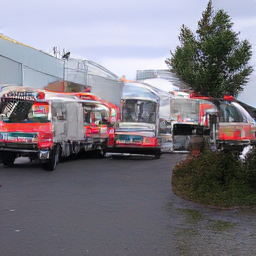}
\cfgimg{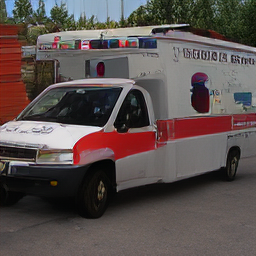}
\cfgimg{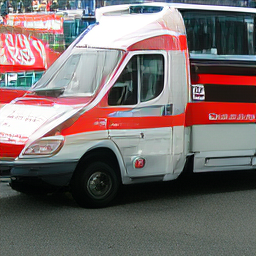}
\cfgimg{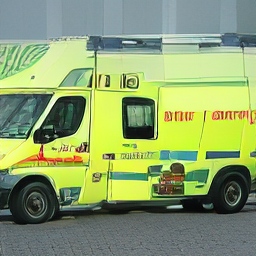} 

\caption{Qualitative comparison of models without classifier-free guidance.\label{fig:visual_comps_nocfg}}
\end{figure*}
\section{Conclusion}
\label{sec:conclusion}

In this work, we introduced a lightweight reinforcement learning framework for tuning large autoregressive image generators using a combination of instance-level and distribution-level rewards. By formulating token-based image synthesis as an MDP and optimizing the policy with GRPO, our method enables stable, value-free policy updates that directly target semantic alignment and perceptual quality. Complementary rewards--CLIPScore and HPSv2 for alignment, and a leave-one-out FID reward for distributional fidelity--guide the model toward both realistic and diverse generations, with further improvements using adaptive entropy regularization.

Across LlamaGen-B/L/XL and VQGAN backbones, using the same hyper-parameters, our approach consistently improves standard metrics with only a few hundred tuning iterations, substantially outperforming conventional MLE-based continuation. Experiments with and without classifier-free guidance further show that the improvements come from refining the underlying AR policy rather than amplifying conditioning signals. Qualitative comparisons similarly demonstrate that our method yields improvements in coherence and semantic alignment.

\paragraph{Limitations and future work.}
The framework currently relies on pretrained perceptual models for alignment rewards, and performance may vary with their biases or failure modes. Future work may explore richer distribution-level rewards, improved credit assignment strategies, integration with larger-scale AR backbones, and task-conditioned or user-in-the-loop reward formulations. Extending this policy-based tuning approach to video, 3D, or multimodal AR models also presents promising directions.

\section*{Acknowledgments}
The authors gratefully acknowledge the computational resources provided by TRUBA, ROMER, and the MareNostrum supercomputing infrastructure. This study is supported in part by the ADEP project with grant no. ADEP-312-2024-11525. G. Cinbis is supported by the ``Young Scientist Awards Program (BAGEP)'' of Science Academy, Türkiye.

\clearpage
\appendix
\setcounter{section}{0}
\renewcommand{\thesection}{\Alph{section}}
\renewcommand{\theHsection}{appendix.\Alph{section}}

\phantomsection
\section*{Appendix}
\label{app:appendix}

\section{Introduction}
\label{sec:intro-supmat}

This appendix provides additional technical details, ablation descriptions, and extended qualitative analyses that complement the main paper. We begin by presenting the full formulation of our distribution-level reward using Leave--One--Out (LOO) approach, followed by the detailed implementation of our adaptive entropy regularization mechanism and the complete GRPO training algorithm with hyperparameters used in all experiments.

In the qualitative experiments section, we compare images generated by two models: (i) the \emph{pretrained LlamaGen-L model} (baseline), and (ii) the \emph{policy-fine-tuned LlamaGen-L model} obtained using our GRPO-based training with both sample-level and distribution-level rewards. For each selected ImageNet class, we generate images using identical sampling configurations for both models and display all resulting outputs without cherry-picking. These comparisons highlight the systematic improvements brought by our approach, including sharper local structure, more coherent global geometry, and better alignment with class semantics.

\section{Details of the distribution-level reward using Leave--One--Out (LOO) estimators}
\label{app:loo}

The method section of the original paper provides an intuitive summary of our FID reward definition. The equations below provide a more detailed definition for reference. Please refer to the original paper to put these equations into context.

Let $x_i \in \mathbb{R}^D$ be the Inception \cite{inceptionnet} features for
the $i$-th generated image in a batch $\{x_i\}_{i=1}^N$. 
Let $(\mu_r,\sigma_r) \in \mathbb{R}^D \times \mathbb{R}_{\ge 0}^D$
be the reference (real) per-dimension mean and standard deviation. Due to
its cheapness we use diagonal moments only and denote elementwise product by $\odot$.

\paragraph{Batch moments.}
\begin{align}
S_1 &= \sum_{i=1}^N x_i, \quad
S_2 = \sum_{i=1}^N x_i \odot x_i, \\
\hat\mu &= \tfrac{1}{N}S_1, \quad
\widehat{m_2} = \tfrac{1}{N}S_2, \\
\hat v &= \max(\widehat{m_2} - \hat\mu \odot \hat\mu, 0), \quad
\hat\sigma = \sqrt{\hat v + \varepsilon}.
\end{align}

\paragraph{Diagonal FID.}
\begin{equation}
\mathrm{FID}_{\mathrm{diag}}
(\mu,\sigma;\mu',\sigma') =
\|\mu-\mu'\|_2^2 + \|\sigma-\sigma'\|_2^2.
\end{equation}

\paragraph{Vectorized LOO moments.}
For each $i$:
\begin{align}
S_{1,-i} &= S_1 - x_i, \quad
S_{2,-i} = S_2 - x_i \odot x_i, \\
\hat\mu_{-i} &= \tfrac{1}{N-1}S_{1,-i}, \quad
\widehat{m_2}_{-i} = \tfrac{1}{N-1}S_{2,-i}, \\
\hat v_{-i} &= \max(\widehat{m_2}_{-i} - \hat\mu_{-i}\odot\hat\mu_{-i},0), \\
\hat\sigma_{-i} &= \sqrt{\hat v_{-i} + \varepsilon}.
\end{align}

\paragraph{LOO--Batch reward.}
\begin{align}
F_{\mathrm{batch}} &=
\mathrm{FID}_{\mathrm{diag}}(\mu_r,\sigma_r;\hat\mu,\hat\sigma),\\
F_{-i} &=
\mathrm{FID}_{\mathrm{diag}}(\mu_r,\sigma_r;\hat\mu_{-i},\hat\sigma_{-i}),\\
r_i^{\text{LOO-batch}} &= F_{-i} - F_{\mathrm{batch}}.
\end{align}
Positive $r_i$ means removing $x_i$ worsens FID \cite{fid}, i.e.\ $x_i$ is helpful.

\paragraph{EMA updates.}
\begin{align}
    \tilde\mu^{(t+1)} &= (1-\alpha)\mu^{(t)} + \alpha\hat\mu,\\
    \tilde m_2^{(t+1)} &= (1-\alpha) m_2^{(t)} + \alpha\widehat{m_2}.
\end{align}

\paragraph{EMA--LOO reward.}
Let
\begin{equation}
F^{\text{full}} =
\mathrm{FID}_{\mathrm{diag}}
(\mu_r,\sigma_r;\mu^{(t+1)},\sigma^{(t+1)}).
\end{equation}

\noindent Define the LOO EMA by replacing $(\hat\mu,\widehat{m_2})$ with
$(\hat\mu_{-i},\widehat{m_2}_{-i})$, yielding $\mu_{-i}^{(t+1)}$ and $m_{2,-i}^{(t+1)}$.
For brevity, let
$d_i^{(t+1)} \!=\! m_{2,-i}^{(t+1)} - \mu_{-i}^{(t+1)}\!\odot\!\mu_{-i}^{(t+1)}$.
Then
\begin{align}
\sigma_{-i}^{(t+1)} &= \sqrt{\max\!\big(d_i^{(t+1)},0\big) + \varepsilon},\\
F_i^{\text{LOO}} &= \mathrm{FID}_{\mathrm{diag}}\!\big(\mu_r,\sigma_r;\,\mu_{-i}^{(t+1)},\sigma_{-i}^{(t+1)}\big),\\
r_i^{\text{EMA-LOO}} &= F_i^{\text{LOO}} - F^{\text{full}}.
\end{align}

This mathematical framework enables the calculation of individual rewards based on global distribution metrics, facilitating stable policy optimization. We now demonstrate the effectiveness of this approach through a simplified 2D simulation.

\subsection{2D illustration of distribution-level reward} 
\label{sec:toy2d}

\begin{figure}[t]
    \centering
    \begin{subfigure}{0.49\linewidth}
        \includegraphics[height=3.2cm]{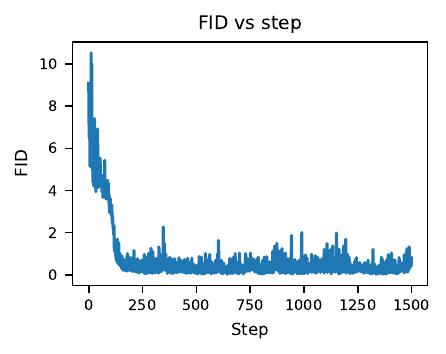}
    \end{subfigure}
    \hfill
    \begin{subfigure}{0.49\linewidth}
        \includegraphics[height=3.2cm]{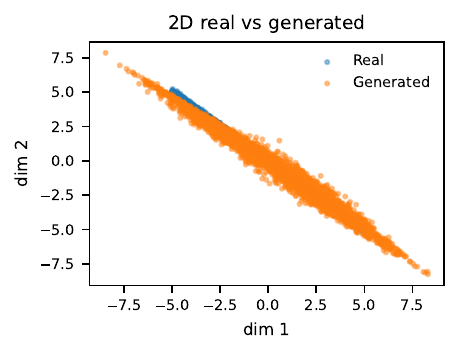}
    \end{subfigure}
    \caption{\textbf{2D example for distribution-level reward.} A simple line dataset is used to demonstrate the convergence of the policy with the proposed LOO-FID reward.}
    \label{fig:toy_loofid}
\end{figure}

To provide an intuitive view of how the leave-one-out (LOO) FID reward shapes the optimization dynamics, we simulate a simplified 2D setting where the policy learns to match a target distribution. The policy is a diagonal Gaussian $\pi_\theta(x|z)$, whose parameters are produced by a lightweight neural network. Given a latent input $z \sim \mathcal{N}(0, I)$, the network outputs the mean and log standard deviation, $\mu_\theta(z)$ and $\log\sigma_\theta(z)$, defining $\pi_\theta(x|z) = \mathcal{N}(\mu_\theta(z), \mathrm{diag}(\sigma^2_\theta(z)))$. Samples are drawn using the reparameterization trick, $x = \mu_\theta(z) + \sigma_\theta(z) \odot \epsilon$, with $\epsilon \sim \mathcal{N}(0, I)$, trained using the proposed LOO-FID reward only.
The results shown in Figure~\ref{fig:toy_loofid} for the example case using a line dataset demonstrate that the proposed EMA-based statistics estimator and the LOO-FID reward drive convergence without destabilizing variance or mode collapse, mirroring the behavior we observe in high-dimensional image generation.

\section{Entropy Regularization: Implementation Details}
\label{app:adaptive}

In the main paper (Sec.~3.3), we introduce an adaptive entropy regularization
term that prevents premature collapse of the token distribution while preserving
exploration. Here we provide the full implementation details of this term.

\mypar{Token Entropy and Normalization} Let $\pi_\theta(\cdot \mid x_{<t}, c)$ denote the autoregressive policy over a
vocabulary $\mathcal{V}$ of size $K$, with logits $z_t \in \mathbb{R}^K$ at
step $t$. At each decoding step, we obtain token probabilities empirically by
applying a softmax to these logits:
\begin{equation}
    p_{t,v} = \pi_\theta(v \mid x_{<t}, c)
    = \frac{\exp(z_{t,v})}{\sum_{u \in \mathcal{V}} \exp(z_{t,u})}.
\end{equation}
The Shannon entropy at step $t$ is
\begin{equation}
    H_t = - \sum_{v \in \mathcal{V}} p_{t,v} \log p_{t,v}.
\end{equation}
For a sequence of length $T$, we define the mean token entropy
\begin{equation}
    \bar{H} = \frac{1}{T} \sum_{t=1}^{T} H_t.
\end{equation}
The maximum entropy for a uniform distribution over the vocabulary is
\begin{equation}
    H_{\max} = \log K.
\end{equation}
We therefore define the normalized entropy fraction
\begin{equation}
    \widehat{H}
    = \frac{1}{T} \sum_{t=1}^{T} \frac{H_t}{H_{\max}}
    \in [0,1].
\end{equation}
Following the main paper, we maintain $\widehat{H}$ near a target
$\widehat{H}_{\mathrm{target}}$ by adapting the effective entropy coefficient
$c_{\mathrm{eff}}$. In practice, the quantities above are computed independently for each training
example, and the entropy term in the final loss corresponds to the average of
these per-instance mean entropies across the batch.

\mypar{Base Schedule for the Entropy Coefficient} In all of our experiments we use a single scalar coefficient
$c_{\mathrm{eff}}$ to weight the entropy bonus. Rather than keeping
this coefficient fixed, we follow a simple schedule
$c_{\mathrm{sched}}(p)$ that depends on the normalized training
progress $p \in [0,1]$ (current iteration divided by the total number
of iterations).

The schedule has three phases:

\begin{itemize}
    \item \textbf{Warmup (early training).} During the first
    $5\%$ of training we increase the coefficient linearly from
    $0$ up to a nominal value $c_0$.
    \item \textbf{Flat region (mid training).} For most of training
    ($p \in [0.05, 0.85]$) we keep the coefficient approximately
    constant at $c_0$. This encourages exploration while the policy
    is still changing rapidly.
    \item \textbf{Cosine decay (late training).} In the last $15\%$
    of training we smoothly reduce the coefficient towards a smaller
    value using a cosine decay, so that the model can focus more on
    reward optimization once the policy has stabilized.
\end{itemize}

Numerically, we set $c_0 = 2.2\times 10^{-3}$, with lower and upper
bounds $c_{\min} = 7\times 10^{-5}$ and
$c_{\max} = 4\times 10^{-3}$. The exact values are not critical; we
found that any schedule with a similar magnitude and shape (moderate
entropy pressure early on, gently reduced near the end) behaves
similarly in practice. The function $c_{\mathrm{sched}}(p)$ in our
code simply implements this warmup–flat–decay pattern.

\mypar{Adaptive Entropy Regulation} The schedule above controls the \emph{overall} scale of the entropy
bonus as training progresses. In addition, we adapt the coefficient
on-line so that the mean token entropy stays near a desired target.

Let $\widehat{H} \in [0,1]$ denote the normalized mean token entropy,
obtained by averaging the per-token entropies and dividing by the
maximum possible entropy $\log K$ (for a vocabulary of size $K$).
\[
    e = \widehat{H}_{\mathrm{target}} - \widehat{H}.
\]
When $\widehat{H}$ is below the target ($e>0$), we would like to
\emph{increase} the entropy pressure; when it is above the target
($e<0$), we would like to reduce it.

To avoid reacting to tiny fluctuations, we use a small deadband
$\delta = 0.015$: if $|e| \le \delta$ we leave the coefficient
unchanged. Outside this band we apply a simple exponential feedback
with gain $k = 3.0$, and then clamp the result to the range
$[c_{\min}, c_{\max}]$:
\begin{equation}
    c_{\mathrm{eff}}
    =
    \mathrm{clip}\!\Big(
        c_{\mathrm{sched}}(p) \,
        \exp\big( k e \big),\;
        c_{\min},\;
        c_{\max}
    \Big).
    \label{eq:ceff}
\end{equation}
Here $\mathrm{clip}(\cdot, c_{\min}, c_{\max})$ denotes scalar
clamping. Intuitively, $c_{\mathrm{eff}}$ increases when the
model becomes overly ``confident'' (entropy below the natural pretrained level) and decreases when the
distribution becomes overly diffuse (high entropy), keeping the
sampling behaviour in a reasonable range throughout training.
Unless otherwise specified, all hyperparameters associated with the entropy
regularization mechanism remain fixed across all experiments in which entropy regularization is
applied. No per-experiment tuning or adjustment of these quantities is
performed, ensuring that all reported results reflect a consistent regularization scheme, 
and we empirically observe that the method is stable and only weakly sensitive to its hyperparameters.

\mypar{Final Entropy-Regularized Objective} Let $L_{\mathrm{GRPO}}(\theta)$ denote the GRPO \cite{grpo} objective (with group-relative
advantages and PPO-style clipping) as defined in the main paper. The final
loss combines the policy objective with the entropy bonus:
\begin{equation}
    L(\theta)
    =
    L_{\mathrm{GRPO}}(\theta)
    \;-\;
    c_{\mathrm{eff}} \;
    \frac{1}{T}\sum_{t=1}^{T} H_t,
    \label{eq:entropy-final-loss}
\end{equation}
which is exactly the form stated in the main paper:
\[
    L(\theta) = L_{\mathrm{GRPO}}(\theta)
    - c_{\mathrm{eff}} \frac{1}{T} \sum_{t} H_t.
\]
The function \texttt{EntropySchedule()} in Algorithm~1 implements
Eqs.~\eqref{eq:ceff} and~\eqref{eq:entropy-final-loss}, using the numerically
specified constants above.

\section{Algorithm and Training Details}
\label{app:training}

We summarize our complete training process in Algorithm~\ref{alg:grpo-merged}.
All components described in the previous subsections (policy-based tuning,
reward modeling, and entropy regularization) are integrated
into a single GRPO iteration for policy optimization. The algorithm shows the training
loop used in all our experiments.

During training, we generate groups of 12 (G) images per iteration with 12 distinct class prompts. Unless otherwise specified, the classifier-free guidance (CFG) scale is fixed to $1.5$, top-$k$ sampling is disabled, and the top-$p$ parameter is set to $1.0$. We use a learning rate of $1\times10^{-5}$, an EMA decay $\alpha=0.5$ for batch estimates, and a clipping parameter $\epsilon=0.2$ for sequence level probability ratio $\rho_j(\theta)$. For any model we select 10\% higher of pretrained models' token entropy as target entropy $\widehat{H}_{\mathrm{target}}$. Full set of hyperparameters can be found in Table~\ref{tab:hyperparameters}.

\begin{table}[t]
\centering
\caption{Model-specific hyperparameters. Unless otherwise specified, all models
use $G=12$, $K=2$, $\beta=3.0$, $\epsilon=0.2$, $\alpha=0.5$,
reward weights
$w_{\mathrm{clip}}=w_{\mathrm{hps}}=w_{\mathrm{fid}}=1.0$,
and a tokenizer with 16,384 codebook entries.}
\label{tab:hyperparameters}

\small
\begin{tabular}{lcccccc}
\toprule
Model &
Layers &
$d_{\text{hid}}$ &
Heads &
$d_{\text{code}}$ &
$T$ &
Batch \\
\midrule
LlamaGen-B  & 12 & 768  & 12 & 8   & $16\times16$ & 144 \\
LlamaGen-L  & 24 & 1024 & 16 & 8   & $16\times16$ & 144 \\
LlamaGen-XL & 36 & 1280 & 20 & 8   & $24\times24$ & 64 \\
VQGAN       & 48 & 1024 & 16 & 256 & $16\times16$ & 64 \\
\bottomrule
\end{tabular}
\end{table}

\section{CLIP-space distribution metrics}

In addition to Inception-based FID, Precision, and Recall, we evaluate the
generated image distribution in the CLIP feature space. Since CLIP embeddings
capture semantic similarity between images and text, these metrics provide a
complementary perspective on semantic distributional quality. We report
CLIP-FID, CLIP-Precision, and CLIP-Recall under both classifier-free guidance
(CFG scale $1.5$) and without CFG in Table~\ref{tab:clip_space_metrics}. Across
all model scales, our method consistently improves CLIP-FID and
CLIP-Precision, while maintaining comparable or improved CLIP-Recall. These
results further support that the proposed policy-based tuning improves semantic
distributional alignment without sacrificing distributional coverage.

\begin{table}[t]
\centering
\caption{Distributional metrics computed in CLIP feature space. Lower is
better for CLIP-FID, while higher is better for CLIP-Precision and
CLIP-Recall.}
\label{tab:clip_space_metrics}

\renewcommand{\arraystretch}{0.95}
\setlength{\tabcolsep}{4pt}
\resizebox{0.95\columnwidth}{!}{%
\begin{tabular}{c l | c c | c c | c c}
\toprule
 &  & \multicolumn{2}{c|}{\textbf{CLIP-FID} $\downarrow$}
    & \multicolumn{2}{c|}{\textbf{CLIP-Precision} $\uparrow$}
    & \multicolumn{2}{c}{\textbf{CLIP-Recall} $\uparrow$} \\
 & \textbf{Model}
    & \textbf{Base} & \textbf{Ours}
    & \textbf{Base} & \textbf{Ours}
    & \textbf{Base} & \textbf{Ours} \\
\midrule
 & LlamaGen-B  & 6.00 & \textbf{5.12} & 0.75 & \textbf{0.80} & \textbf{0.26} & 0.22 \\
\smash{\rotatebox[origin=c]{90}{\scriptsize\textbf{CFG 1.5}}}
 & LlamaGen-L  & 4.16 & \textbf{3.80} & 0.78 & \textbf{0.79} & \textbf{0.33} & \textbf{0.33} \\
 & LlamaGen-XL & 3.55 & \textbf{3.07} & 0.78 & \textbf{0.79} & 0.47 & \textbf{0.49} \\
\midrule
 & LlamaGen-B  & 9.08 & \textbf{5.12} & 0.69 & \textbf{0.80} & \textbf{0.28} & 0.23 \\
\smash{\rotatebox[origin=c]{90}{\scriptsize\textbf{No CFG}}}
 & LlamaGen-L  & 5.99 & \textbf{3.81} & 0.72 & \textbf{0.80} & \textbf{0.37} & \textbf{0.37} \\
 & LlamaGen-XL & 5.86 & \textbf{3.39} & 0.73 & \textbf{0.80} & \textbf{0.45} & \textbf{0.45} \\
\bottomrule
\end{tabular}%
}
\end{table}

\begin{algorithm*}[t]
\small
\caption{Single GRPO Iteration (single pass)}
\label{alg:grpo-merged}
\begin{algorithmic}[1]
\Require policy $\pi_\theta$ (autoregressive), decoder $\mathcal{D}$; class tokens $c$, prompts $\mathcal{P}$
\Require length $T$; groups $G$, epochs $K$; sampling $(\tau, k, p)$; clip $\epsilon$
\Require weights $w_c,w_h$ (CLIP \cite{clipscore},HPS \cite{hpscorev2}); optional FID weight $w_{\mathrm{fid}}$; optional EMA ref $\pi_{\theta_{\mathrm{ref}}}$ with weight $\beta$
\Statex
\State \textbf{Generate and decode $G$ samples per prompt}
\For{$i=1..G$}
  \State $x_i \gets \mathrm{Generate}(\pi_\theta, c, T;\ \mathrm{CFG},\tau,k,p)$ \Comment{token sequence}
  \State $I_i \gets \mathcal{D}(x_i)$ \Comment{image}
\EndFor
\Statex
\State \textbf{Compute scalar rewards and group advantage}
\State $r^{\text{CLIP}}_i,\ r^{\text{HPS}}_i \gets \mathrm{CLIP/HPS}(I_i,\mathcal{P})$
\State $r^{\text{FID}}_i \gets \mathrm{FID\_reward}(I_i)$ \textbf{if} enabled \textbf{else} $0$
\State $r_i \gets w_c\,r^{\text{CLIP}}_i + w_h\,r^{\text{HPS}}_i + w_{\mathrm{fid}}\,r^{\text{FID}}_i$
\State $\bar{r} \gets \tfrac{1}{G}\sum_{j=1}^G r_j$, \quad $s_{r} \gets \sqrt{\tfrac{1}{G}\sum_{j=1}^G (r_j-\bar{r})^2}$
\State $A_i \gets (r_i - \bar{r})\big/(s_{r} + 10^{-4})$ \hfill 
\Statex
\State \textbf{Cache old sequence log-probs (no masks)}
\For{$i=1..G$}
  \State $\ell^{\mathrm{old}}_i[t] \gets \log \pi_{\theta_{\mathrm{old}}}(x_{i,t}\mid c,x_{i,<t})\ \ \forall t$
  \State $\bar{\ell}^{\mathrm{old}}_i \gets \tfrac{1}{T}\sum_{t=1}^{T} \ell^{\mathrm{old}}_i[t]$ \Comment{mean over tokens}
\EndFor
\Statex
\State \textbf{$K$ optimization epochs (single pass)}
\For{$e=1..K$}
  \State $\mathrm{zero\_grad}()$;\quad $\mathcal{L}\gets 0$
  \For{$i=1..G$}
    \State $\ell_i[t] \gets \log \pi_{\theta}(x_{i,t}\mid c,x_{i,<t})\ \ \forall t$
    \State $\bar{\ell}_i \gets \tfrac{1}{T}\sum_{t=1}^{T} \ell_i[t]$
    \State $\rho_i(\theta) \gets \exp\!\big(\bar{\ell}_i - \bar{\ell}^{\mathrm{old}}_i\big)$ \hfill (ratio, cf. Eq. 1 from main paper)
    \State $\tilde \rho_i \gets \mathrm{clip}\!\big(\rho_i(\theta),\, 1-\epsilon,\, 1+\epsilon\big)$
    \State $\mathcal{L}_{\mathrm{GRPO}} \gets \mathbb{E}\!\left[\min\!\big(-\rho_i(\theta)\,A_i,\ -\tilde \rho_i\,A_i\big)\right]$ \hfill (cf. Eq. 2 from main paper)
    \State $H \gets \mathrm{TokenEntropy}(x_i)$
    \State $c_{\mathrm{eff}} \gets \mathrm{EntropySchedule}()$
    \State $\mathcal{L} \mathrel{+}= \mathcal{L}_{\mathrm{GRPO}} - c_{\mathrm{eff}}\cdot \mathbb{E}[H]$
    \If{$\beta>0$ and $\pi_{\theta_{\mathrm{ref}}}$ available}
      \State $\bar{\ell}^{\mathrm{ref}}_i \gets \tfrac{1}{T}\sum_t \log \pi_{\theta_{\mathrm{ref}}}(x_{i,t}\mid c,x_{i,<t})$
      \State $\Delta_i \gets \bar{\ell}^{\mathrm{ref}}_i - \bar{\ell}_i$
      \State $\mathrm{KL}_{\mathrm{approx}} \gets e^{\Delta_i} - \Delta_i - 1$ \hfill (cf. Eq. 3 from main paper)
      \State $\mathcal{L} \mathrel{+}= \beta\cdot \mathbb{E}[\mathrm{KL}_{\mathrm{approx}}]$
    \EndIf
  \EndFor
  \State $\mathrm{backprop}(\mathcal{L})$;\quad $\mathrm{clip\_grad}(\theta,1.0)$;\quad $\mathrm{optimizer.step}()$
\EndFor
\end{algorithmic}
\end{algorithm*}

\section{Computational Costs}
We report the wall-clock time of one training iteration on a single NVIDIA A100 GPU, with all reward computations executed on-device to minimize host--device transfers and (since we use a single GPU) without network communication.
Each iteration uses a batch of 144 class-conditional prompts. Image generation for 144 samples takes $\sim$43.4\,s, and the backward pass plus parameter update takes $\sim$2.4\,s.
Reward computation adds $\sim$11.37\,s in total: CLIPScore $\sim$0.14\,s, HPS $\sim$11.20\,s, and diagonal LOO-FID $\sim$0.03\,s.
Thus, the diagonal LOO-FID component introduces negligible overhead ($\sim$0.03\,s; $<0.1\%$ of the iteration time and $\sim$0.3\% of reward time).

\section{EMA versus Batch LOO-FID}
We compare our EMA-based LOO-FID formulation against a variant that computes LOO-FID using only the current minibatch statistics. Both variants improve over the pretrained baseline; however, EMA LOO-FID consistently yields better FID and preserves Recall. This supports our hypothesis that tracking the evolving generator distribution through EMA provides a more stable and informative distribution-level reward.

\begin{table}[h]
\centering
\caption{Batch versus EMA LOO-FID.}
\label{tab:ema_batch}
\small
\begin{tabular}{lcccc}
\toprule
Model & FID$\downarrow$ & IS$\uparrow$ & Prec.$\uparrow$ & Recall$\uparrow$ \\
\midrule
LlamaGen-L & 4.64 & 196.78 & 0.78 & 0.63 \\
Batch LOO-FID & 4.06 & 214.60 & 0.80 & 0.62 \\
EMA LOO-FID (ours) & \textbf{3.83} & \textbf{215.90} & 0.79 & \textbf{0.63} \\
\bottomrule
\end{tabular}
\end{table}

\section{Adaptive Entropy Regularization}

To isolate the contribution of the adaptive entropy mechanism, we compare it against the best fixed and scheduled entropy coefficients. Adaptive entropy achieves the best FID and Recall, indicating improved preservation of
distributional coverage while maintaining competitive image quality. 

\begin{table}[h]
\centering
\caption{Adaptive versus fixed entropy regularization.}
\label{tab:adaptive_entropy}
\begin{tabular}{lcccc}
\toprule
Entropy regularization & FID$\downarrow$ & IS$\uparrow$ & Prec.$\uparrow$ & Recall$\uparrow$ \\
\midrule
Scheduled ($2\times10^{-3}$) & 4.10 & 226.66 & 0.80 & 0.61 \\
Fixed ($2\times10^{-3}$) & 4.12 & 226.39 & \textbf{0.81} & 0.61 \\
Adaptive (ours) & \textbf{3.83} & 215.90 & 0.79 & \textbf{0.63} \\
\bottomrule
\end{tabular}
\end{table}

\section{Additional Qualitative Experiments}
\label{app:qualitative}
In this section, we provide an extended set of qualitative comparisons between the pretrained LlamaGen \cite{llamagen} models and our policy fine-tuned model (LlamaGen-L only). For these qualitative comparisons, we only select the classes to visualize. For each chosen class, we report all images generated by both the pretrained and fine-tuned model under the same sampling configuration, without \textbf{cherry-picking or manual selection of individual outputs}. Across these examples, the fine-tuned policy generally produces images with sharper local structure, more consistent global geometry, and improved alignment to the semantic content of the class, reflecting the types of refinements encouraged by our reward design. Images are best viewed at high resolution, and we recommend zooming in for optimal visual inspection.

\providecommand{\cfgimg}[1]{\includegraphics[width=0.19\linewidth]{#1}}

\begin{figure*}[t]
\centering

\textbf{Monitor}\\[2pt]
{\small LlamaGen-L}\\[2pt]
\cfgimg{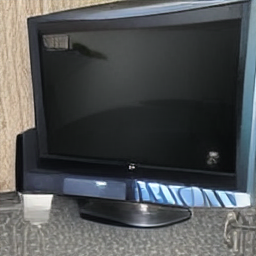}
\cfgimg{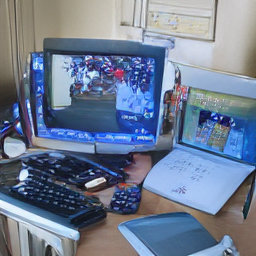}
\cfgimg{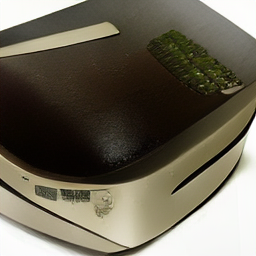}
\cfgimg{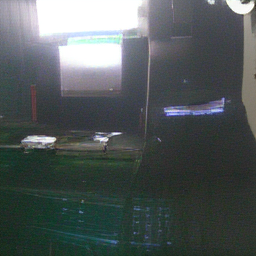}
\cfgimg{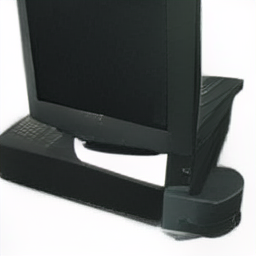}\\[4pt]

{\small LlamaGen-L + ours}\\[2pt]
\cfgimg{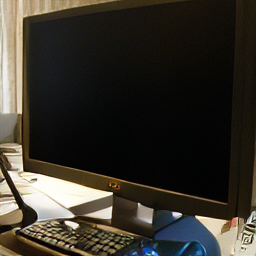}
\cfgimg{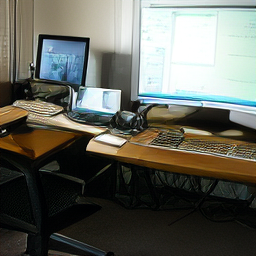}
\cfgimg{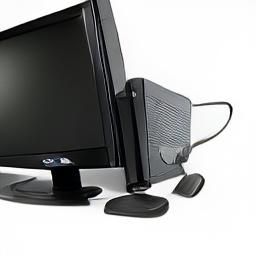}
\cfgimg{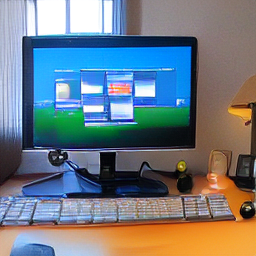}
\cfgimg{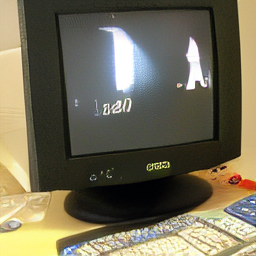}\\[12pt]

\textbf{Wine Bottle}\\[2pt]
{\small LlamaGen-L}\\[2pt]
\cfgimg{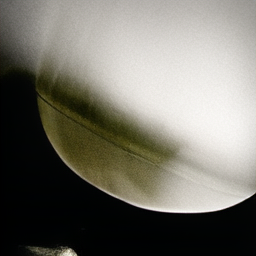}
\cfgimg{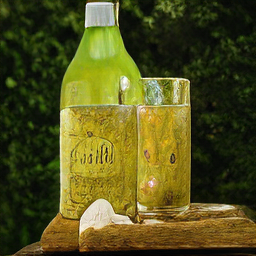}
\cfgimg{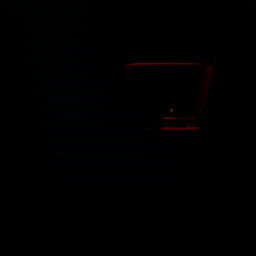}
\cfgimg{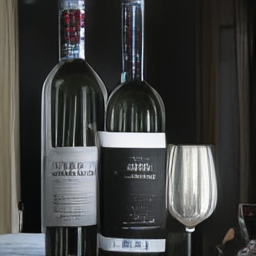}
\cfgimg{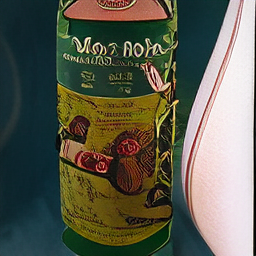}\\[4pt]

{\small LlamaGen-L + ours}\\[2pt]
\cfgimg{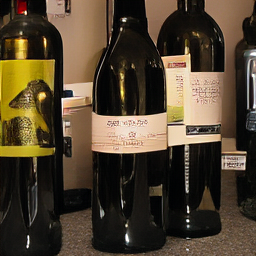}
\cfgimg{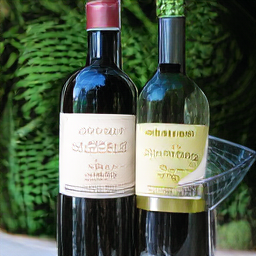}
\cfgimg{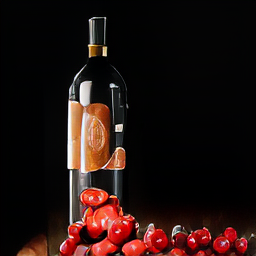}
\cfgimg{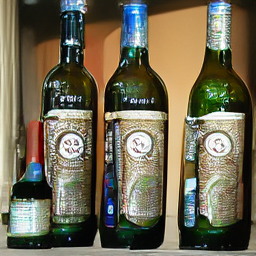}
\cfgimg{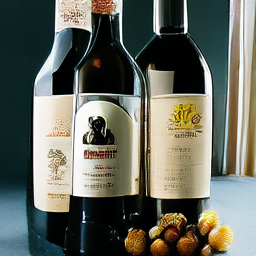}\\[4pt]

\caption{
Additional qualitative comparison between the pretrained \texttt{LlamaGen-L} and our policy fine-tuned model with classifier-free guidance on selected ImageNet classes. For both the \textbf{Monitor} and \textbf{Wine Bottle} classes, our approach produces images that more accurately reflect their respective categories. In contrast, the pretrained model fails to generate a class-consistent image for the \textbf{Monitor} class in the third example and produces an unrecognizable result for the \textbf{Wine Bottle} class in the third example.
}
\label{fig:main_to_supmat_cfg}
\end{figure*}
\providecommand{\cfgimg}[1]{\includegraphics[width=0.19\linewidth]{#1}}

\begin{figure*}[t]
\centering

\textbf{Thunder Snake}\\[2pt]
{\small LlamaGen-L}\\[2pt]
\cfgimg{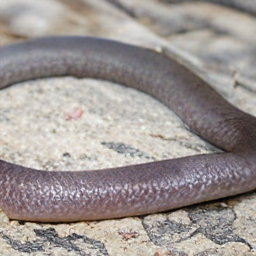}
\cfgimg{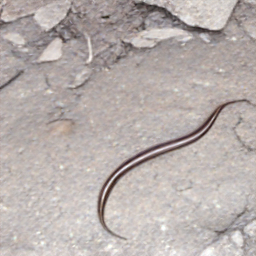}
\cfgimg{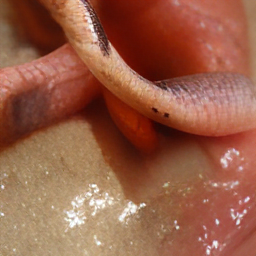}
\cfgimg{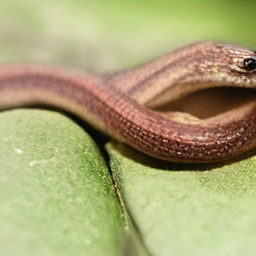}
\cfgimg{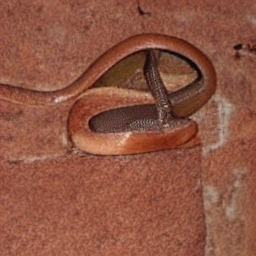}\\[4pt]

{\small LlamaGen-L + ours}\\[2pt]
\cfgimg{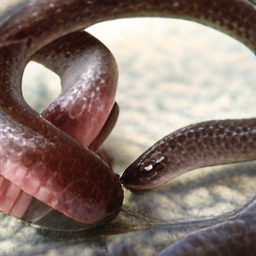}
\cfgimg{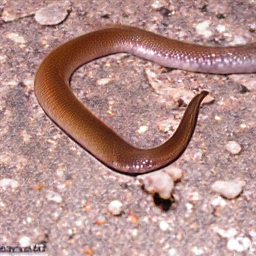}
\cfgimg{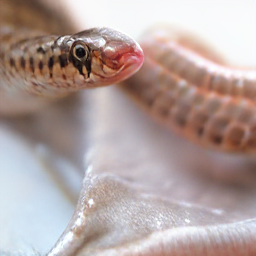}
\cfgimg{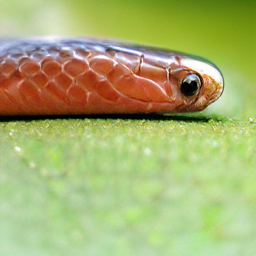}
\cfgimg{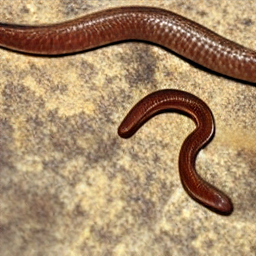}\\[12pt]

\textbf{Goldfish}\\[2pt]
{\small LlamaGen-L}\\[2pt]
\cfgimg{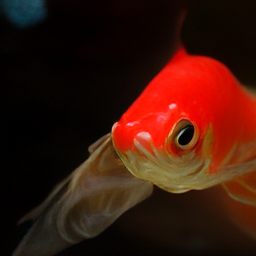}
\cfgimg{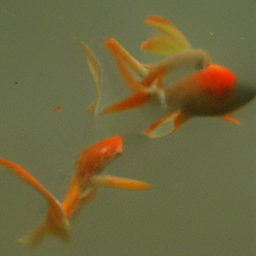}
\cfgimg{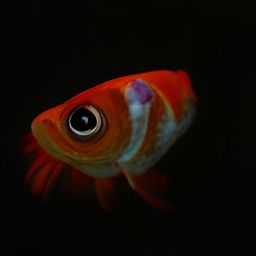}
\cfgimg{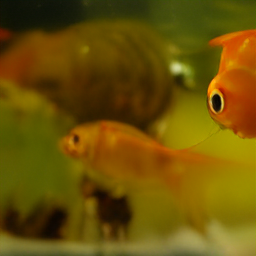}
\cfgimg{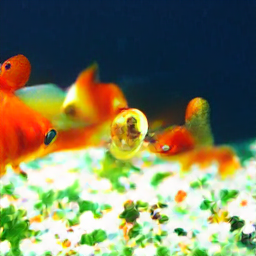}\\[4pt]

{\small LlamaGen-L + ours}\\[2pt]
\cfgimg{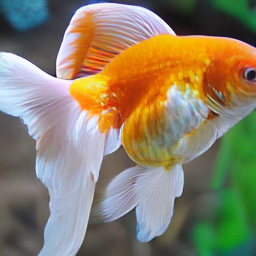}
\cfgimg{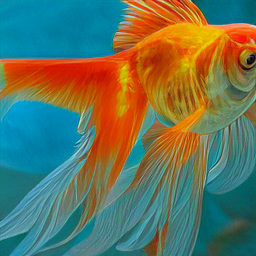}
\cfgimg{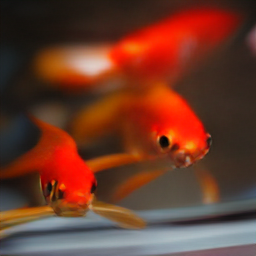}
\cfgimg{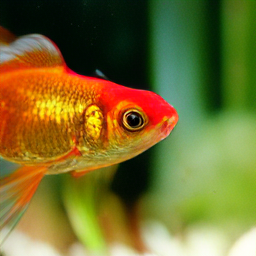}
\cfgimg{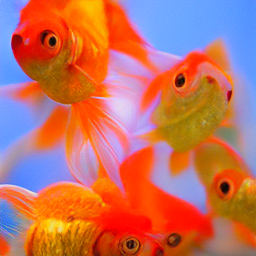}\\[4pt]

\caption{
Additional qualitative comparison between pretrained \texttt{LlamaGen-L} and our policy fine-tuned model with
classifier-free guidance. \textbf{Thunder snake:} our fine-tuned model produces images that convey the class-specific semantics more clearly, offering more descriptive and recognizable visual attributes compared to the pretrained model. \textbf{Goldfish:} pretrained model often exhibits inconsistencies in body structure and overall geometry, whereas our fine-tuned model generates shapes that are more coherent and anatomically plausible.\\
}
\label{fig:cfg_page_1}
\end{figure*}

\providecommand{\cfgimg}[1]{\includegraphics[width=0.19\linewidth]{#1}}

\begin{figure*}[t]
\centering

\textbf{Hen}\\[2pt]
{\small LlamaGen-L}\\[2pt]
\cfgimg{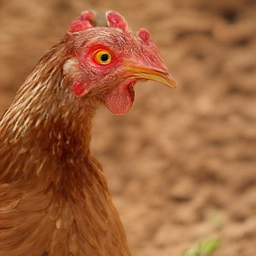}
\cfgimg{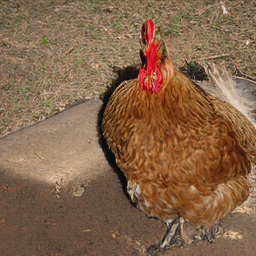}
\cfgimg{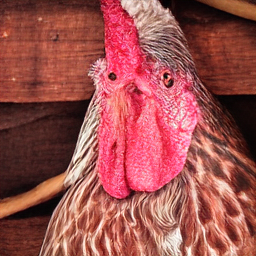}
\cfgimg{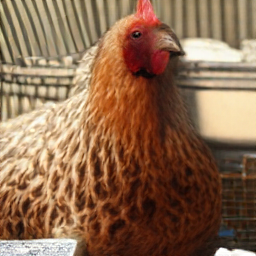}
\cfgimg{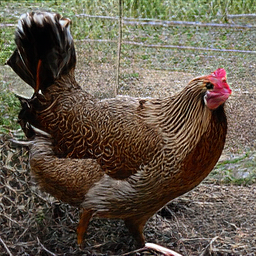}\\[4pt]

{\small LlamaGen-L + ours}\\[2pt]
\cfgimg{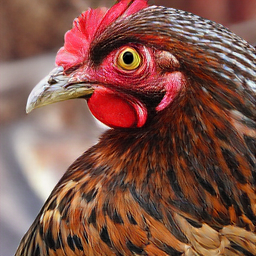}
\cfgimg{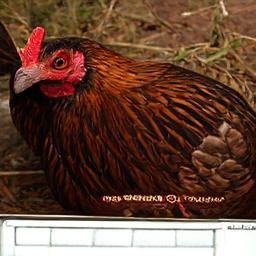}
\cfgimg{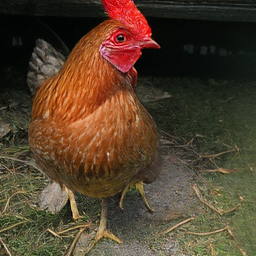}
\cfgimg{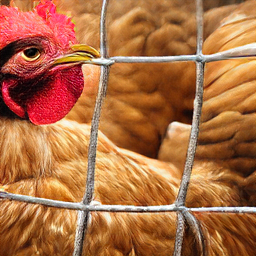}
\cfgimg{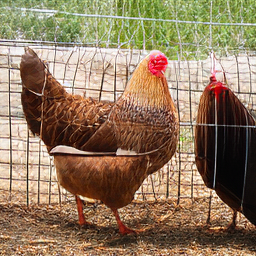}\\[12pt]

\textbf{Ostrich}\\[2pt]
{\small LlamaGen-L}\\[2pt]
\cfgimg{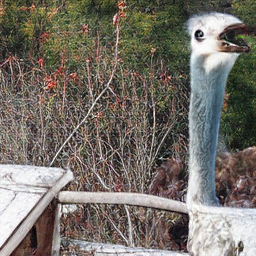}
\cfgimg{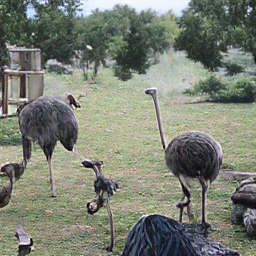}
\cfgimg{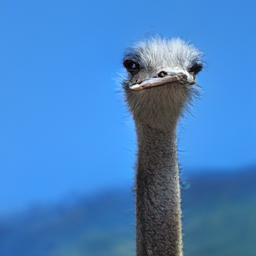}
\cfgimg{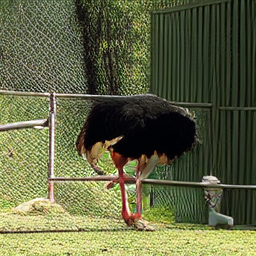}
\cfgimg{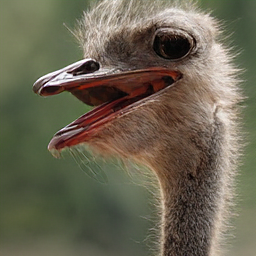}\\[4pt]

{\small LlamaGen-L + ours}\\[2pt]
\cfgimg{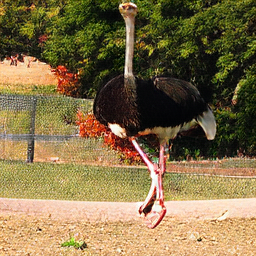}
\cfgimg{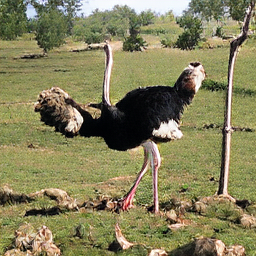}
\cfgimg{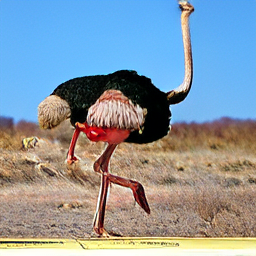}
\cfgimg{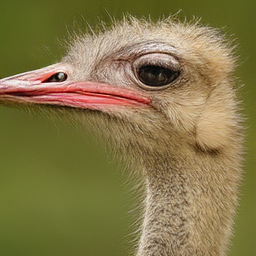}
\cfgimg{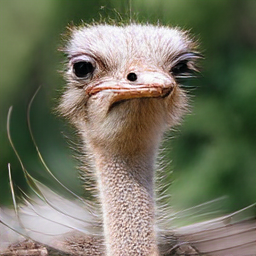}\\[4pt]

\caption{
Additional qualitative comparison between pretrained
\texttt{LlamaGen-L} and our policy fine-tuned model with
classifier-free guidance. \textbf{Hen:} Our model produces faces with clearer structure and more coherent anatomical details. \textbf{Ostrich:} The overall body geometry and structural consistency are improved in our results.\\
}
\label{fig:cfg_page_2}
\end{figure*}

\providecommand{\cfgimg}[1]{\includegraphics[width=0.19\linewidth]{#1}}

\begin{figure*}[t]
\centering

\textbf{Scorpion}\\[2pt]
{\small LlamaGen-L}\\[2pt]
\cfgimg{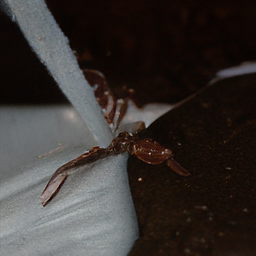}
\cfgimg{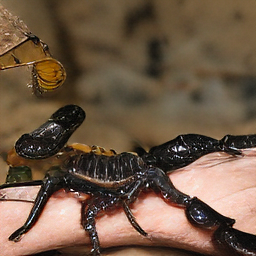}
\cfgimg{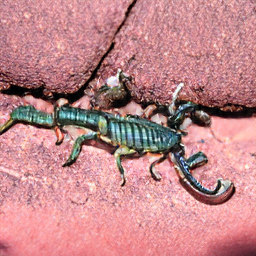}
\cfgimg{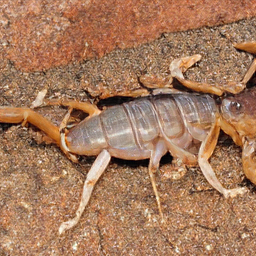}
\cfgimg{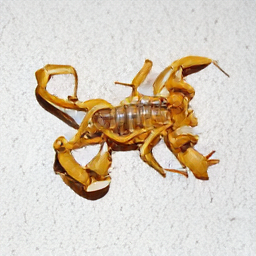}\\[4pt]

{\small LlamaGen-L + ours}\\[2pt]
\cfgimg{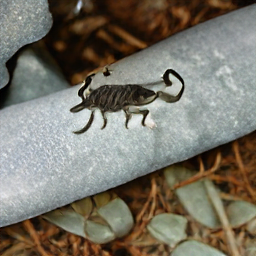}
\cfgimg{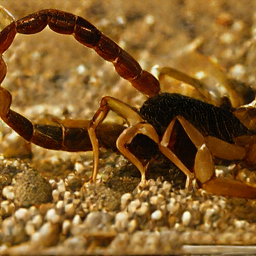}
\cfgimg{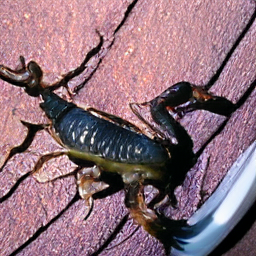}
\cfgimg{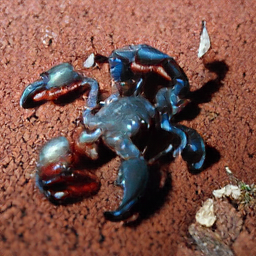}
\cfgimg{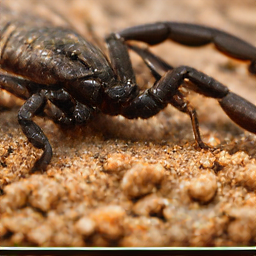}\\[12pt]

\textbf{Kite}\\[2pt]
{\small LlamaGen-L}\\[2pt]
\cfgimg{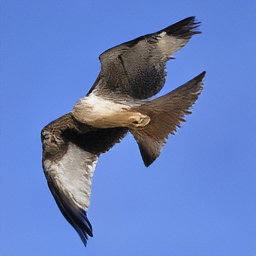}
\cfgimg{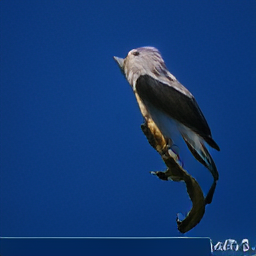}
\cfgimg{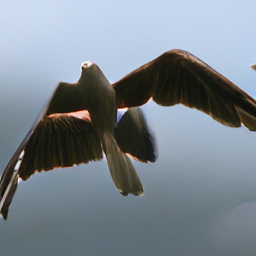}
\cfgimg{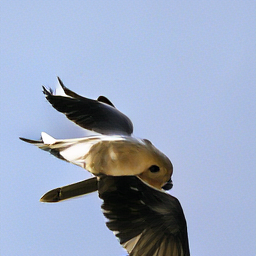}
\cfgimg{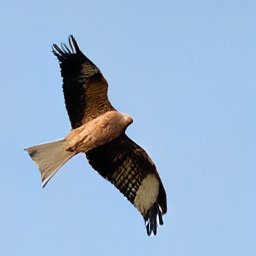}\\[4pt]

{\small LlamaGen-L + ours}\\[2pt]
\cfgimg{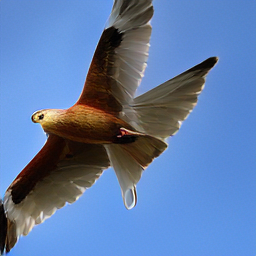}
\cfgimg{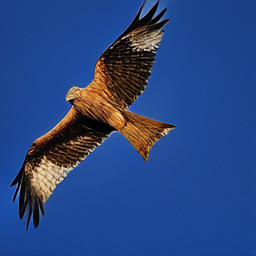}
\cfgimg{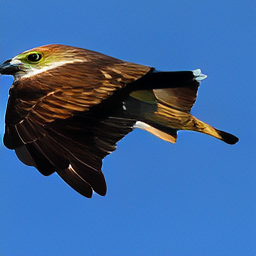}
\cfgimg{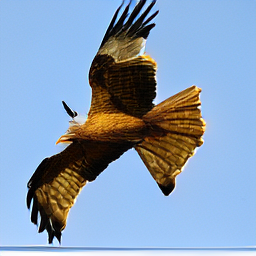}
\cfgimg{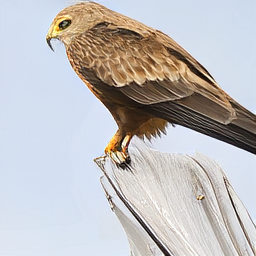}\\[4pt]

\caption{
Additional qualitative comparison between pretrained
\texttt{LlamaGen-L} and our policy fine-tuned model with
classifier-free guidance. \textbf{Scorpion:} Both models produce images of comparable visual quality for this class. 
\textbf{Kite:} Our model generates images that more faithfully capture the characteristic visual features of the class, whereas the pretrained model often displays noticeable geometric distortions.\\
}
\label{fig:cfg_page_3}
\end{figure*}

\providecommand{\cfgimg}[1]{\includegraphics[width=0.19\linewidth]{#1}}

\begin{figure*}[t]
\centering

\textbf{Bald Eagle}\\[2pt]
{\small LlamaGen-L}\\[2pt]
\cfgimg{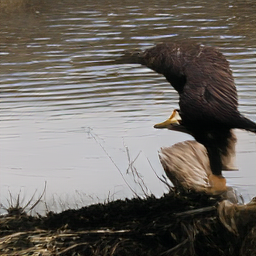}
\cfgimg{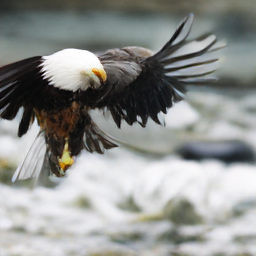}
\cfgimg{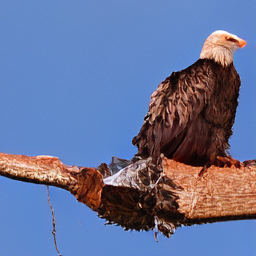}
\cfgimg{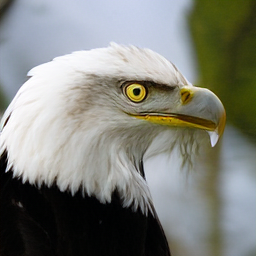}
\cfgimg{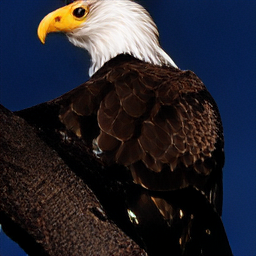}\\[4pt]

{\small LlamaGen-L + ours}\\[2pt]
\cfgimg{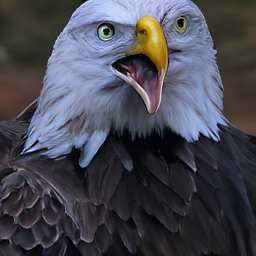}
\cfgimg{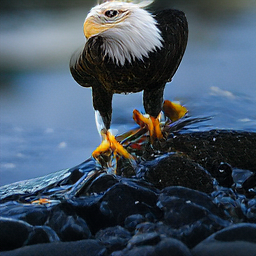}
\cfgimg{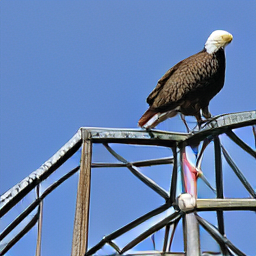}
\cfgimg{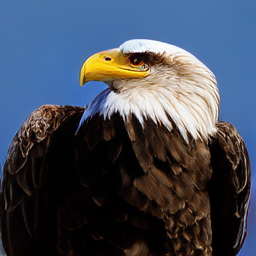}
\cfgimg{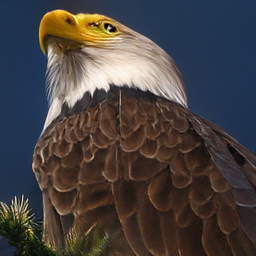}\\[12pt]

\textbf{Vulture}\\[2pt]
{\small LlamaGen-L}\\[2pt]
\cfgimg{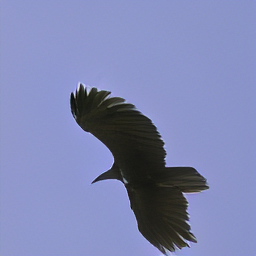}
\cfgimg{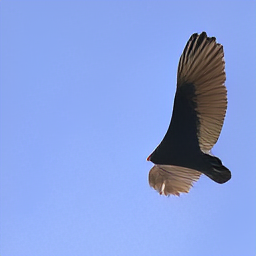}
\cfgimg{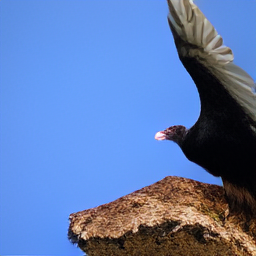}
\cfgimg{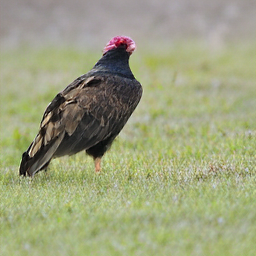}
\cfgimg{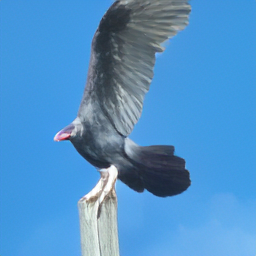}\\[4pt]

{\small LlamaGen-L + ours}\\[2pt]
\cfgimg{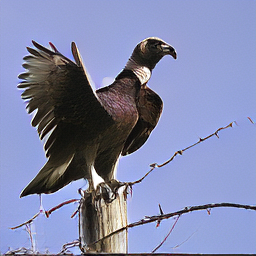}
\cfgimg{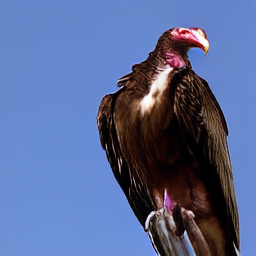}
\cfgimg{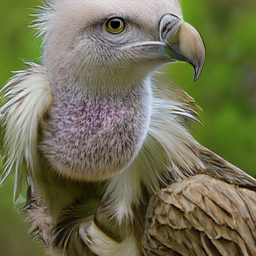}
\cfgimg{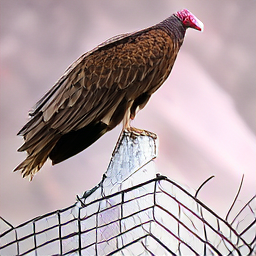}
\cfgimg{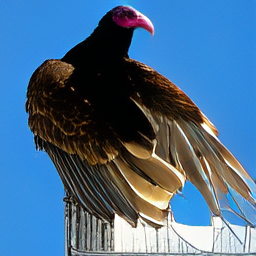}\\[4pt]

\caption{
Additional qualitative comparison between pretrained
\texttt{LlamaGen-L} and our policy fine-tuned model with
classifier-free guidance. \textbf{Bald Eagle:} Our generations exhibit higher aesthetic quality and more coherent detail overall. 
\textbf{Vulture:} The pretrained model often produces images that are difficult to associate with the target class and appear visually generic, whereas our model yields outputs that are more clearly identifiable and better aligned with the characteristic features of the class.\\
}
\label{fig:cfg_page_4}
\end{figure*}

\providecommand{\cfgimg}[1]{\includegraphics[width=0.19\linewidth]{#1}}

\begin{figure*}[t]
\centering

\textbf{Jellyfish}\\[2pt]
{\small LlamaGen-L}\\[2pt]
\cfgimg{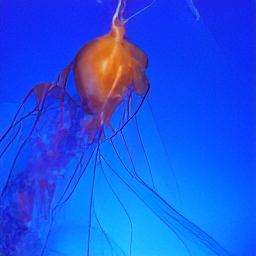}
\cfgimg{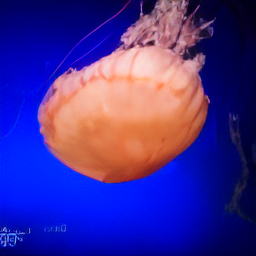}
\cfgimg{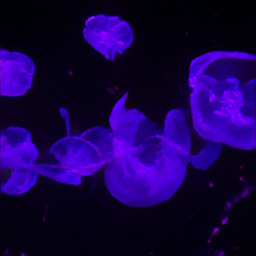}
\cfgimg{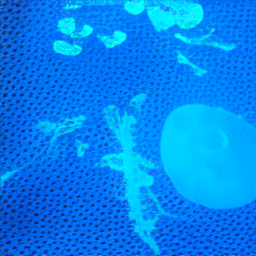}
\cfgimg{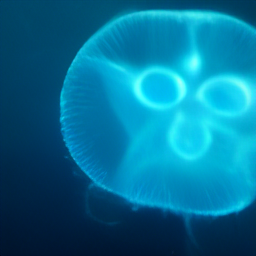}\\[4pt]

{\small LlamaGen-L + ours}\\[2pt]
\cfgimg{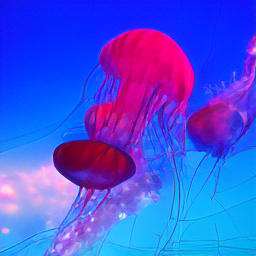}
\cfgimg{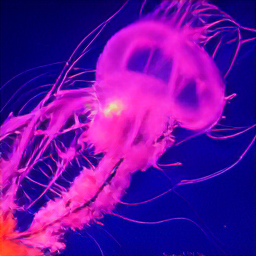}
\cfgimg{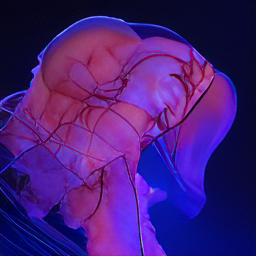}
\cfgimg{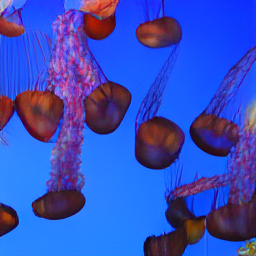}
\cfgimg{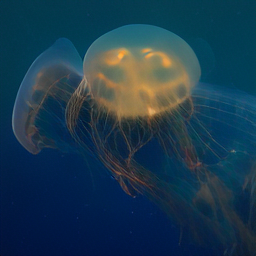}\\[12pt]

\textbf{Spotted Salamander}\\[2pt]
{\small LlamaGen-L}\\[2pt]
\cfgimg{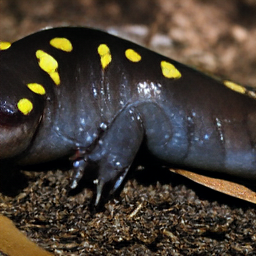}
\cfgimg{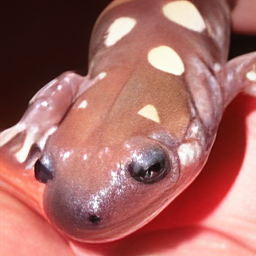}
\cfgimg{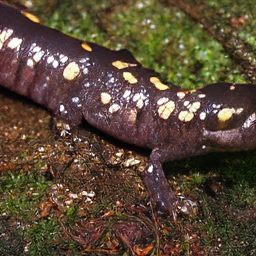}
\cfgimg{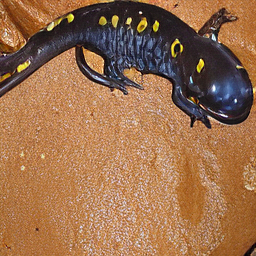}
\cfgimg{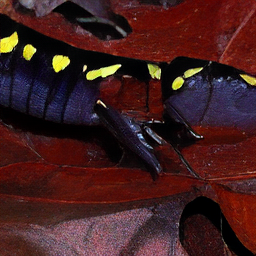}\\[4pt]

{\small LlamaGen-L + ours}\\[2pt]
\cfgimg{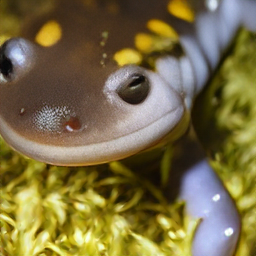}
\cfgimg{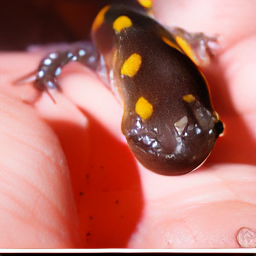}
\cfgimg{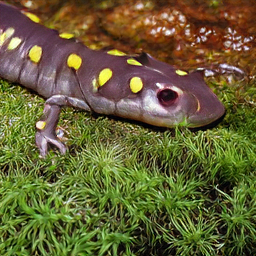}
\cfgimg{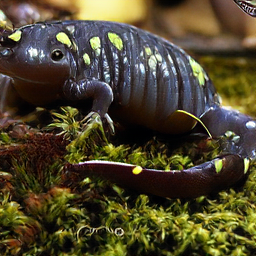}
\cfgimg{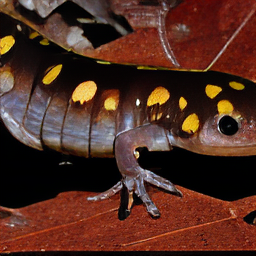}\\[4pt]

\caption{
Additional qualitative comparison between pretrained
\texttt{LlamaGen-L} and our policy fine-tuned model with
classifier-free guidance. \textbf{Jellyfish:} The pretrained model often produces shapes that lack clear structure and show irregular forms. Our model generates jellyfish with more coherent body geometry, smoother outlines, and tentacles that better reflect the characteristic of the class. \textbf{Spotted Salamander:} The pretrained model sometimes produces body shapes that lack proper limb proportion. Our model generally yields more anatomically coherent salamanders with clearer limb structure, resulting in images that better match the class’ defining visual attributes.
\\
}
\label{fig:cfg_comps_5}
\end{figure*}

\providecommand{\cfgimg}[1]{\includegraphics[width=0.19\linewidth]{#1}}

\begin{figure*}[t]
\centering

\textbf{Beagle}\\[2pt]
{\small LlamaGen-L}\\[2pt]
\cfgimg{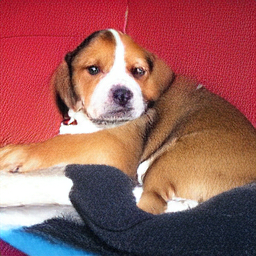}
\cfgimg{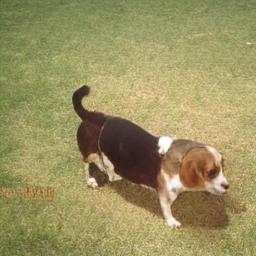}
\cfgimg{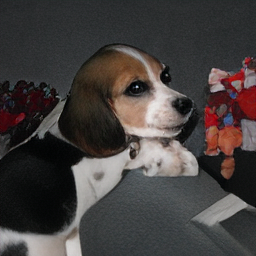}
\cfgimg{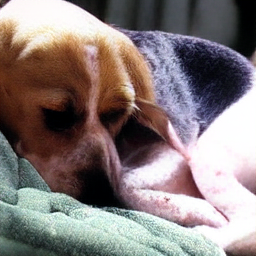}
\cfgimg{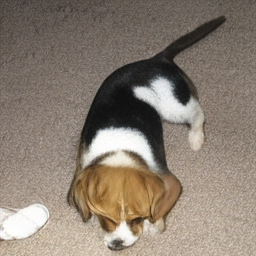}\\[4pt]

{\small LlamaGen-L + ours}\\[2pt]
\cfgimg{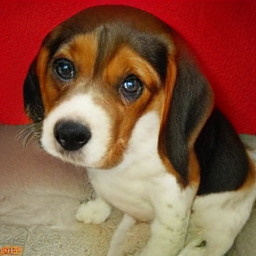}
\cfgimg{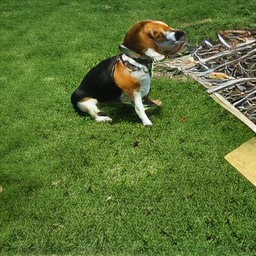}
\cfgimg{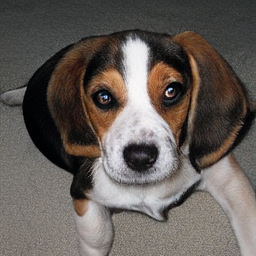}
\cfgimg{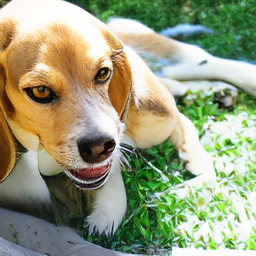}
\cfgimg{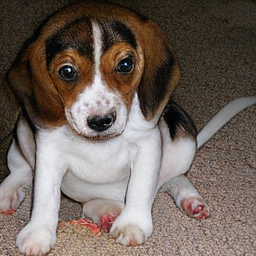}\\[12pt]

\textbf{Norwegian Elkhound}\\[2pt]
{\small LlamaGen-L}\\[2pt]
\cfgimg{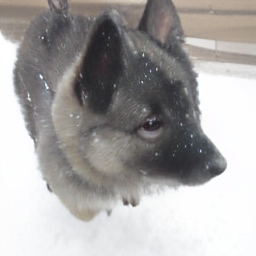}
\cfgimg{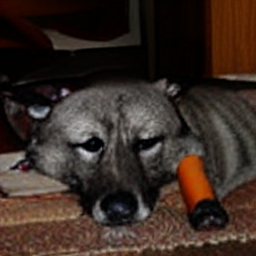}
\cfgimg{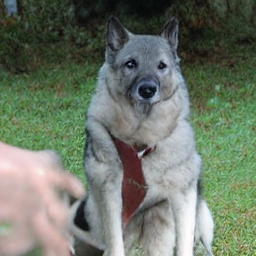}
\cfgimg{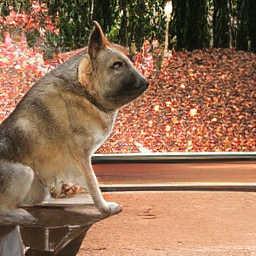}
\cfgimg{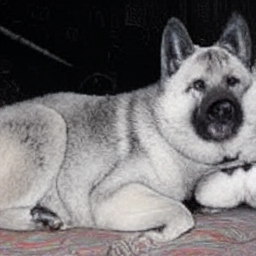}\\[4pt]

{\small LlamaGen-L + ours}\\[2pt]
\cfgimg{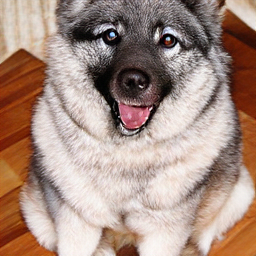}
\cfgimg{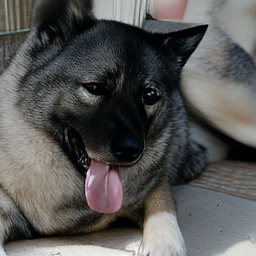}
\cfgimg{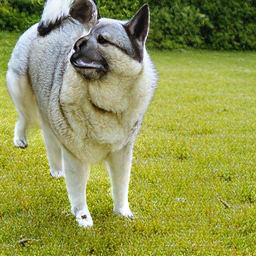}
\cfgimg{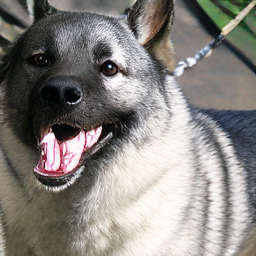}
\cfgimg{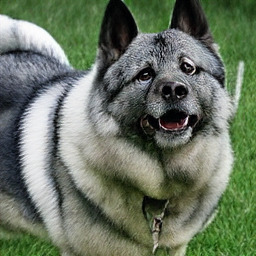}\\[4pt]

\caption{
Additional qualitative comparison between pretrained
\texttt{LlamaGen-L} and our policy fine-tuned model with
classifier-free guidance. \textbf{Beagle:} The pretrained model generally captures the overall dog shape but often produces faces with blurred or softened features. Our model yields clearer facial definition. Overall, the fine-tuned model provides more recognizable, aesthetically better and visually coherent beagle depictions. \textbf{Norwegian Elkhound:} The pretrained model often produces dogs with somewhat blurred facial regions, and the overall body shape can appear loosely defined. Our model generates images with clearer facial structure, resulting in more recognizable and visually consistent elkhound depictions.
\\
}
\label{fig:cfg_comps_6}
\end{figure*}

\providecommand{\cfgimg}[1]{\includegraphics[width=0.19\linewidth]{#1}}

\begin{figure*}[t]
\centering

\textbf{Golden Retriever}\\[2pt]
{\small LlamaGen-L}\\[2pt]
\cfgimg{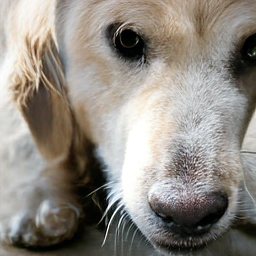}
\cfgimg{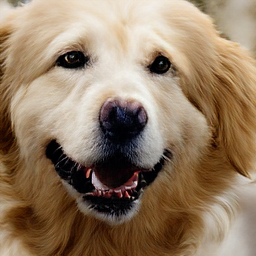}
\cfgimg{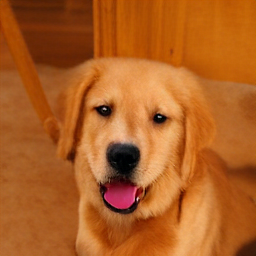}
\cfgimg{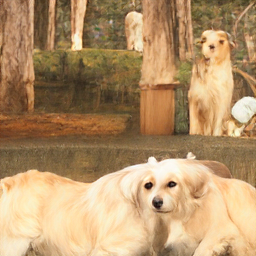}
\cfgimg{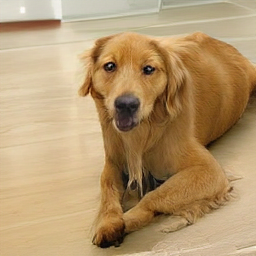}\\[4pt]

{\small LlamaGen-L + ours}\\[2pt]
\cfgimg{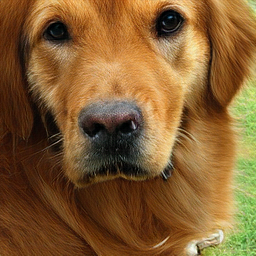}
\cfgimg{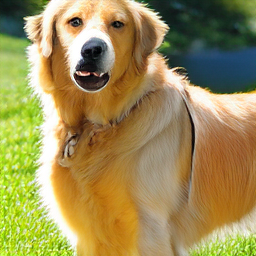}
\cfgimg{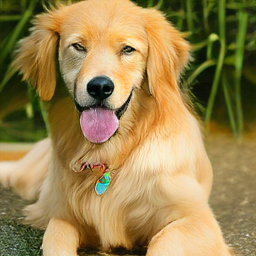}
\cfgimg{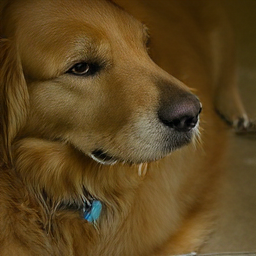}
\cfgimg{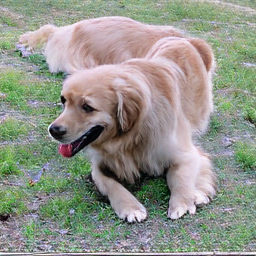}\\[12pt]

\textbf{Bee}\\[2pt]
{\small LlamaGen-L}\\[2pt]
\cfgimg{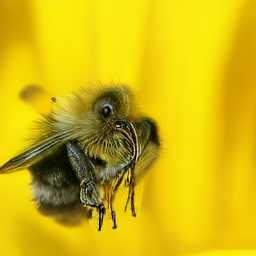}
\cfgimg{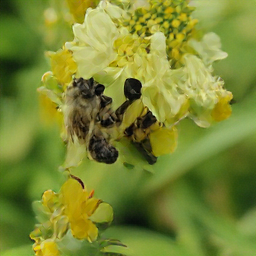}
\cfgimg{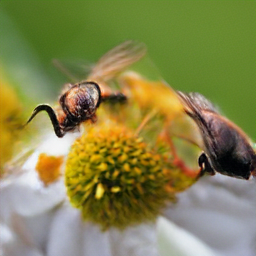}
\cfgimg{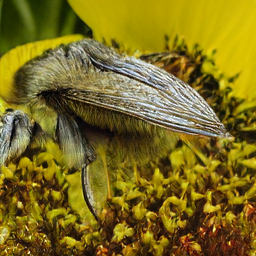}
\cfgimg{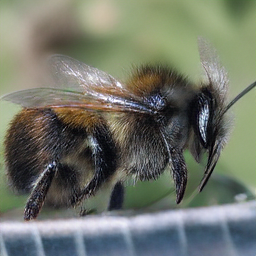}\\[4pt]

{\small LlamaGen-L + ours}\\[2pt]
\cfgimg{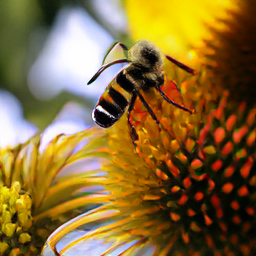}
\cfgimg{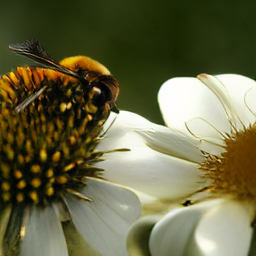}
\cfgimg{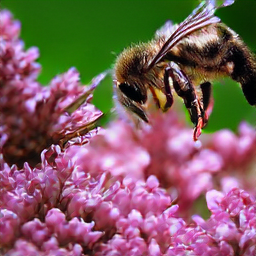}
\cfgimg{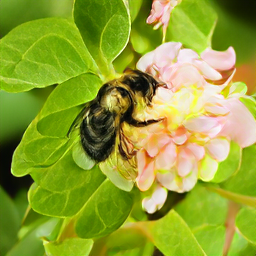}
\cfgimg{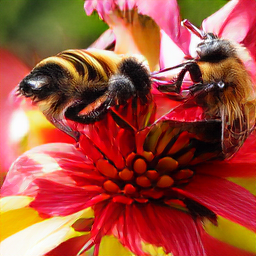}\\[4pt]

\caption{
Additional qualitative comparison between pretrained
\texttt{LlamaGen-L} and our policy fine-tuned model with
classifier-free guidance. \textbf{Golden Retriever:} The pretrained model generally captures the overall dog shape but often produces facial regions that appear soft or slightly distorted, and the fur texture lacks the characteristic smooth, layered appearance of the breed. Our model produces dogs with clearer facial definition, more coherent head and ear geometry, and fur patterns that appear more natural, giving the resulting images a more recognizable and visually consistent golden retriever appearance. \textbf{Bee:} The pretrained model often produces bodies and wings with irregular geometry. Our model yields bees with more coherent body structure, better-defined wing shapes, and markings that more closely resemble the expected striped pattern, resulting in images that are visually clearer and more class-consistent.
\\
}
\label{fig:cfg_comps_7}
\end{figure*}

\providecommand{\cfgimg}[1]{\includegraphics[width=0.19\linewidth]{#1}}

\begin{figure*}[t]
\centering

\textbf{Starfish}\\[2pt]
{\small LlamaGen-L}\\[2pt]
\cfgimg{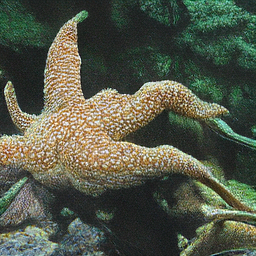}
\cfgimg{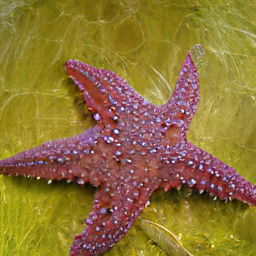}
\cfgimg{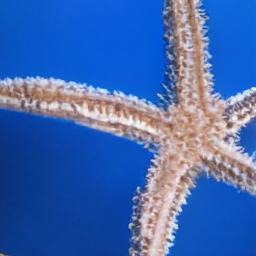}
\cfgimg{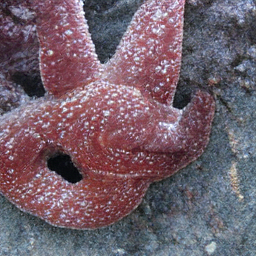}
\cfgimg{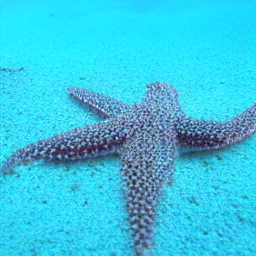}\\[4pt]

{\small LlamaGen-L + ours}\\[2pt]
\cfgimg{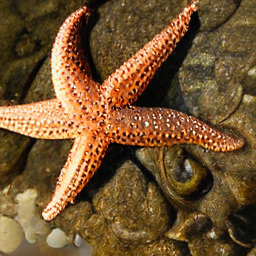}
\cfgimg{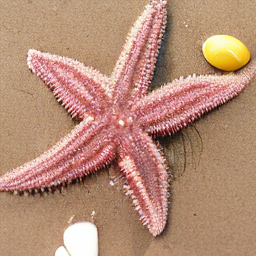}
\cfgimg{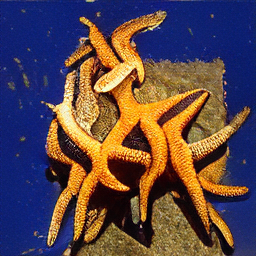}
\cfgimg{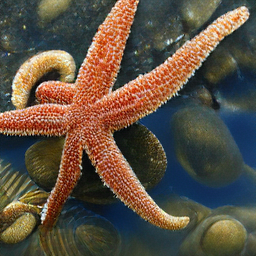}
\cfgimg{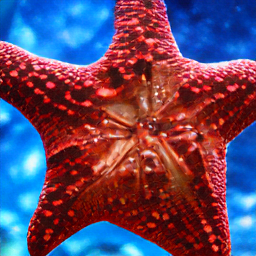}\\[12pt]

\textbf{Hare}\\[2pt]
{\small LlamaGen-L}\\[2pt]
\cfgimg{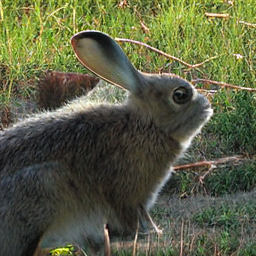}
\cfgimg{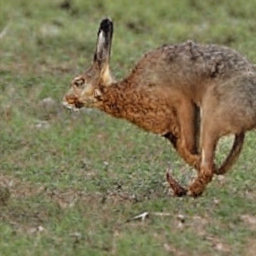}
\cfgimg{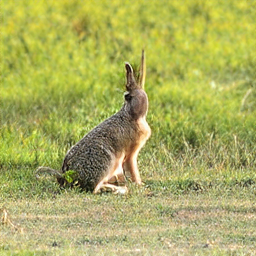}
\cfgimg{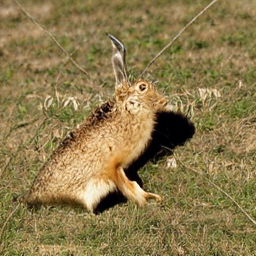}
\cfgimg{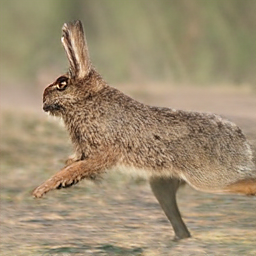}\\[4pt]

{\small LlamaGen-L + ours}\\[2pt]
\cfgimg{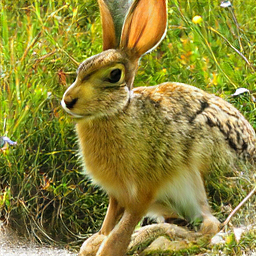}
\cfgimg{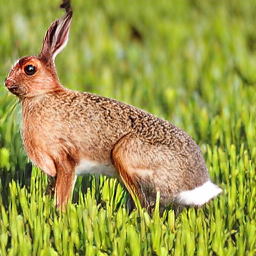}
\cfgimg{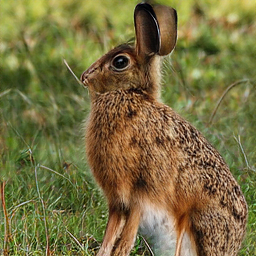}
\cfgimg{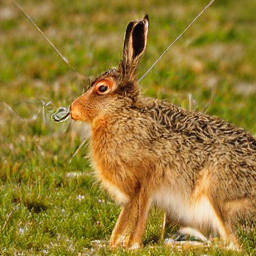}
\cfgimg{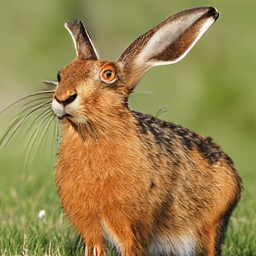}\\[4pt]

\caption{
Additional qualitative comparison between pretrained
\texttt{LlamaGen-L} and our policy fine-tuned model with
classifier-free guidance. \textbf{Starfish:} The pretrained model sometimes produces starfish with uneven arm shapes or irregular overall geometry. Our model generates starfish with more consistent five-arm structure, clearer separation between the arms, resulting in images that better reflect the expected morphology of the class. \textbf{Hare:} Both models produce plausible hare images, but the pretrained outputs often appear softer and less structured. Our generations show clearer facial and ear definition and slightly improved overall aesthetic coherence.
\\
}
\label{fig:cfg_comps_8}
\end{figure*}

\providecommand{\cfgimg}[1]{\includegraphics[width=0.19\linewidth]{#1}}

\begin{figure*}[t]
\centering

\textbf{Backpack}\\[2pt]
{\small LlamaGen-L}\\[2pt]
\cfgimg{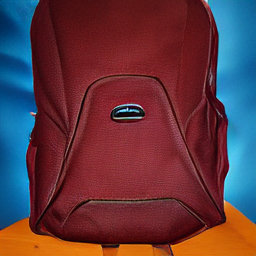}
\cfgimg{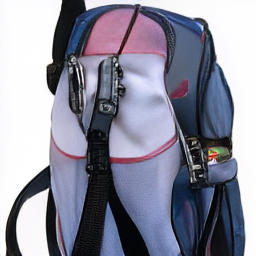}
\cfgimg{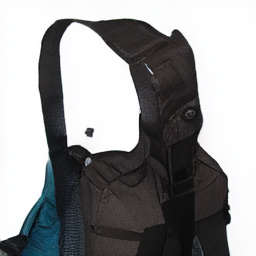}
\cfgimg{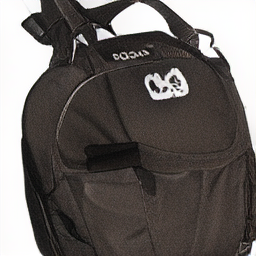}
\cfgimg{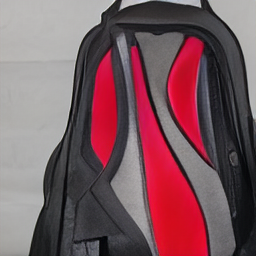}\\[4pt]

{\small LlamaGen-L + ours}\\[2pt]
\cfgimg{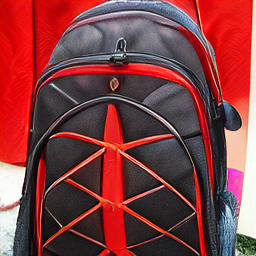}
\cfgimg{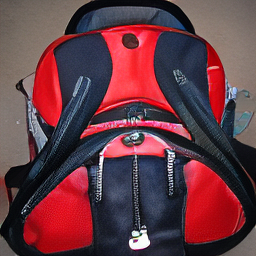}
\cfgimg{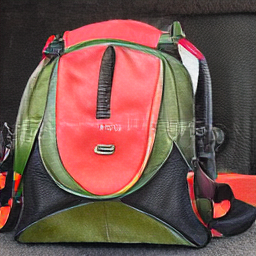}
\cfgimg{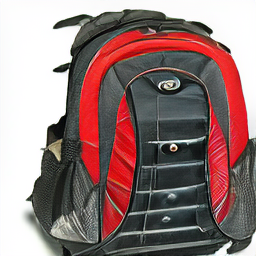}
\cfgimg{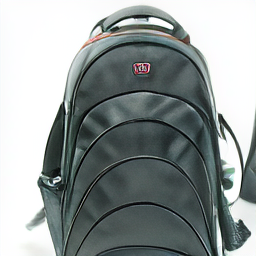}\\[12pt]

\textbf{Bakery}\\[2pt]
{\small LlamaGen-L}\\[2pt]
\cfgimg{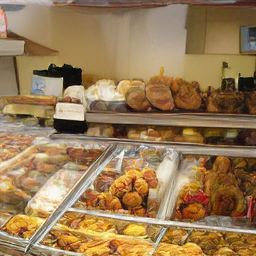}
\cfgimg{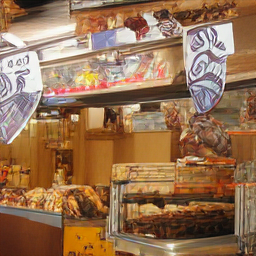}
\cfgimg{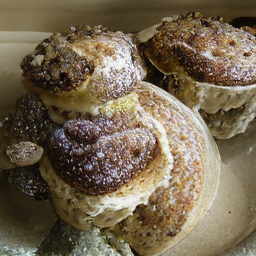}
\cfgimg{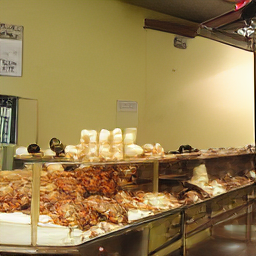}
\cfgimg{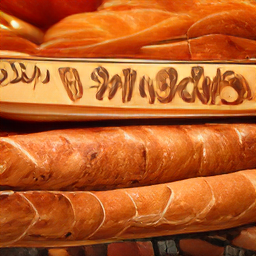}\\[4pt]

{\small LlamaGen-L + ours}\\[2pt]
\cfgimg{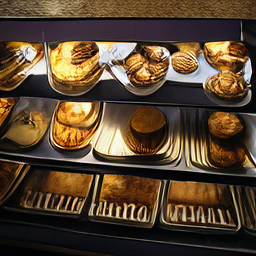}
\cfgimg{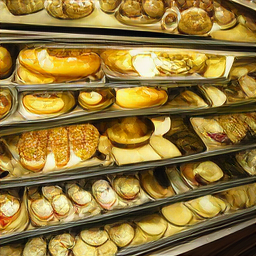}
\cfgimg{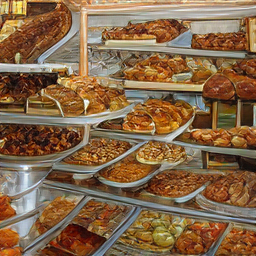}
\cfgimg{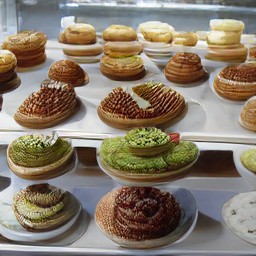}
\cfgimg{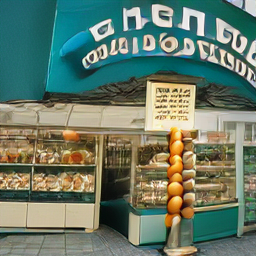}\\[4pt]

\caption{
Additional qualitative comparison between pretrained
\texttt{LlamaGen-L} and our policy fine-tuned model with
classifier-free guidance. \textbf{Backpack:} The pretrained model often produces backpacks with slightly distorted shapes. Our generations show better-defined structural elements, giving the images a cleaner and more coherent overall aesthetic. \textbf{Bakery:} Both models produce recognizable indoor scenes, but the pretrained outputs more often contain warped shelves or irregular object placement.
}
\label{fig:cfg_page_6}
\end{figure*}

\providecommand{\cfgimg}[1]{\includegraphics[width=0.19\linewidth]{#1}}

\begin{figure*}[t]
\centering

\textbf{Balloon}\\[2pt]
{\small LlamaGen-L}\\[2pt]
\cfgimg{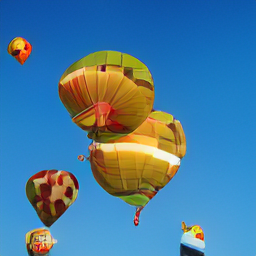}
\cfgimg{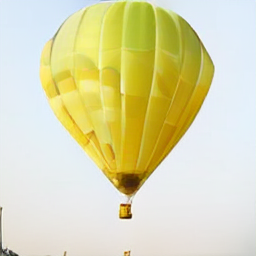}
\cfgimg{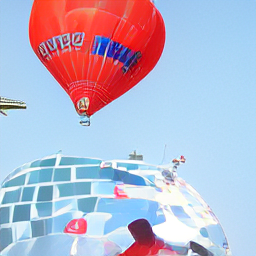}
\cfgimg{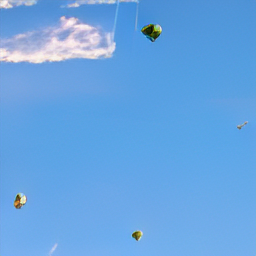}
\cfgimg{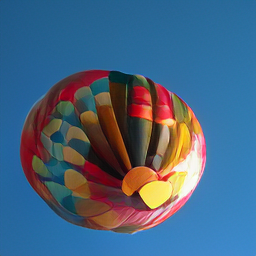}\\[4pt]

{\small LlamaGen-L + ours}\\[2pt]
\cfgimg{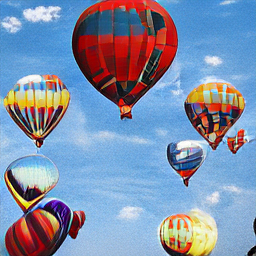}
\cfgimg{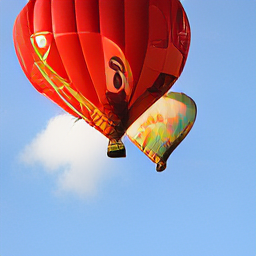}
\cfgimg{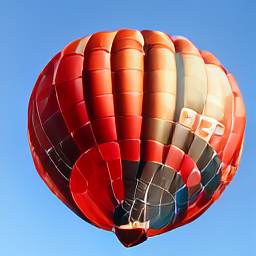}
\cfgimg{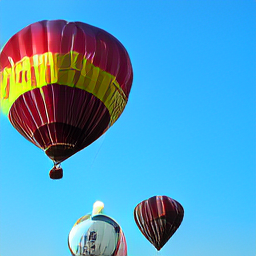}
\cfgimg{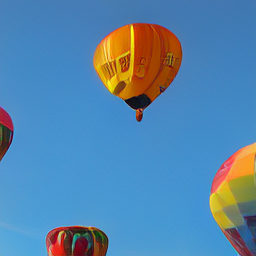}\\[12pt]

\textbf{Boathouse}\\[2pt]
{\small LlamaGen-L}\\[2pt]
\cfgimg{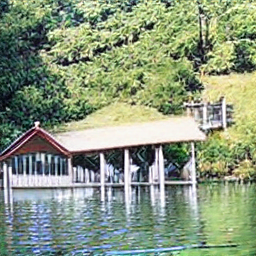}
\cfgimg{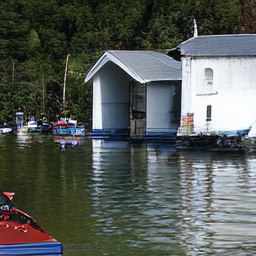}
\cfgimg{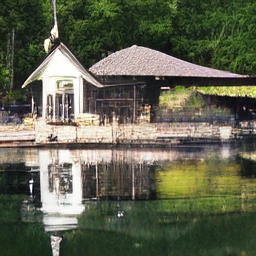}
\cfgimg{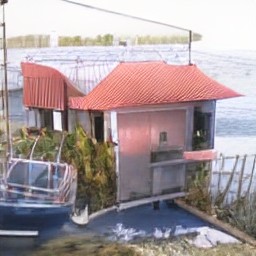}
\cfgimg{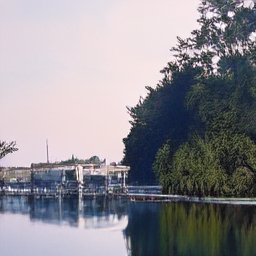}\\[4pt]

{\small LlamaGen-L + ours}\\[2pt]
\cfgimg{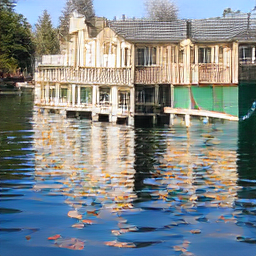}
\cfgimg{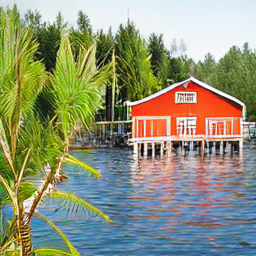}
\cfgimg{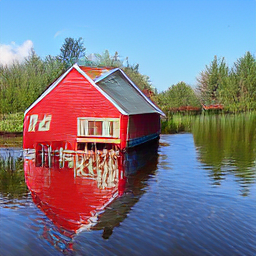}
\cfgimg{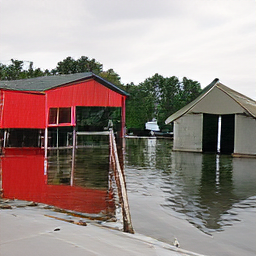}
\cfgimg{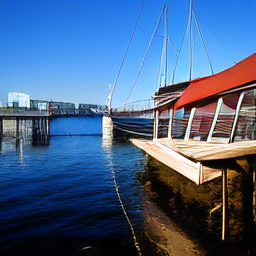}\\[4pt]

\caption{
Additional qualitative comparison between pretrained
\texttt{LlamaGen-L} and our policy fine-tuned model with
classifier-free guidance. \textbf{Balloon:} The pretrained model often produces balloons with irregular outlines or distorted surface shading. Our generations show smoother, more symmetric shapes, resulting in images that appear more visually coherent. \textbf{Boathouse:} The pretrained model sometimes produces buildings with less distinct edges or inconsistent roof geometry. Our generations tend to show clearer structural lines and more stable scene composition, yielding images that look more coherent and visually well-formed.
}
\label{fig:cfg_page_7}
\end{figure*}

\providecommand{\cfgimg}[1]{\includegraphics[width=0.19\linewidth]{#1}}

\begin{figure*}[t]
\centering

\textbf{Cinema}\\[2pt]
{\small LlamaGen-L}\\[2pt]
\cfgimg{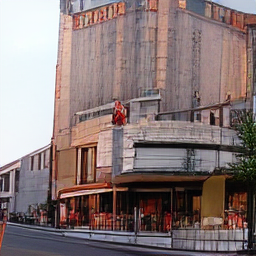}
\cfgimg{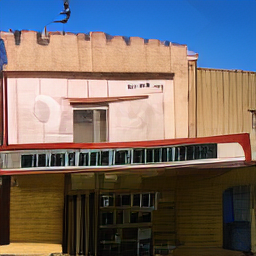}
\cfgimg{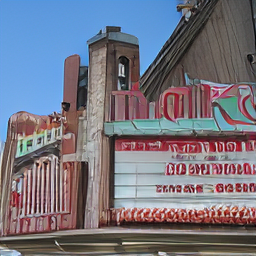}
\cfgimg{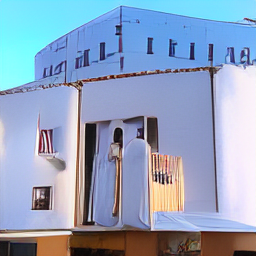}
\cfgimg{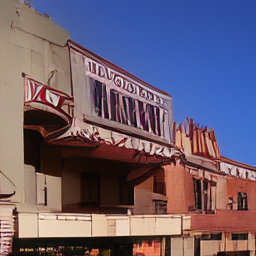}\\[4pt]

{\small LlamaGen-L + ours}\\[2pt]
\cfgimg{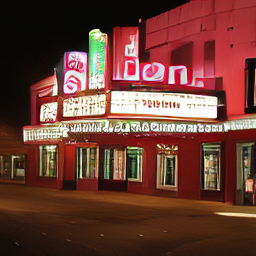}
\cfgimg{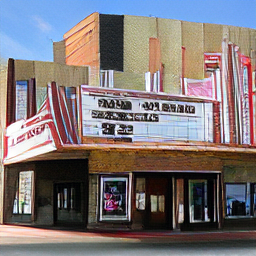}
\cfgimg{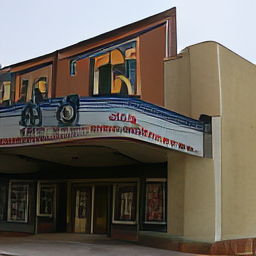}
\cfgimg{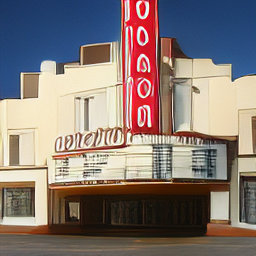}
\cfgimg{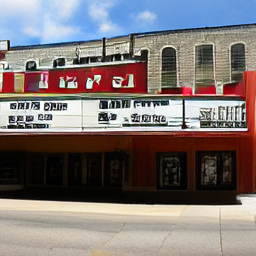}\\[12pt]

\textbf{Coffee Mug}\\[2pt]
{\small LlamaGen-L}\\[2pt]
\cfgimg{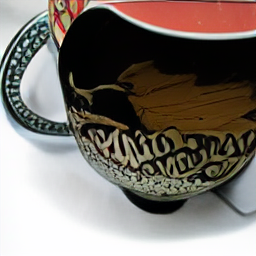}
\cfgimg{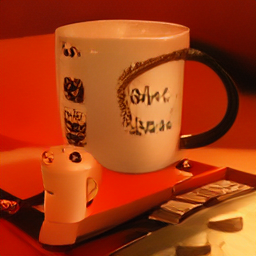}
\cfgimg{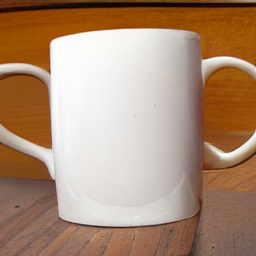}
\cfgimg{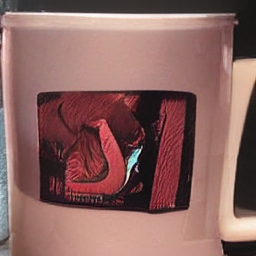}
\cfgimg{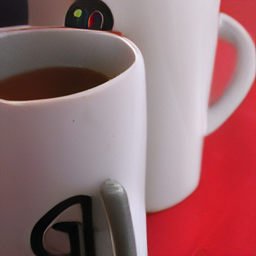}\\[4pt]

{\small LlamaGen-L + ours}\\[2pt]
\cfgimg{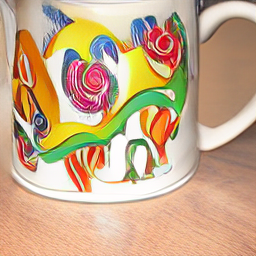}
\cfgimg{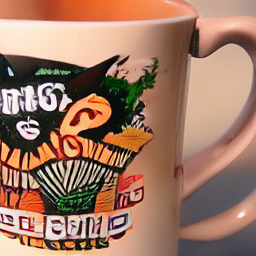}
\cfgimg{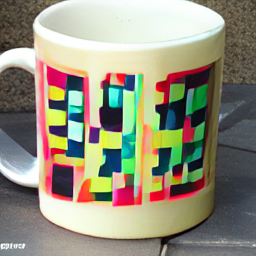}
\cfgimg{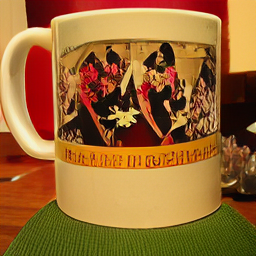}
\cfgimg{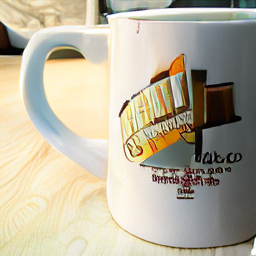}\\[4pt]

\caption{
Additional qualitative comparison between pretrained
\texttt{LlamaGen-L} and our policy fine-tuned model with
classifier-free guidance. \textbf{Cinema:} Both models produce plausible cinema façades. In our generations, the building structure and marquee region typically appear more coherent and readable, whereas the pretrained model more often exhibits distortions in the signage and surrounding geometry. \textbf{Coffee Mug:} The pretrained model produces mugs that frequently exhibits distorted geometry. Our fine-tuned model generates more coherent and visually structured prints on the mug surface. Overall, our samples better capture the intended decorative patterns while maintaining plausible mug geometry.
}
\label{fig:cfg_page_8}
\end{figure*}

\providecommand{\cfgimg}[1]{\includegraphics[width=0.19\linewidth]{#1}}

\begin{figure*}[t]
\centering

\textbf{Crate}\\[2pt]
{\small LlamaGen-L}\\[2pt]
\cfgimg{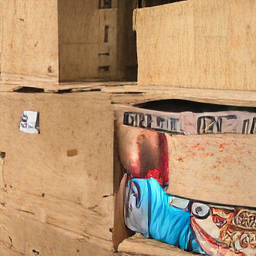}
\cfgimg{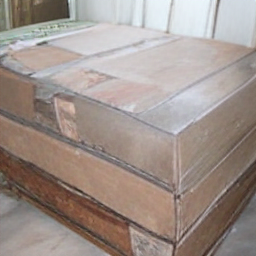}
\cfgimg{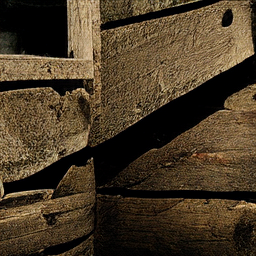}
\cfgimg{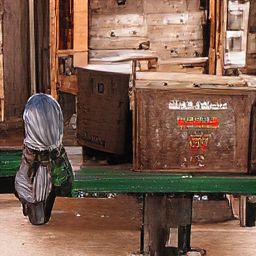}
\cfgimg{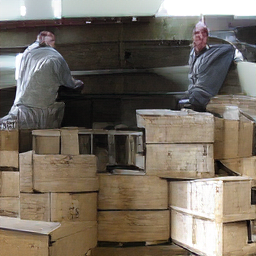}\\[4pt]

{\small LlamaGen-L + ours}\\[2pt]
\cfgimg{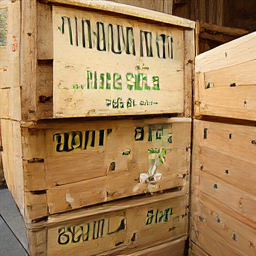}
\cfgimg{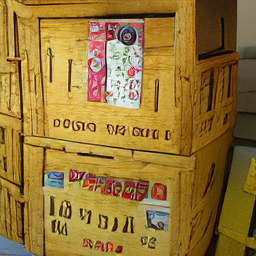}
\cfgimg{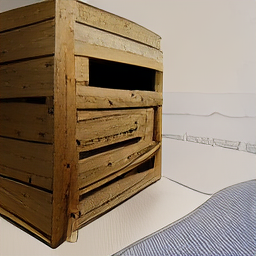}
\cfgimg{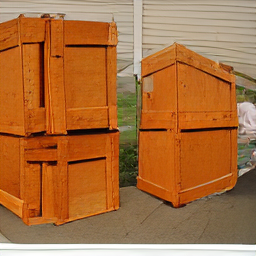}
\cfgimg{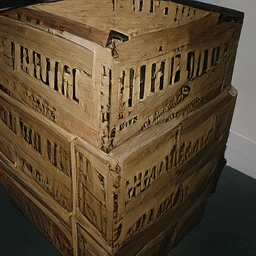}\\[12pt]

\textbf{File}\\[2pt]
{\small LlamaGen-L}\\[2pt]
\cfgimg{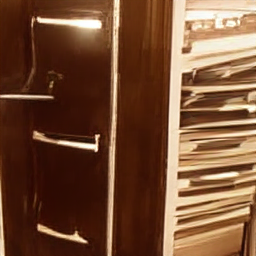}
\cfgimg{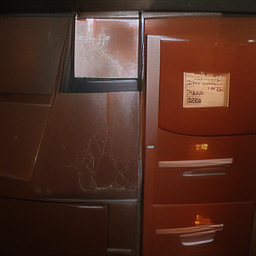}
\cfgimg{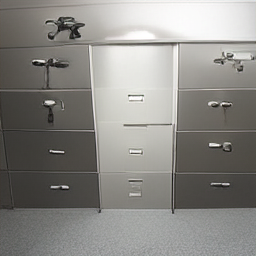}
\cfgimg{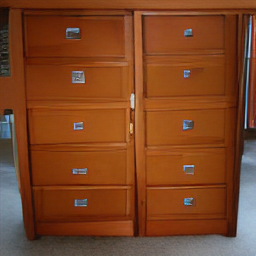}
\cfgimg{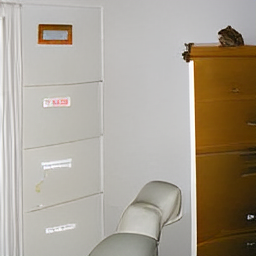}\\[4pt]

{\small LlamaGen-L + ours}\\[2pt]
\cfgimg{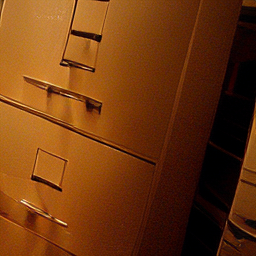}
\cfgimg{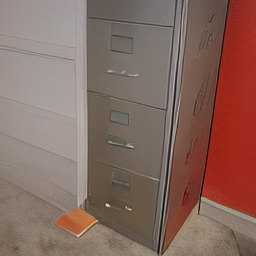}
\cfgimg{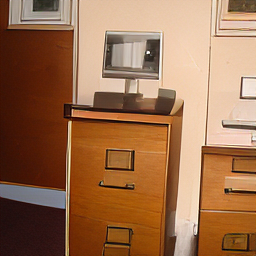}
\cfgimg{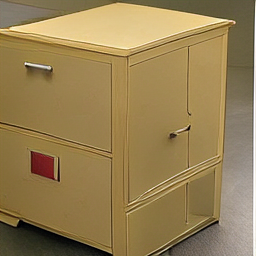}
\cfgimg{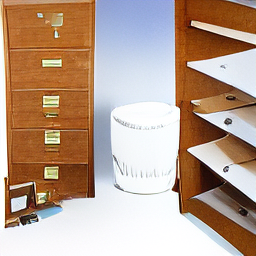}\\[4pt]

\caption{
Additional qualitative comparison between pretrained
\texttt{LlamaGen-L} and our policy fine-tuned model with
classifier-free guidance. \textbf{Crate:} The pretrained model frequently produces crates with inconsistent wood textures, or partially corrupted object boundaries. Our generations exhibit more regular box geometry, clearer structure, and more coherent wood patterns, resulting in crates that appear more stable and visually plausible overall. \textbf{File:} Our model produces cleaner drawer alignment, more consistent handle placement, and generally better-defined cabinet geometry, yielding images that look more orderly and realistic.
}
\label{fig:cfg_page_9}
\end{figure*}

\providecommand{\cfgimg}[1]{\includegraphics[width=0.19\linewidth]{#1}}

\begin{figure*}[t]
\centering

\textbf{Forklift}\\[2pt]
{\small LlamaGen-L}\\[2pt]
\cfgimg{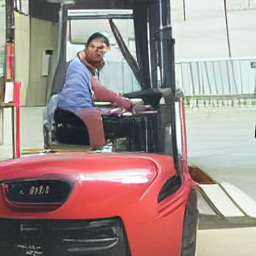}
\cfgimg{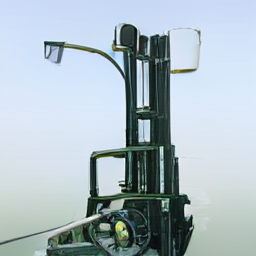}
\cfgimg{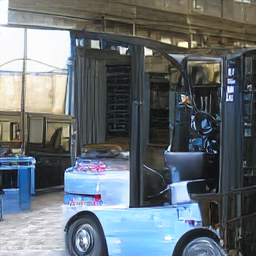}
\cfgimg{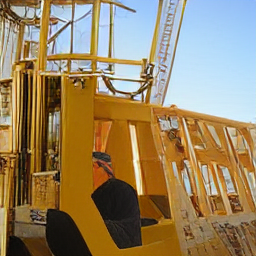}
\cfgimg{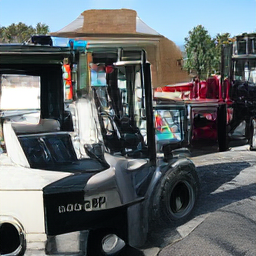}\\[4pt]

{\small LlamaGen-L + ours}\\[2pt]
\cfgimg{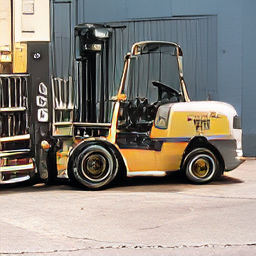}
\cfgimg{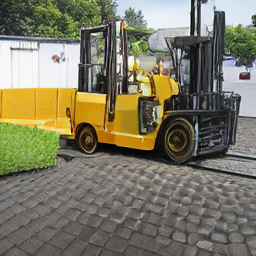}
\cfgimg{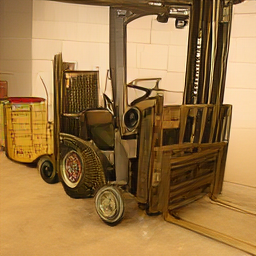}
\cfgimg{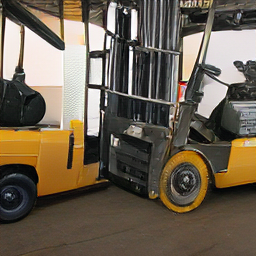}
\cfgimg{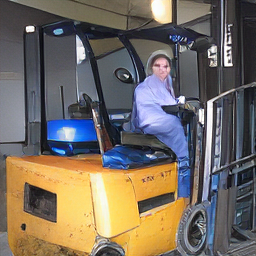}\\[12pt]

\textbf{Greenhouse}\\[2pt]
{\small LlamaGen-L}\\[2pt]
\cfgimg{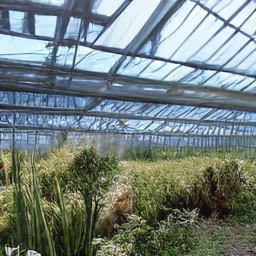}
\cfgimg{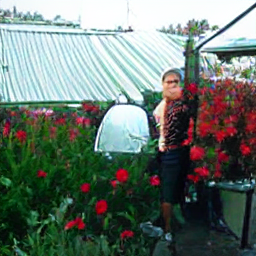}
\cfgimg{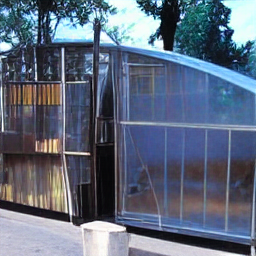}
\cfgimg{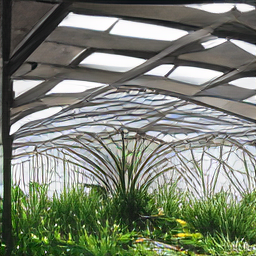}
\cfgimg{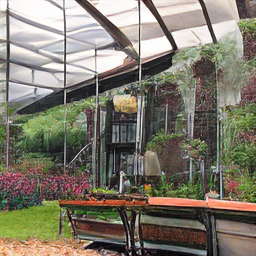}\\[4pt]

{\small LlamaGen-L + ours}\\[2pt]
\cfgimg{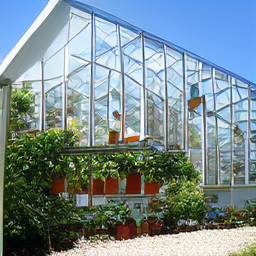}
\cfgimg{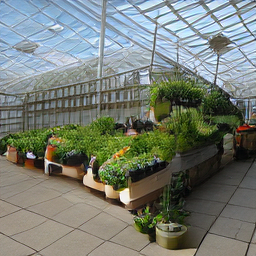}
\cfgimg{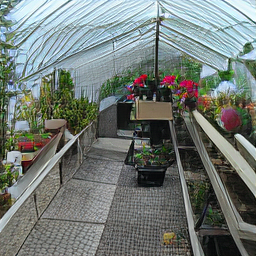}
\cfgimg{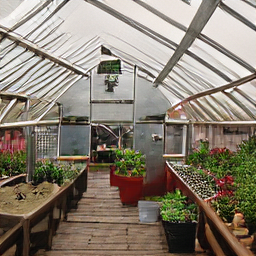}
\cfgimg{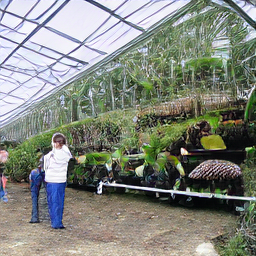}\\[4pt]

\caption{
Additional qualitative comparison between pretrained
\texttt{LlamaGen-L} and our policy fine-tuned model with
classifier-free guidance. \textbf{Forklift:} The pretrained model often produces forklifts with incomplete or distorted mechanical parts, and several samples show structural artifacts. Our generations exhibit more coherent vehicle geometry resulting in forklifts that appear more plausible and visually stable. \textbf{Greenhouse:} Our model produces images with clearer glass framing, more coherent interior organization, and more realistic plant arrangements, leading to greenhouses that read as cleaner and more structurally consistent.
}
\label{fig:cfg_page_10}
\end{figure*}

\providecommand{\cfgimg}[1]{\includegraphics[width=0.19\linewidth]{#1}}

\begin{figure*}[t]
\centering

\textbf{Handkerchief}\\[2pt]
{\small LlamaGen-L}\\[2pt]
\cfgimg{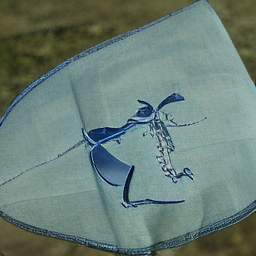}
\cfgimg{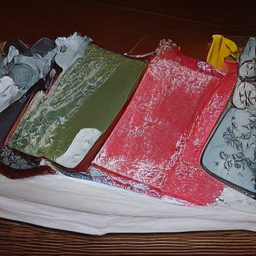}
\cfgimg{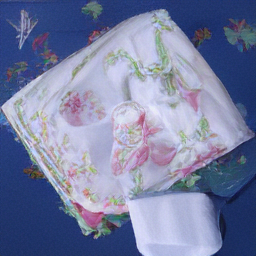}
\cfgimg{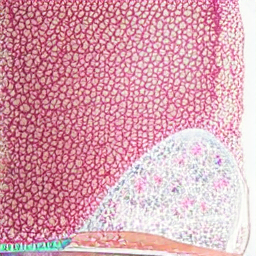}
\cfgimg{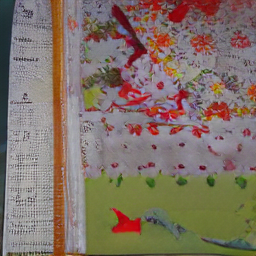}\\[4pt]

{\small LlamaGen-L + ours}\\[2pt]
\cfgimg{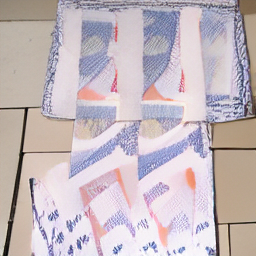}
\cfgimg{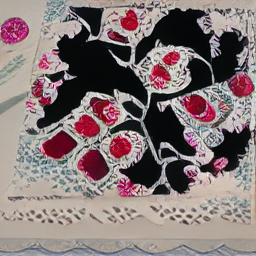}
\cfgimg{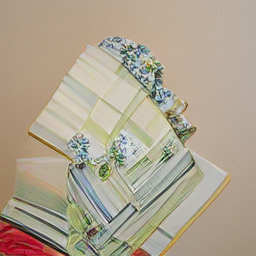}
\cfgimg{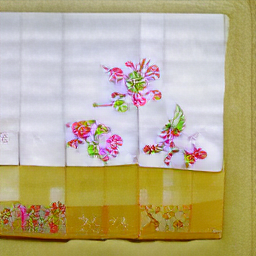}
\cfgimg{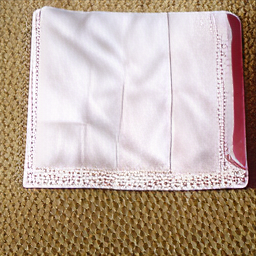}\\[12pt]

\textbf{Hourglass}\\[2pt]
{\small LlamaGen-L}\\[2pt]
\cfgimg{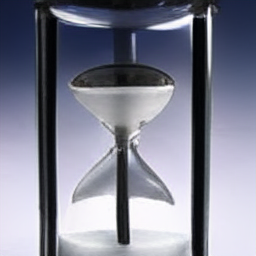}
\cfgimg{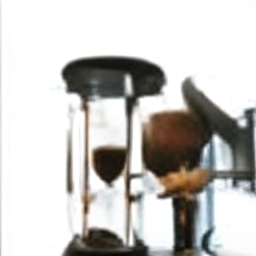}
\cfgimg{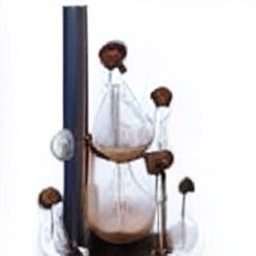}
\cfgimg{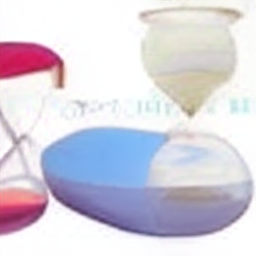}
\cfgimg{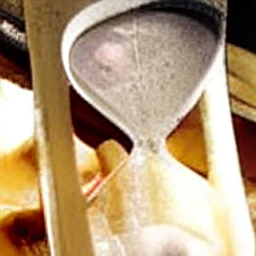}\\[4pt]

{\small LlamaGen-L + ours}\\[2pt]
\cfgimg{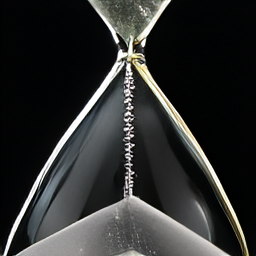}
\cfgimg{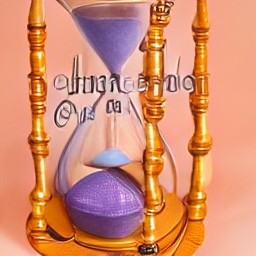}
\cfgimg{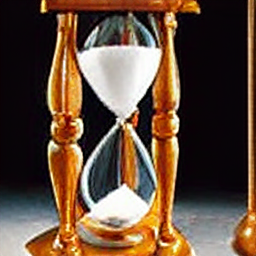}
\cfgimg{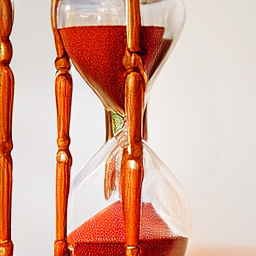}
\cfgimg{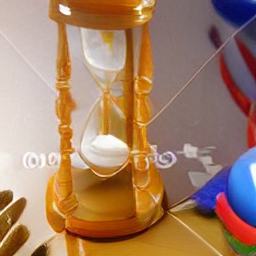}\\[4pt]

\caption{
Additional qualitative comparison between pretrained
\texttt{LlamaGen-L} and our policy fine-tuned model with
classifier-free guidance. \textbf{Handkerchief: } Both models produce images of comparable visual quality for this class. \textbf{Hourglass:} Pretrained samples frequently exhibit distorted frame shapes. Our model produces hourglasses with more symmetric frame geometry, clearer separation between the upper and lower chambers, and smoother visual structure, resulting in images that more reliably reflect the intended object.
}
\label{fig:cfg_page_11}
\end{figure*}

\providecommand{\cfgimg}[1]{\includegraphics[width=0.19\linewidth]{#1}}

\begin{figure*}[t]
\centering

\textbf{Ipod}\\[2pt]
{\small LlamaGen-L}\\[2pt]
\cfgimg{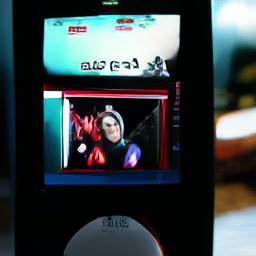}
\cfgimg{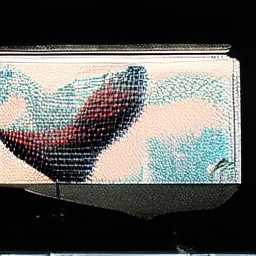}
\cfgimg{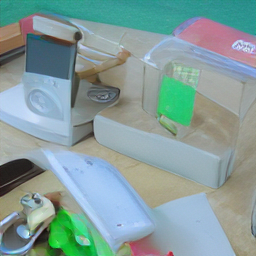}
\cfgimg{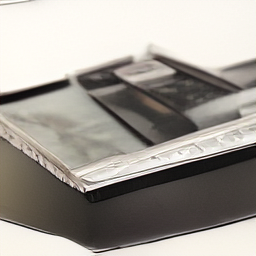}
\cfgimg{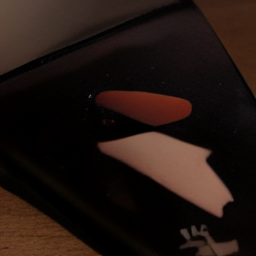}\\[4pt]

{\small LlamaGen-L + ours}\\[2pt]
\cfgimg{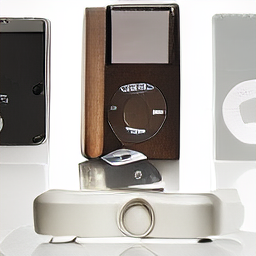}
\cfgimg{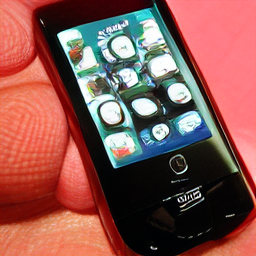}
\cfgimg{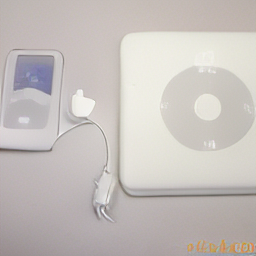}
\cfgimg{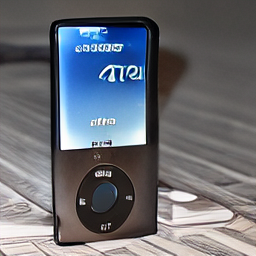}
\cfgimg{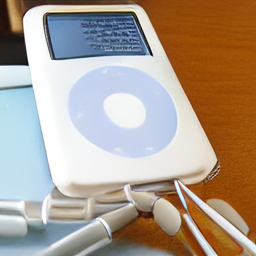}\\[12pt]

\textbf{Mailbox}\\[2pt]
{\small LlamaGen-L}\\[2pt]
\cfgimg{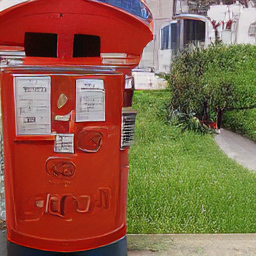}
\cfgimg{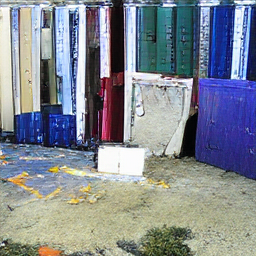}
\cfgimg{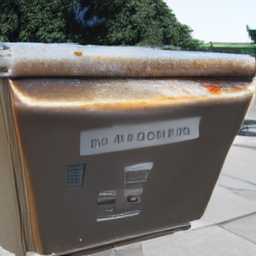}
\cfgimg{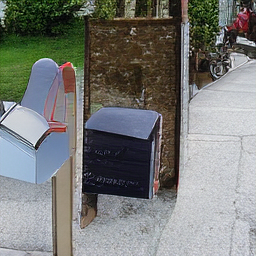}
\cfgimg{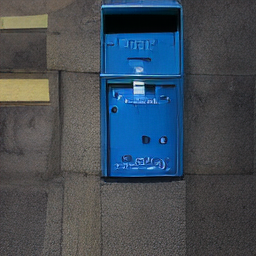}\\[4pt]

{\small LlamaGen-L + ours}\\[2pt]
\cfgimg{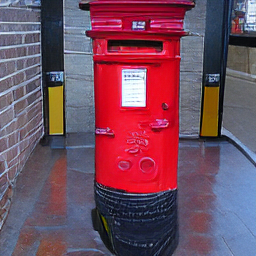}
\cfgimg{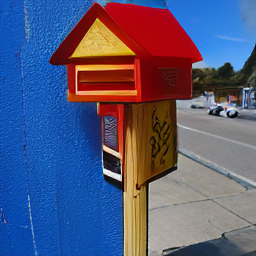}
\cfgimg{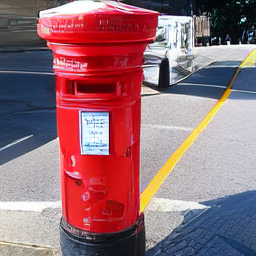}
\cfgimg{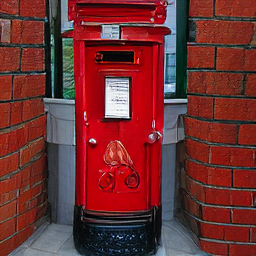}
\cfgimg{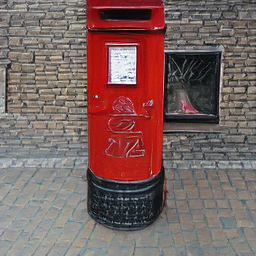}\\[4pt]

\caption{
Additional qualitative comparison between pretrained
\texttt{LlamaGen-L} and our policy fine-tuned model with
classifier-free guidance. \textbf{iPod:} Our model produces units with clearer geometry, more recognizable screen interfaces, and more stable overall proportions, yielding images that better reflect the intended device. \textbf{Mailbox:} Pretrained samples frequently contain irregular shapes or unstable surface details. Our generations offer cleaner cylindrical forms, more consistent mailbox openings, and improved color and material coherence, resulting in images that are easier to interpret as real-world mailboxes.
}
\label{fig:cfg_page_12}
\end{figure*}

\providecommand{\cfgimg}[1]{\includegraphics[width=0.19\linewidth]{#1}}

\begin{figure*}[t]
\centering

\textbf{Palace}\\[2pt]
{\small LlamaGen-L}\\[2pt]
\cfgimg{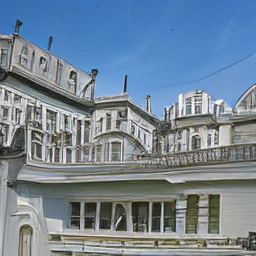}
\cfgimg{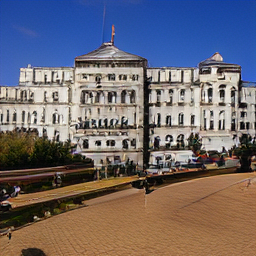}
\cfgimg{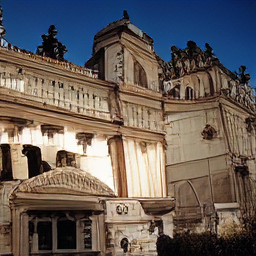}
\cfgimg{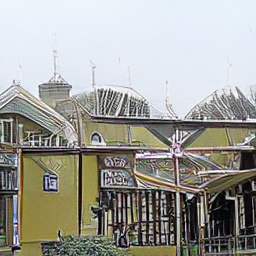}
\cfgimg{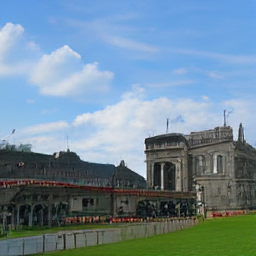}\\[4pt]

{\small LlamaGen-L + ours}\\[2pt]
\cfgimg{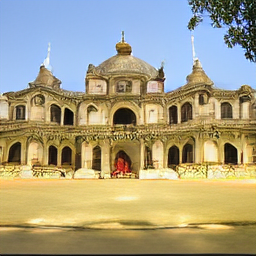}
\cfgimg{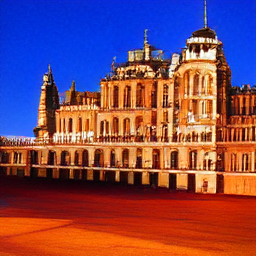}
\cfgimg{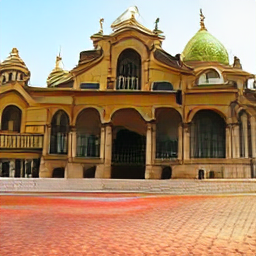}
\cfgimg{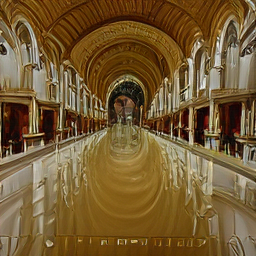}
\cfgimg{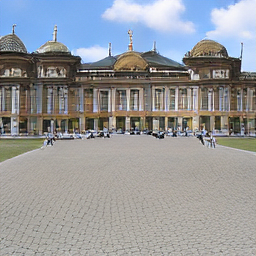}\\[12pt]

\textbf{Pillow}\\[2pt]
{\small LlamaGen-L}\\[2pt]
\cfgimg{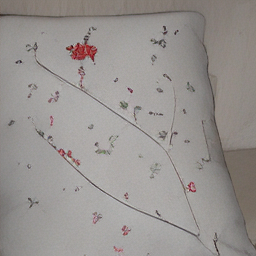}
\cfgimg{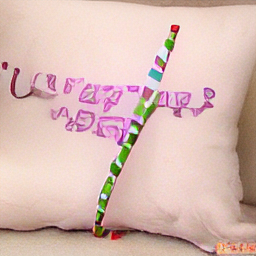}
\cfgimg{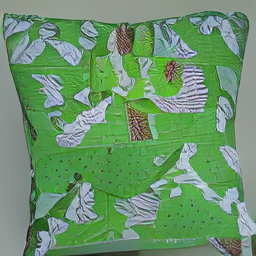}
\cfgimg{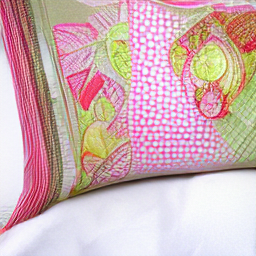}
\cfgimg{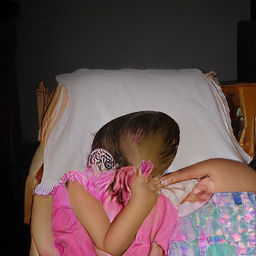}\\[4pt]

{\small LlamaGen-L + ours}\\[2pt]
\cfgimg{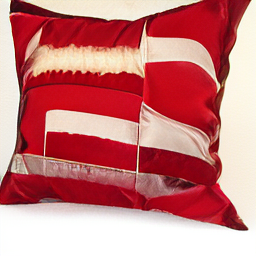}
\cfgimg{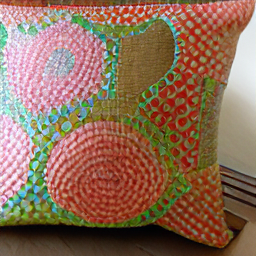}
\cfgimg{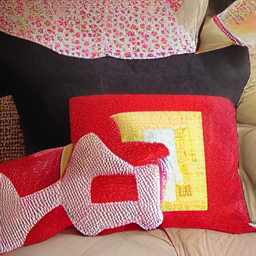}
\cfgimg{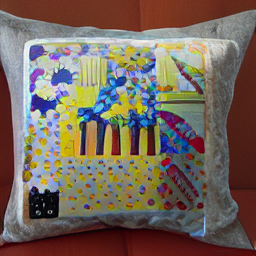}
\cfgimg{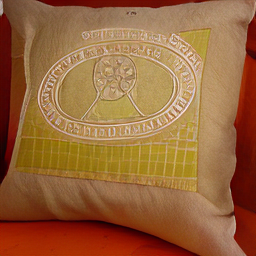}\\[4pt]

\caption{
Additional qualitative comparison between pretrained
\texttt{LlamaGen-L} and our policy fine-tuned model with
classifier-free guidance. \textbf{Palace:} Our model produces palaces with clearer structural organization, more stable rooflines, and better-preserved details, resulting in images that read as more coherent and architecturally plausible. \textbf{Pillow:} Pretrained samples occasionally contain distorted shapes or fabric textures that appear uneven or partially collapsed. Our generations show smoother contours, more natural fabric shading, and more consistent pillow geometry.
}
\label{fig:cfg_page_13}
\end{figure*}

\providecommand{\cfgimg}[1]{\includegraphics[width=0.19\linewidth]{#1}}

\begin{figure*}[t]
\centering

\textbf{Table Lamp}\\[2pt]
{\small LlamaGen-L}\\[2pt]
\cfgimg{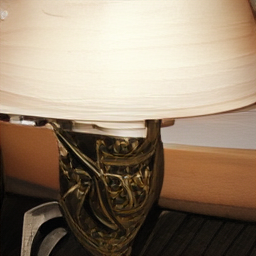}
\cfgimg{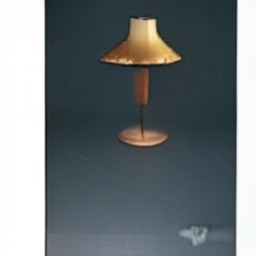}
\cfgimg{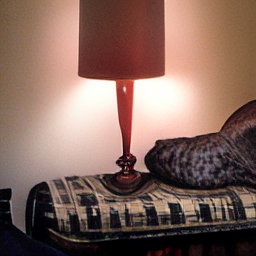}
\cfgimg{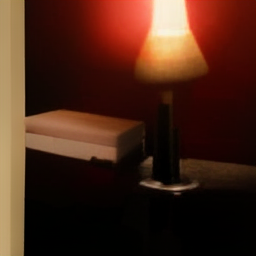}
\cfgimg{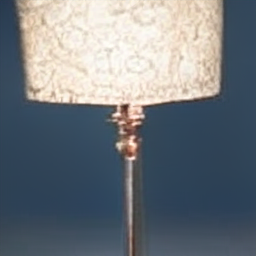}\\[4pt]

{\small LlamaGen-L + ours}\\[2pt]
\cfgimg{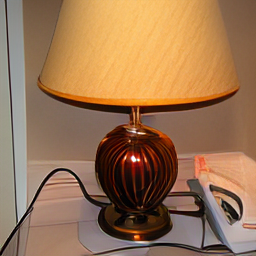}
\cfgimg{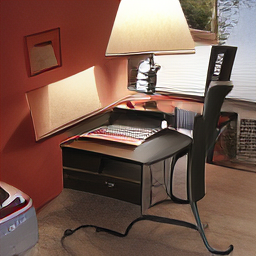}
\cfgimg{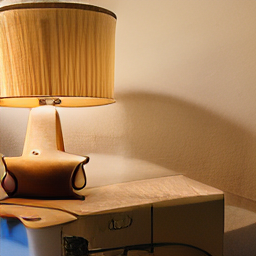}
\cfgimg{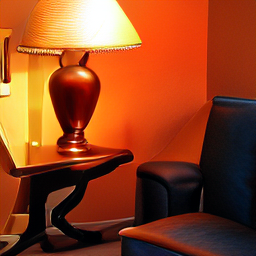}
\cfgimg{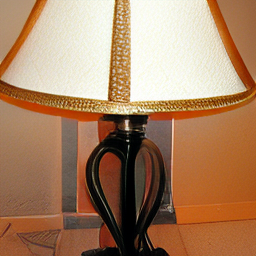}\\[12pt]

\textbf{Vase}\\[2pt]
{\small LlamaGen-L}\\[2pt]
\cfgimg{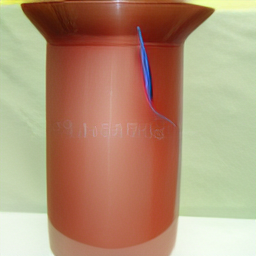}
\cfgimg{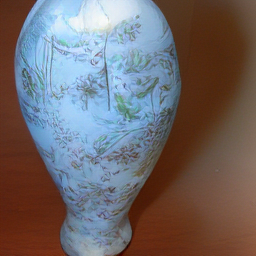}
\cfgimg{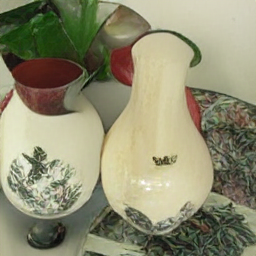}
\cfgimg{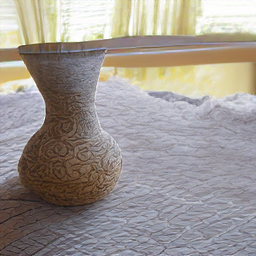}
\cfgimg{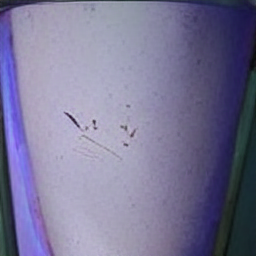}\\[4pt]

{\small LlamaGen-L + ours}\\[2pt]
\cfgimg{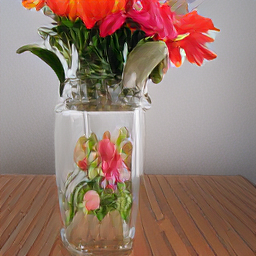}
\cfgimg{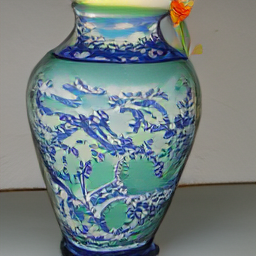}
\cfgimg{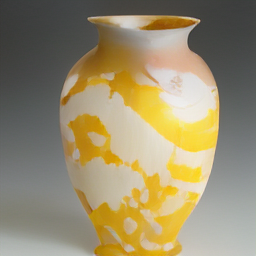}
\cfgimg{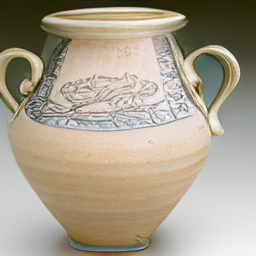}
\cfgimg{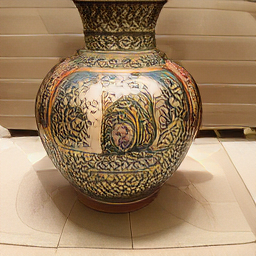}\\[4pt]

\caption{
Additional qualitative comparison between pretrained
\texttt{LlamaGen-L} and our policy fine-tuned model with
classifier-free guidance. \textbf{Table Lamp:} Our generations contain more stable lamp silhouettes, more consistent geometry, and more details within the scene, resulting in table lamps that appear more realistic and structurally consistent. \textbf{Vase:} Our model yields vases with smoother profiles, better-defined openings and necks, and more coherent decorative textures, producing images that more reliably reflect the intended object category.
}
\label{fig:cfg_page_14}
\end{figure*}

\providecommand{\cfgimg}[1]{\includegraphics[width=0.19\linewidth]{#1}}

\begin{figure*}[t]
\centering

\textbf{Water Tower}\\[2pt]
{\small LlamaGen-L}\\[2pt]
\cfgimg{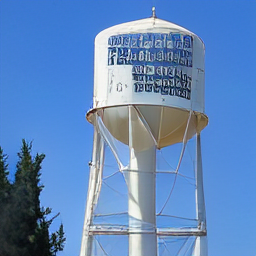}
\cfgimg{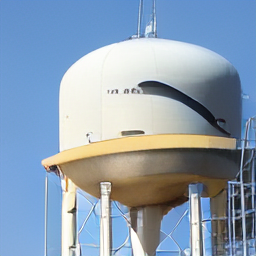}
\cfgimg{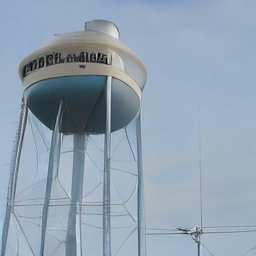}
\cfgimg{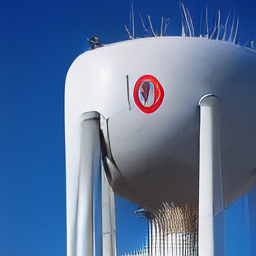}
\cfgimg{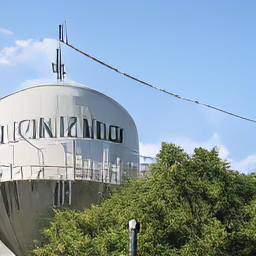}\\[4pt]

{\small LlamaGen-L + ours}\\[2pt]
\cfgimg{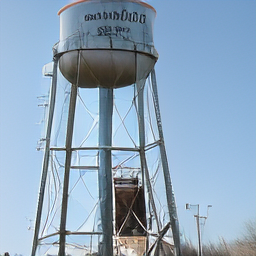}
\cfgimg{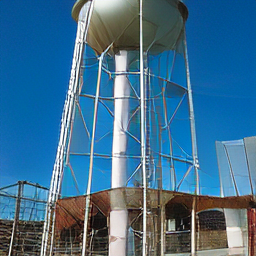}
\cfgimg{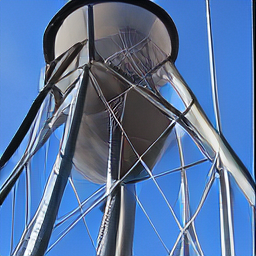}
\cfgimg{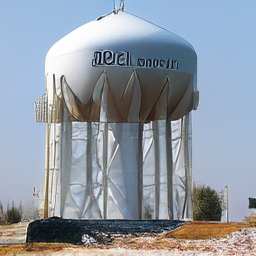}
\cfgimg{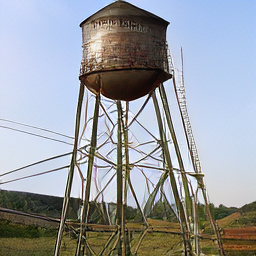}\\[12pt]

\textbf{Mushroom}\\[2pt]
{\small LlamaGen-L}\\[2pt]
\cfgimg{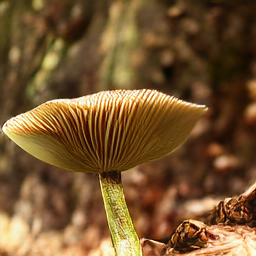}
\cfgimg{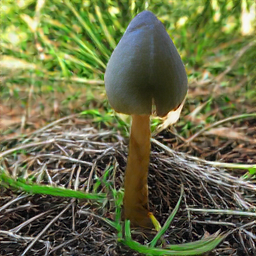}
\cfgimg{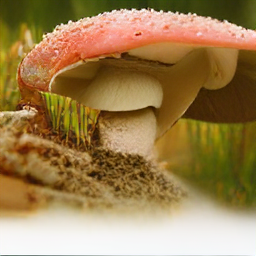}
\cfgimg{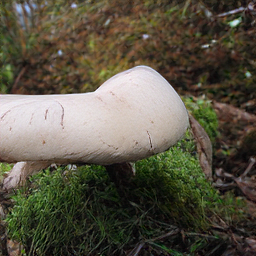}
\cfgimg{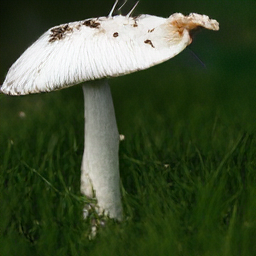}\\[4pt]

{\small LlamaGen-L + ours}\\[2pt]
\cfgimg{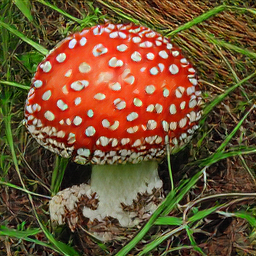}
\cfgimg{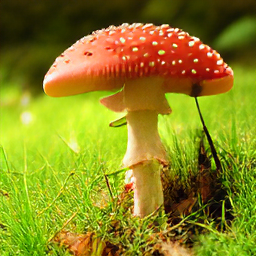}
\cfgimg{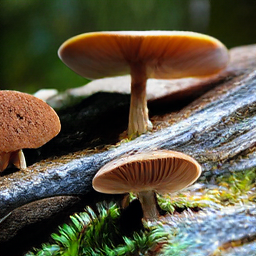}
\cfgimg{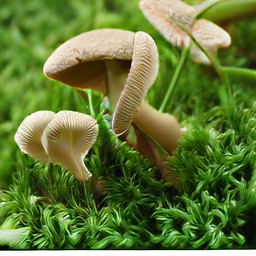}
\cfgimg{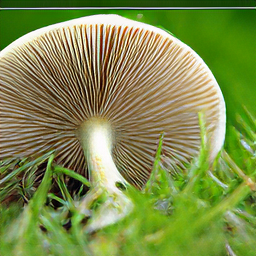}\\[4pt]

\caption{
Additional qualitative comparison between pretrained
\texttt{LlamaGen-L} and our policy fine-tuned model with
classifier-free guidance. \textbf{Water Tower:} The pretrained model often produces towers with warped tank shapes or inconsistent support structures. Our model generates water towers with cleaner cylindrical forms, more coherent leg geometry, and better separation between structural elements, resulting in images that read as more stable and realistic. \textbf{Mushroom:} Pretrained outputs occasionally display uneven cap shapes, or unclear gill structure. Our generations have smoother cap curvature, and more consistent geometry, yielding samples that more faithfully capture the natural morphology of the class.
}
\label{fig:cfg_page_15}
\end{figure*}

\providecommand{\cfgimg}[1]{\includegraphics[width=0.19\linewidth]{#1}}

\begin{figure*}[t]
\centering

\textbf{Pineapple}\\[2pt]
{\small LlamaGen-L}\\[2pt]
\cfgimg{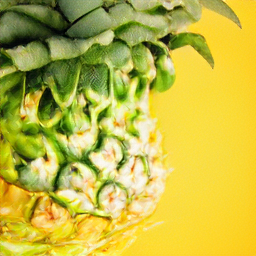}
\cfgimg{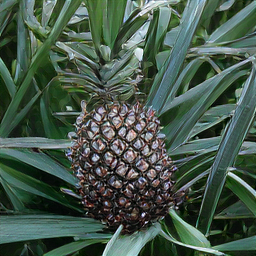}
\cfgimg{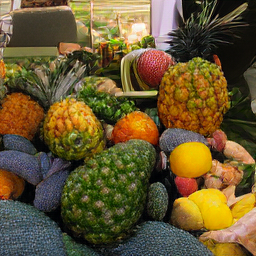}
\cfgimg{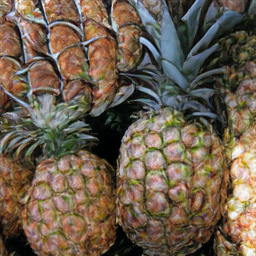}
\cfgimg{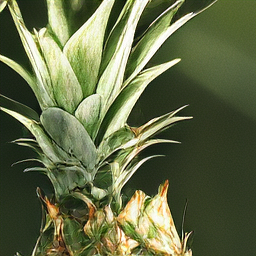}\\[4pt]

{\small LlamaGen-L + ours}\\[2pt]
\cfgimg{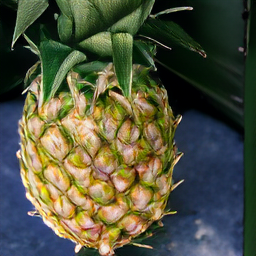}
\cfgimg{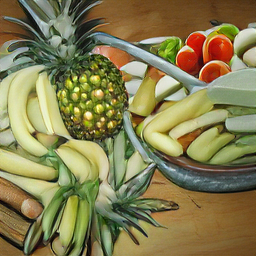}
\cfgimg{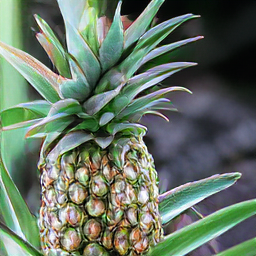}
\cfgimg{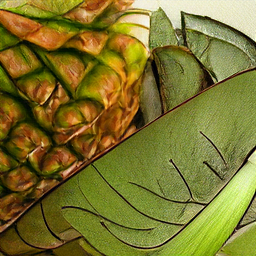}
\cfgimg{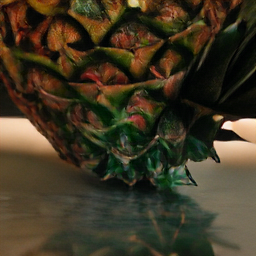}\\[12pt]

\textbf{Coral Reef}\\[2pt]
{\small LlamaGen-L}\\[2pt]
\cfgimg{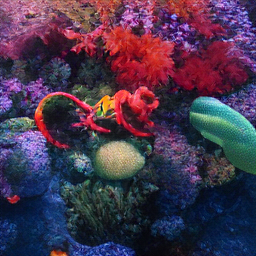}
\cfgimg{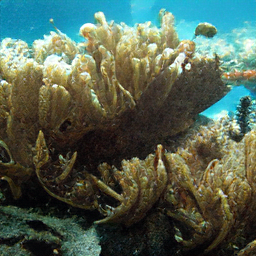}
\cfgimg{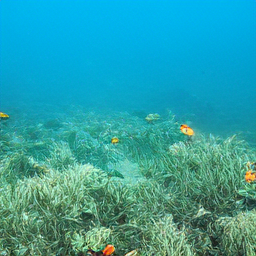}
\cfgimg{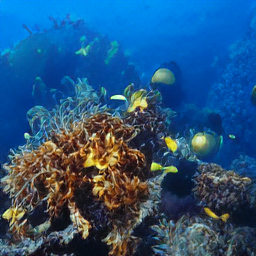}
\cfgimg{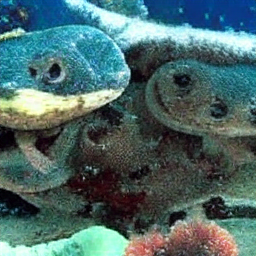}\\[4pt]

{\small LlamaGen-L + ours}\\[2pt]
\cfgimg{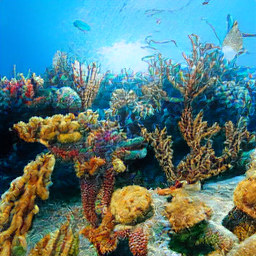}
\cfgimg{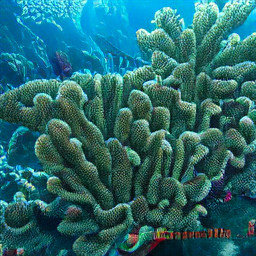}
\cfgimg{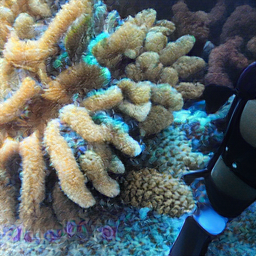}
\cfgimg{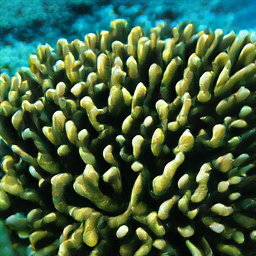}
\cfgimg{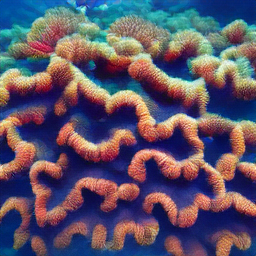}\\[4pt]

\caption{
Additional qualitative comparison between pretrained
\texttt{LlamaGen-L} and our policy fine-tuned model with
classifier-free guidance. \textbf{Pineapple:} Our generations show more coherent skin structure, better-organized leaf crowns, and improved overall fruit geometry, yielding images that more reliably capture the distinctive visual features of pineapples. \textbf{Coral Reef:} Pretrained samples sometimes contain muddled textures or blended coral forms that make the scene visually ambiguous. Our model produces clearer reef structures with more distinct elements, richer textural diversity, and better separation between foreground coral and background water, resulting in images that are more interpretable and easy to identify.
}
\label{fig:cfg_page_16}
\end{figure*}

\clearpage
\bibliographystyle{splncs04}
\bibliography{main}

\end{document}